\documentclass[twoside,11pt]{article}

\usepackage{jmlr2e}
\usepackage{graphicx}
\usepackage{forest}
\usepackage{tikz-qtree}
\usepackage{enumerate}
\usepackage{multirow}
\usepackage{indentfirst}
\usepackage{bm}
\usepackage{array}
\usepackage{subfigure}
\usepackage{soul}
\usepackage{amsmath}
\usepackage{color}
\usepackage{booktabs}
\usepackage{geometry}
\usepackage{colortbl}
\usepackage[normalem]{ulem}
\usepackage{longtable}
\usepackage{amssymb}
\usepackage{appendix}
\usepackage{hyperref}
\renewcommand{\arraystretch}{1.5} 

\usepackage[utf8]{inputenc}
\usepackage[TS1,T1]{fontenc}
\usepackage{array, booktabs}
\usepackage{graphicx}
\usepackage{xcolor}
\newcommand{\foo}{\color{gray}\makebox[0pt]{\textbullet}\hskip-0.5pt\vrule width 1pt\hspace{\labelsep}}
\usepackage{chronology}

\usepackage{lastpage}
\jmlrheading{23}{2024}{1-\pageref{LastPage}}{1/21; Revised 5/22}{9/22}{21-0000}{Wenjin Niu, Zijun Gao, Liyan Song \& Lingbo Li}

\ShortHeadings{Comprehensive Review and Empirical Evaluation of Causal Discovery}
{Niu, Gao, Song \& Li}
\firstpageno{1}

\begin{document}

\title{Comprehensive Review and Empirical Evaluation of Causal Discovery Algorithms for Numerical Data}

\author{\name Wenjin Niu \email wenjin.niu@warwick.ac.uk \\
       \addr The School of Engineering\\
       University of Warwick\\
       CV4 7AL Coventry, U.K.
       \AND
       \name Zijun Gao \email zijun.gao.1@warwick.ac.uk \\
       \addr The School of Engineering\\
       University of Warwick\\
       CV4 7AL Coventry, U.K.
       \AND
       \name Liyan Song \email songly@hit.edu.cn \\
       \addr Faculty of Computing\\
       Harbin Institute of Technology (HIT)\\
       92 Xida Street, Nangang District, 154001, Harbin, China
       \AND
       \name Lingbo Li \email lingbo.li.1@warwick.ac.uk \\
       \addr The School of Engineering\\
       University of Warwick\\
       CV4 7AL Coventry, U.K.\\
       State Key Laboratory for Novel Software Technology\\
       Nanjing University\\
       163 Xianlin Avenue, Nanjing, Jiangsu Province, China}

\editor{}

\maketitle

\begin{abstract}

Causal analysis has become an essential component in understanding the underlying causes of phenomena across various fields.
Despite its significance, existing literature on causal discovery algorithms is fragmented, with inconsistent methodologies, i.e., there is no universal classification standard for existing methods, and a lack of comprehensive evaluations, i.e., data characteristics are often ignored to be jointly analyzed when benchmarking algorithms.
This study addresses these gaps by conducting an exhaustive review and empirical evaluation for causal discovery methods on numerical data, aiming to provide a clearer and more structured understanding of the field.
Our research begins with a comprehensive literature review spanning over two decades, analyzing over 200 academic articles and identifying more than 40 representative algorithms.
This extensive analysis leads to the development of a structured taxonomy tailored to the complexities of causal discovery, categorizing methods into six main types.
To address the lack of comprehensive evaluations, our study conducts an extensive empirical assessment of 29 causal discovery algorithms on multiple synthetic and real-world datasets.
We categorize synthetic datasets based on size, linearity, and noise distribution, employing five evaluation metrics, and summarize the top-3 algorithm recommendations, providing guidelines for users in various data scenarios.
Our results highlight a significant impact of dataset characteristics on algorithm performance.
Moreover, a metadata extraction strategy with an accuracy exceeding 80\% is developed to assist users in algorithm selection on unknown datasets.
Based on these insights, we offer professional and practical guidelines to help users choose the most suitable causal discovery methods for their specific dataset.

\end{abstract}


\section{Introduction}
\label{chp:introduction}
Causality, as a dynamically evolving interdisciplinary field, has been gaining increasing attention from both academia and industry \citep{nogueira2021causal,menegozzo2021industrial,masson2021climate,ganguly2023review}. 
Causal analysis employs a systematic approach to uncovering the underlying causes of phenomena, primarily addressing the question of ``Why” behind observed trends.
Since Granger's seminal work in 1969 \citep{granger1969}, which introduced a mathematical concept of causality, the field has expanded from philosophy to economics \citep{imbens2004nonparametric} and other domains such as medicine \citep{mani2000causal}, environmental science \citep{li2014discovering}, and dynamics \citep{hu2015comparison}. 
Recently, the rapid advancement of Artificial Intelligence (AI) has opened new avenues for causal analysis. 
The integration of machine learning has enhanced the precision and efficiency of data processing for causal inference, while causal learning has established a more reliable and trustworthy framework for machine learning \citep{Scholkopf2019,makhlouf2020survey}. 
These two disciplines mutually reinforce each other, driving significant advancements in scientific research.

\begin{figure}[ht]
    \centering
    \includegraphics[scale=0.8]{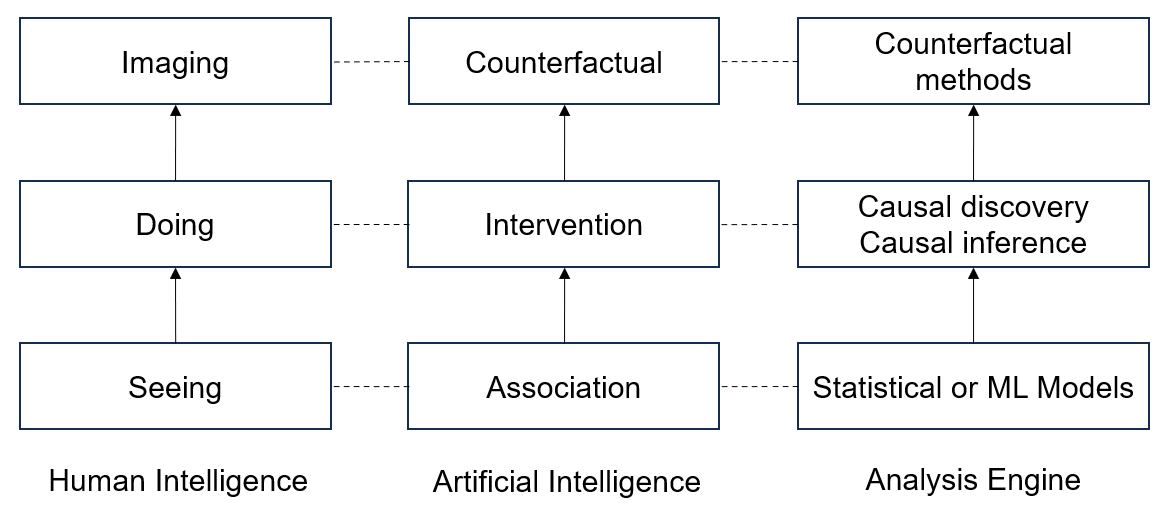}
    \caption{Ladder of Causation \citep{runge2023}}
    \label{fig:Ladder_of_causation}
     \end{figure}

Given that correlation does not imply causation, causal research necessitates a thorough investigation beyond simple association analysis. 
Pearl's work (\citeyear{pearl2000}) provided a widely accepted framework, known as the ``Ladder of Causation”, which delineates three stages: \textit{association}, \textit{intervention}, and \textit{counterfactual}. 
The initial stage, \textit{association}, involves observing relationships between variables, yet it is insufficient for identifying confounders or selection bias that may lead to spurious causation \citep{cheng2019robust}. 
The second stage, \textit{intervention}, involves controlled experiments to quantify the causal impact of one variable on another.
The final stage, \textit{counterfactual} analysis, requires a deep understanding of the causal mechanisms underlying the phenomena.
Figure \ref{fig:Ladder_of_causation} shows the causation ladder and corresponding analysis engine \citep{bareinboim2016causal}.

This section elaborates on the fundamental concepts, evolution, and classifications of causal analysis, addresses the limitations of previous studies, and introduces the innovative contributions and specific objectives of the current work.
Research on causal analysis is primarily categorized into two areas \citep{Nogueira2022}: \textit{causal inference} and \textit{causal discovery}. 
\textit{Causal inference} typically progresses from cause to effect, focusing on the quantitative problems of ``Intervention Stage” within the causation ladder framework \citep{peters2017elements}.
The central concept of inference involves controlled trials \citep{Yao2021}, where both experimental and control groups are observed to determine the effects of interventions.
When the causal graph is known, observational data can be used to predict intervention effects of experimental tests. Common approaches for causal effect estimation include covariate adjustment \citep{pearl2009causality,stekhoven2012causal,maathuis2015generalized}, optimal adjustment \citep{sekhon2008multivariate,runge2021necessary,henckel2022graphical}, and the path method \citep{nandy2017estimating}.
In contrast, \textit{causal discovery} seeks to identify causal relationships from observed outcomes, emphasizing the qualitative problems to learn causal structures \citep{gelman2011causality}. 
Upon uncovering causal mechanisms, it becomes possible to infer outcomes based on hypothetical scenarios that have not occurred.
Despite their inverse logical relationship, \textit{causal discovery} and \textit{causal inference} differ significantly in their research methodologies, algorithms, and applications.

In practical scenarios, a significant challenge in causal analysis is managing diverse and complex data types. 
Standard data formats include time series \citep{eichler2012causal}, cross-sectional, and panel data. 
Panel data, which has two dimensions (time and samples), combines elements of cross-sectional data, where all sample observations at a specific time point are included, and time series data, where observations of the same sample are recorded at different time points.
Since time series data does not adhere to the assumptions of independent and identically distributed (i.i.d.) random variables, it necessitates specialized research on data processing techniques and causal analysis algorithms. 
Furthermore, due to the importance and widespread use of time series in real-world applications, many researchers have focused extensively on this data type \citep{assaad2021mixed,biswas2024consistent}.
If a time series can be viewed as a ``list", then the i.i.d. variable is a ``set" without self-causes. Therefore, causal discovery for i.i.d. data, which differs from time-series causality, has also attracted a lot of research attention \citep{xie2019efficient}.

However, existing literature lacks a universally applicable algorithm for causal analysis \citep{edinburgh2021}.
Unlike \textit{causal inference}, the precision of \textit{causal discovery} heavily relies on the selection of an appropriate causality model.
Hence, users often face challenges in selecting the appropriate causal discovery algorithm on unknown datasets, leading to unsatisfactory results or unnecessary expenditure of computational resources and time.
The significance of this study including providing guidelines to help users quickly identify the most suitable algorithm for unknown datasets, and assisting researchers in organizing and benchmarking numerous existing algorithms.
Therefore, we concentrates on two pivotal elements of causal analysis: \textbf{(1) \textit{causal discovery}} and \textbf{(2) \textit{time series and i.i.d. data analysis}}. 
An exhaustive investigation was conducted to delve into the methodologies of causal discovery, encompassing principles, algorithmic strategies, and recent advancements in the field.


Extensive efforts have been made to review and reevaluate causal discovery algorithms in a unified, more extensive, and systematic way.
To address the challenge of limited benchmark datasets for causal discovery, many researchers have focused on developing data simulators in fields such as industrial systems \citep{menegozzo2022cipcad}, neurology \citep{tu2019neuropathic}, and biology \citep{ma2023local}.
There have been some studies that conducted experimental studies to systematically evaluate one type of (not all existing) causal discovery methods.
For example, Sogawa et al. (\citeyear{sogawa2010experimental}) evaluated the identification accuracy and robustness of linear non-Gaussian methods and its variants. Raghu et al. (\citeyear{Raghu2018}) compared the performance of four conditional independence-based algorithms on mixed data with latent variables. Ko et al. (\citeyear{ko2018experimental}) summarized estimation of distribution algorithms (EDAs) and compared their performance on four datasets to infer the best ones.

There have been articles that survey various types of methods. Song et al. (\citeyear{song2016evaluation}) and Käding et al. (\citeyear{kading2021benchmark}) compared causal discovery methods for bivariates on real-world bench datasets.
However, their research only focused on bivariate and did not investigate multivariate algorithms.
Ombadi et al. (\citeyear{ombadi2020evaluation}) evaluated four causal discovery algorithms on hydrometeorological data, aiming to guide researchers in determining which causal method is most appropriate based on the characteristics of hydrological system.
Assaad et al. (\citeyear{Assaad2022}) not only systematically organized time-series methodologies but also performed thorough evaluations of representative algorithms based on distinct causal structures.

Despite these contributions, most existing literature primarily focuses on theoretical summaries. 
Even when experiments were conducted, they mainly assessed the impact of causal structures on model performance. 
Given the often ambiguous causal structure of observational data, it is vital to provide practical and reliable insights from the user's perspective.

Unlike previous surveys, this paper adopts a data-oriented approach, categorizing causal relationships into four types: i.i.d. causality, time-delay causality, instantaneous causality, and causal pairs. 
To address the research gap that often overlooks the user's perspective, this study performs a comprehensive analysis starting from the intrinsic characteristics of the data (data assumptions).
Motivated by this approach, comparative experiments are conducted, treating data assumptions as experimental factors and employing various algorithms as experimental subjects. 
These comparisons aim to establish the relationship between algorithms and specific data features within each causality category.
Specifically, our pursuit is dedicated to identifying the optimal algorithm, considering various factors, including data size, linearity, stationarity, and noise attributes. 
By leveraging the extracted data features, users can choose the most appropriate algorithm for their specific needs. The main contributions of this paper are summarized as follows:

\begin{enumerate}
    \item \textbf{\textit{Survey and Taxonomy:}}
We comprehensively collect and categorize methods and algorithms for causal discovery, summarizing the characteristics and applications of these algorithms.

    \item \textbf{\textit{Benchmarking:}}
We conduct extensive benchmarking of selected state-of-the-art algorithms across diverse datasets using multiple evaluation metrics to access performance and applicability.
    \item \textbf{\textit{Practical Guidelines:}}
We provide practical insights and recommendations on the optimal algorithm for specific datasets, offering decision-making suggestions in various application fields.
\end{enumerate}





In light of this, our experiment focus on the analysis of algorithm performance and result effectiveness, with the aim of addressing the following research questions:\\

\textbf{RQ 1 (\textit{Comparison of algorithm performance}):} Among the assessed algorithms, which one demonstrates superior effectiveness or efficiency under specific data characteristics?

This research question can be considered as a benchmark and baseline for answering other questions.
By evaluating the impact of data features on algorithm performance, we establish a foundation that informs subsequent steps in our experimental analysis.\\

\textbf{RQ 2 (\textit{Real-world applicability}):} Are the insights derived from the synthetic datasets consistent with those acquired from the real datasets?

This research question is crucial for determining the effectiveness of insights gained from RQ 1, as it connects the findings from synthetic data to real-world scenarios. 
This question ensures that our conclusions are not limited to controlled experimental conditions but are also valid in practical applications.\\

\textbf{RQ 3 (\textit{Generalization to unknown datasets})}: This RQ can be further divided into the following sub-research questions:
\begin{itemize}
  \item \textbf{RQ 3.1 (\textit{Metadata recognition for algorithm selection}):} Is it feasible to precisely capture the representative attributes of unknown datasets using their metadata to ascertain the optimal algorithm based on our previous conclusions?
  \item \textbf{RQ 3.2 (\textit{Practical recommendations for users}):} Upon successfully identifying the optimal algorithm for an unknown dataset in RQ 3.1, what practical recommendations can we provide to users for selecting appropriate methods for their specific datasets?
\end{itemize}

This research question is challenging with respect to experimental justifications of the other RQs as it involves extending the results of RQ 1 and RQ 2 to a broader range of applications. 
RQ 3.1 focuses on the feasibility of applying our findings to new datasets by analyzing their metadata, ensuring that our methods are robust and versatile. 
RQ 3.2 aims to translate these validated approaches into practical, user-friendly guidelines that assist practitioners in choosing the best algorithms for their unique datasets, thus bridging the gap between theoretical research and practical implementation.

\begin{figure}[ht]
    \centering
    \includegraphics[scale=0.53]{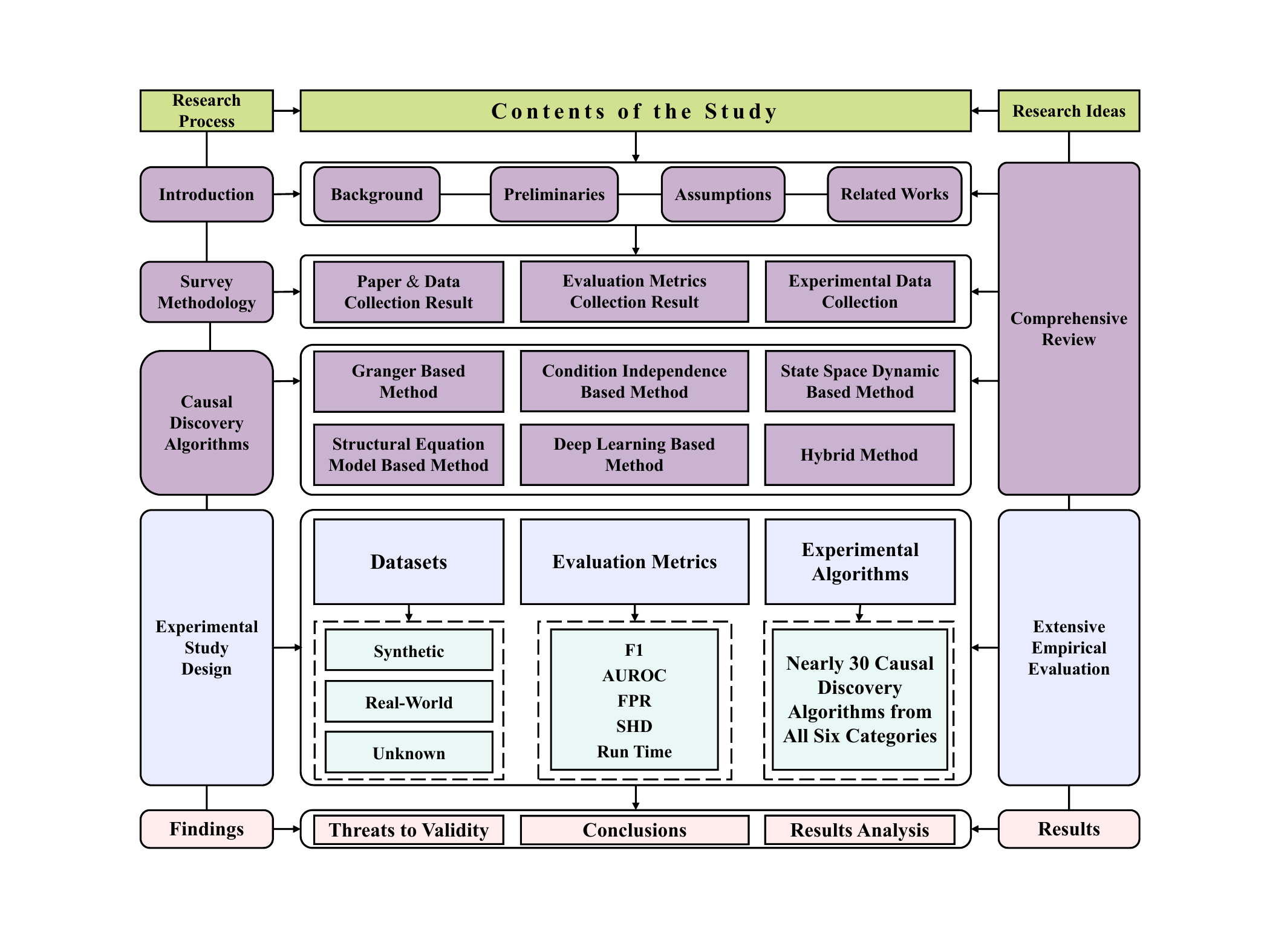}
    \caption{The Overall Organization of the Study}
    \label{fig:Paper_framework_diagram}
     \end{figure}

The rest of this paper is organized as follows. 
Section \ref{chp:literature-review} presents the methodology of conducting the literature review, in which Section \ref{chp:Preliminaries} covering the fundamental preliminaries of the domain, Section \ref{chp:Assumptions_for_Causality} discussing assumptions for causality, and Section \ref{chp:Related_Works} reviewing related algorithms and relevant surveys on causal analysis and identifying research gaps.
Section \ref{chp:research-methods} details the survey methodology, describing the data collection process in Section \ref{chp:Data_Collection}, and the analysis of the survey data in Section \ref{chp:Survey_Analysis}.
Section \ref{chp:Causal_Discovery_Algorithms} focuses on causal discovery algorithms, covering 6 categories of methodologies in Sections \ref{chp:Granger-Based_Method} to \ref{chp:Hybrid_Method}, respectively.
Section \ref{chp:Empirical_Study_Design} outlines the empirical study design, including investigated datasets in Section \ref{chp:Datasets}, evaluation metrics in Section \ref{chp:Evaluation_Metrics}, algorithms in Section \ref{chp:Algorithms}, and environment settings in Section \ref{chp:Environment_Settings}. 
Section \ref{chp:results} presents the results analysis, answering the research questions. 
Potential threats to validity are discussed in Section \ref{chp:Threats_to_Validity}.
Section \ref{chp:conclusion} concludes this study and presents future work directions.

\section{Background and Related Work}
\label{chp:literature-review}

Here we will analyze the preliminaries, related survey literature, and research gaps in causal discovery. 
The section aims to structure the knowledge body of this academic domain systematically.

\subsection{Preliminaries}
\label{chp:Preliminaries}

This section presents the fundamental definitions and corresponding notations associated with causal discovery. One needs to state that matrices are denoted by uppercase bold letters, whereas vectors are indicated by lowercase bold letters. Consider the dataset denoted by \bm{$X$}, which manifests as a $m\times n$ matrix. Here, \bm{$x^n$} designates the $nth$ variable, while \bm{$x$} embodies $m$ observations. This endeavour categorizes observational data into cross-sectional and time-series data, as defined below.

\textbf{Definition 1 (Cross-sectional Data):} Cross-sectional data is a set of observations collected from subjects at one time point.

Note that we mainly focus on a common type of cross-sectional data, namely independent and identically distributed (i.i.d.) data.

\textbf{Definition 2 (Time Series):} Time series is a sequence of data points arranged in temporal order. Given a time series \bm{$x^n$}, an observation at a specific temporal point $t$ is represented as $x^n_t$.

The concept of time lag is introduced to discern the demarcation between time-delay and instantaneous causality.

\textbf{Definition 3 (Time Lag):} Time Lag $\tau$ refers to the temporal interval between a cause and its effect.

In cases where $\tau>0$, it signifies the occurrence of the cause $\tau$ units of time prior to its effect. This phenomenon is referred to as ``time-delay causality''. Nonetheless, circumstances arising from sampling techniques or other factors might occasion an instance wherein $\tau=0$. In such scenarios, the causal latency is deemed insignificant for observation, and this relationship is classified as ``instantaneous causality''.

To further depict the causal interdependencies among variables within the dataset, it becomes imperative to introduce the notion of causal graphs.

\textbf{Definition 4 (Causal Graph):} The causal graph \bm{$G$} is composed of two subsets: a set of nodes \bm{$v$} and a set of edges \bm{$\epsilon$}. If variable \bm{$x^i$} is cause of variable \bm{$x^j$}, denoted as $\bm{x^i}\rightarrow \bm{x^j}$, this relationship manifests as an edge from node $i$ to node $j$ in Directed Acyclic Graphs (DAGs) \citep{pearl1985bayesian}.

However, when there are hidden variables in the dataset, Maximal Ancestral Graphs (MAGs) \citep{richardson2002ancestral} can represent causal relationships. The types of edges in MAGs are as follows:

$\bm{x^i}\rightarrow \bm{x^j}$: $\bm{x^i}$ causes $\bm{x^j}$;

$\bm{x^i}\leftrightarrow \bm{x^j}$: there is a hidden confounder between $\bm{x^i}$ and $\bm{x^j}$;

$\bm{x^i}- \bm{x^j}$: there is a hidden effect variable from both $\bm{x^i}$ and $\bm{x^j}$.

\begin{figure}[ht]
    \centering
    \includegraphics[scale=0.55]{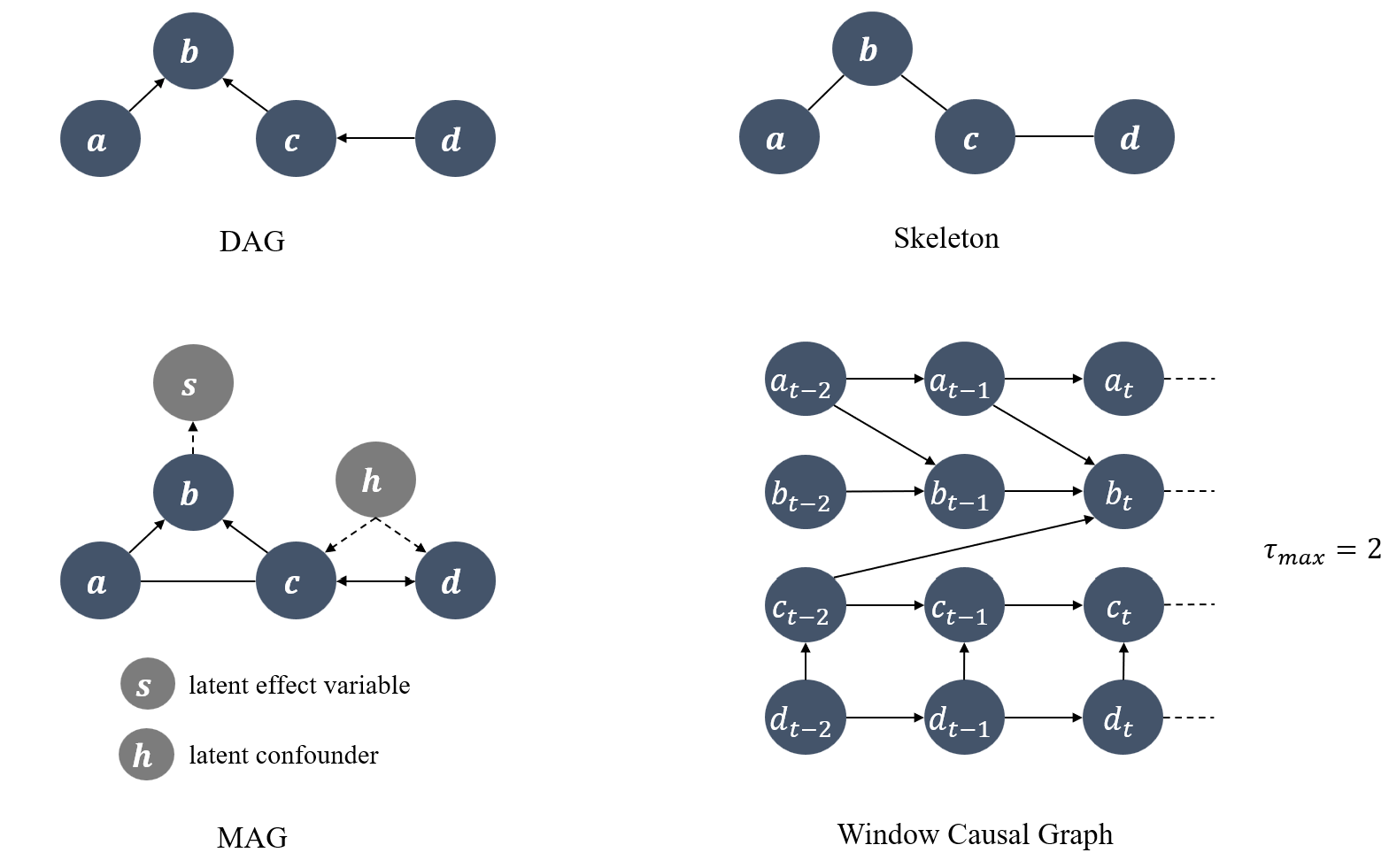}
    \caption{Causal Graphs}
    \label{fig:causal_graph}
     \end{figure}

If we do not consider causal directions, a skeleton graph can signify the causal relations between variables. There are only undirected edges in the skeleton graph that represent causal links.
For time series, a window graph is common for causal discovery, referring to the causal graph within the maximum time lag window \citep{assaad2023root}. Figure \ref{fig:causal_graph} shows examples of these causal graphs. 

Building upon the precedent definitions, we can elucidate the tasks of causal discovery. Given a dataset \bm{$X$}, the objectives of causal discovery are the deduction of the causal graph, the quantification of causal strength, and the determination of causal time lags (when the data is time-series). In essence, the goal is to reconstruct the causal mechanism interconnecting the variables.
Our focus is centred upon exploring causal graphs, driven by the objective of identifying causal relations within the variable set \bm{$v$}. To accomplish this, the outcomes of the causal discovery process are encoded using an adjacency matrix \bm{$A$}.

\textbf{Definition 5 (Adjacency Matrix):} The Adjacency Matrix of causal discovery constitutes a square matrix of dimensions $n\times n$. Specifically, the row vector is set to signify the cause, while the column vector signifies the effect. If the value located at position $(i, j)$ in the matrix is 1, it denotes the presence of a causal link from variable \bm{$x^i$} to \bm{$x^j$}.

\subsection{Assumptions for Causality} 
\label{chp:Assumptions_for_Causality}

When conducting causal discovery tasks, it is crucial to consider certain foundational assumptions. 
All the methods analyzed in this study are based on at least one of these assumptions.
These assumptions help in relating causality to probability densities \citep{spirtes2016causal}.

\subsubsection{The Causal Markov Assumption}

Spirtes and Zhang (\citeyear{spirtes2016causal}) argue that in a dataset, all features are independent of their non-effects (nondescendants in the causal graph) conditional on their direct causes (parents in the causal graph). 
However, it is important to note that the Causal Markov Assumption can be an oversimplification. 
It assumes that causality is the sole reason for associations between all features, which is not always true, as other kinds of associations exist as well.

For instance, consider the case of ice cream sales and drowning incidents.
According to the Causal Markov Assumption, if these two features were present in a dataset and shared an association, it should imply a causal link. 
However, we know this isn’t true. 
During summers, both ice cream sales and swimming activities increase. 
While there might be some conditional dependencies between the two due to common factors, confusing correlation with causality would lead to spurious results.

\subsubsection{The Causal Faithfulness Assumption}
    
The Causal Faithfulness Assumption asserts that all and only the true causal links between features are represented in the observed data. 
This means there are no latent or unobserved confounding variables impacting the observed features and their underlying causal relationships \citep{spirtes2016causal}.

For example, consider a dataset of patients in a lung cancer ward, with smoking habits as one of the features and lung cancer as the target variable. 
To determine whether smoking causes lung cancer, the causal faithfulness assumption holds if there are no unmeasured factors (like genetic vulnerability) linked to both smoking habits and lung cancer but are absent from the observed dataset. 
If the assumption does not hold, it indicates the presence of confounding factors. 
For instance, if the dataset lacks information about the patients' medical history or genetic predisposition, the observed causal link between smoking and lung cancer may be spurious. 
In such a scenario, smoking habits may appear to cause lung cancer in the data, but the true causal relationship is distorted by latent confounders.
    
\subsubsection{Markov Equivalence Classes (MEC)}
    
Markov Equivalence Classes (MEC) are another important concept in causal discovery.
MECs group statistically indistinguishable causal models that make the same predictions about the probability distributions of observed variables \citep{spirtes2016causal}.
In a given set of variables V, a Markov Equivalence Class represents an aggregation of DAGs that exhibit the same pattern of conditional independence associations among variables in V. 

For example, consider two DAGs, G1 and G2. G1 and G2 lie in the same Markov Equivalence Class if, for each pair of variables v1 and v2 in V, v1 and v2 are conditionally independent given all other variables v3 in V in G1 if and only if they are conditionally independent given v3 in G2. Simply put, for G1 and G2 to lie in the same Markov Equivalence Class, they must imply the same set of conditional independence statements among the variables.

To illustrate this further, consider two plausible causal models, M1 and M2, and three variables P, Q, and R:
    \begin{center}
    \[M1: P \rightarrow Q \rightarrow R\]
    \[M2: P \leftarrow Q \leftarrow R\]
    \end{center}
    
In M1, P causally influences Q, and Q causally influences R.
In M2, R causally influences Q, and Q causally influences P. 
Despite the different causal directions, M1 and M2 lie in the same Markov Equivalence Class because the observed data generated from these models will exhibit the same trends and conditional independence associations \citep{spirtes2016causal}.

\subsubsection{The Causal Sufficiency Assumption}

The Causal Sufficiency Assumption \citep{spirtes2001causation} states that the common causes between any two variables of the variable set \bm{$v$} are entirely contained within \bm{$v$} itself, thereby excluding the presence of latent confounders. This condition is considered a prerequisite for the efficacy of most causal discovery algorithms.

For example, consider one causal model M and three variables, P, L, and R:
    \begin{center}
    \[M: P \leftarrow L \rightarrow R\]
    \end{center}
    
If L is an unobserved variable, indicating that L is a hidden confounder of R and Q, then the model M does not satisfy the causal sufficiency assumption. Under this assumption, a directed edge in DAGs represents a causation from a cause to its effect.

\subsection{Related Works}
\label{chp:Related_Works}

\begin{table}[ht]
\centering
\caption{Timeline of Causal Discovery Algorithms}
\label{tab:timeline}
\scalebox{0.75}{
\begin{tabular}{@{\,}r <{\hskip 1.0pt} !{\foo}>{\raggedright\arraybackslash}p{18cm}}
\toprule
\addlinespace[1.0ex]
1969 & Pairwise Granger Causality \citep{granger1969} {\hskip 3.0pt}\sethlcolor{lightgray}\hl{\textit{one of the first statistical model of causal discovery}}\\
1982 & Multivariate Granger Causality (MVGC) \citep{geweke1982,chen2004,barrett2010multivariate}\\
2001 & PC \citep{spirtes2001causation,colombo2014order} {\hskip 3.0pt}\sethlcolor{lightgray}\hl{\textit{turn observations into causal knowledge}}\\
2002 & GES \citep{chickering2002,chickering2002optimal,chickering2020statistically}\\
2006 & ICALiNGAM \citep{shimizu2006} {\hskip 3.0pt}\sethlcolor{lightgray}\hl{\textit{based on structural equation model}}\\
2008 & ANM \citep{hoyer2008,hoyer2012causal}

Kernel Granger Causality (KGC) \citep{marinazzo2008kernel,kgc2021}

FCI \citep{zhang2008completeness,colombo2012learning}\\
2010 & tsFCI \citep{entner2010causal}

VARLiNGAM \citep{Hyvarinen2010}\\
2011 & IOTA \citep{hempel2011}

DirectLiNGAM \citep{Shimizu2011,hyvarinen2013pairwise}

ARMA-LiNGAM \citep{kawahara2011}\\
2012 & CCM \citep{sugihara2012,ye2015distinguishing} {\hskip 3.0pt}\sethlcolor{lightgray}\hl{\textit{dynamic systems}}

PNL \citep{zhang2012identifiability}

IGCI \citep{janzing2012}\\
2013 & ES \citep{10.5555/2591248.2591250}

TiMINO \citep{peters2013}\\
2014 & Copula Granger Causality (Copula GC) \citep{hu2014}

CMS \citep{ma2014}

PAI \citep{mccracken2014}\\
2015 & oCSE \citep{sun2015}\\
2017 & PSDR-TE \citep{mao2017}\\
2018 & NOTEARS \citep{zheng2018dags,zheng2020learning,fang2023low}

CGNN \citep{goudet2018}

SCDA \citep{Raghu2018}

RECI \citep{blobaum2018cause}\\
2019 & PCMCI \citep{runge2019,runge2020discovering}

TCDF \citep{nauta2019}

GraNDAG \citep{lachapelle2019gradient}

CDS \citep{fonollosa2019conditional}\\
2020 & CD-NOD \citep{JMLR:v21:19-232}

DYNOTEARS \citep{pamfil2020}

RCD \citep{Maeda2020RCDRC,maeda2022rcd}

GOLEM \citep{ng2020role}

NonSENS \citep{monti2020causal}\\
2021 & CAM-UV \citep{Maeda2021CausalAM}

DAG-GNN \citep{yu2019daggnn}

CORL \citep{wang2021ordering}

NGC \citep{Tank2021,wang2023detecting}\\
2022 & GRaSP \citep{lam2022greedy}

ACD \citep{lowe2022amortized}\\
2024 & NBCB, CBNB \citep{bystrova2024causaldiscoverytimeseries}
\end{tabular}
}
\end{table}

In recent years, the topic of causal discovery has garnered significant interest as researchers endeavor to uncover causal relationships from observational and experimental data.
To establish a solid foundation for understanding causal discovery, it is crucial to collect related works in this field.
Pearl's work \citep{pearl1985bayesian,pearl2000} on causal Bayesian networks and the introduction of causal graphical models significantly advanced the field's theoretical foundation. 
These seminal contributions serve as the basis for further research and methodologies.
Peter Spirtes introduced nonparametric Structural Causal Models (SCM) \citep{pearl2009causality} as a formal and intelligible language for articulating causal knowledge and explaining causal notions used in scientific discourse. 
These include concepts like randomization, intervention, direct and indirect effects, confounding, counterfactuals, and attribution. 
The structural language's algebraic component corresponds to the potential-outcome framework \citep{rubin1974estimating}, while its graphical component incorporates Wright's method of path diagrams.
The potential outcome framework, which focuses on estimating potential outcomes to calculate treatment effects, is particularly applicable to A/B tests. It performs effectively in causal inference, even when the complete causal graph is unknown \citep{aliprantis2015distinction}.
When combined, these components provide a robust approach for causal inference, addressing long-standing issues in empirical sciences, such as confounding control, policy evaluation, mediation analysis, and the algorithmization of counterfactuals.

Table \ref{tab:timeline} illustrates the timeline of the development of causal discovery algorithms.
As shown in Table \ref{tab:timeline}, Granger \citep{granger1969} proposed a statistical model to determine causal relationships between bivariables, which became one of the oldest mathematical models in the history of causal discovery.
Spirtes et al. introduced the assumptions and methods that laid the foundation for this field in their pioneering work \citep{spirtes2001causation}, attempting to transform observations in real-world into causal knowledge.
Spirtes and Glymour (\citeyear{spirtes2001causation}) developed the PC algorithm as a fundamental constraint-based algorithm. 
This algorithm starts with an undirected graph and recursively deletes edges based on conditional independence judgments.
Since 2006, Shimizu et al. (\citeyear{shimizu2006}) have designed a Linear Non-Gaussian Acyclic Model (LiNGAM) algorithm based on Structural Equation Model (SEM), which was developed into many variants to address non-linear relationships, time series, mixed data, latent confounders, and other data cases, forming a class of algorithm groups.
LiNGAM-related methods are widely applied across diverse fields, including neuroscience \citep{ji2024metacae,chiyohara2023proprioceptive}, economics \citep{jin2024contemporaneous}, epidemiology \citep{barrera2022link,garcia2020direction}, psychology \citep{mojtabai2024problematic,rosenstrom2023direction}, chemistry \citep{luo2024causal}, and others.
However, the above methods cannot handle nonseparable weakly connected dynamic systems. In response to this issue, Sugihara et al. (\citeyear{sugihara2012}) proposed a Convergent Cross Mapping (CCM) algorithm based on state space method under the assumption of nonlinear deterministic systems, which is suitable for dynamic research fields such as ecology.
In addition, for large sample datasets, causality algorithms combined with constantly evolving deep learning techniques greatly improve the accuracy of causal discovery and have become a popular research method.

In addition to the algorithms and methods mentioned above, we also need to investigate causal discovery from a more holistic perspective. Consequently, an exhaustive analysis must be conducted, combined with authoritative surveys over the past five years, to gain a comprehensive understanding of this field.

It is imperative to differentiate the objectives and conceptual frameworks associated with causal discovery and causal inference to further delve into these two notions, as elucidated by Guo et al. (\citeyear{Guo2020}). 
Guo et al. emphasised that causal inference involves tracing the causal path from cause to effect, aiming to understand the impact of manipulating specific variables on others. 
Within the realm of causal inference, Yao et al. (\citeyear{Yao2021}) have conducted an in-depth investigation into the concepts, methods, and applications of causal inference, contributing significantly to the field. 
Acknowledging the distinct nature of time series data, which differs from i.i.d. data, is essential. 
This distinction presents unique challenges and considerations in analysing causality.

Additionally, to supplement the understanding of these concepts, Nogueira et al. (\citeyear{Nogueira2022}) have analysed and compared causal discovery and inference using software tools, providing practical examples for testing. 
Their work contributed to the existing body of knowledge by exploring the practical application and evaluation of different approaches.

Focusing on time series, Moraffah et al. (\citeyear{Moraffah2021}) provided a comprehensive examination of causality for such data, offering insights into the generation of time series data, methodologies employed in the causal analysis, and evaluation metrics used to assess causal relationships. 
Notably, the study enumerated these evaluation metrics' specific attributes and characteristics, providing valuable information for researchers in selecting appropriate metrics for their analyses.
Regrettably, their research remained confined to theoretical realms and has yet to be realized through practical trial.

Another article on evaluation metrics was proposed by Cheng et al. (\citeyear{Cheng2022}). 
The research conducted thoroughly examines the evaluation methods employed in causal analysis. 
Their analysis referred to a wide range of considerations, including the availability and suitability of software packages, algorithms' effectiveness, and datasets' appropriateness for evaluating causal learning algorithms. 
By investigating these aspects, Cheng et al. provided researchers with valuable guidance for selecting appropriate evaluation methods in causal analysis studies.

Charles K. Assaad et al. (\citeyear{Assaad2022}) have made a notable contribution to the field of time series causal discovery, and their work serves as a pivotal reference for the research project at hand. 
In the theory field, they presented a comprehensive framework comprising seven distinct categories for analyzing causal relationships. 
In the empirical field, they employed ten algorithms to assess these methods' performance across different causal structures.

More recently, Runge, J. et al. (\citeyear{runge2023}) comprehensively summarized methods of causal discovery and proposed a Question-Assumptions-Data (QAD) template, embedding causal discovery into Pearl's causal ladder. They also designed a method selector to match the optimal algorithm to different graph assumptions. However, they did not further discuss parametric assumptions about datasets through experimentation.

Hasan, U. et al. (\citeyear{hasan2023survey}) summarized causal discovery methods for i.i.d. data and time series, and collected source code of relevant algorithms. They also tested and compared the performance of 9 algorithms of i.i.d. data  and 7 algorithms of time series on benchmark datasets. However, they did not further test and analyze the influence of data assumptions on algorithms performance, such as dependency functions and noise distributions.

\begin{table}[ht]
\centering
\scalebox{0.75}{
\renewcommand{\arraystretch}{1.85}
\setlength{\tabcolsep}{12pt}
\begin{tabular}{cccc}
\hline
\textbf{Question} & \textbf{Data} & \textbf{Method} & \textbf{Evaluation} \\
\hline
\multirow{2}{*}{Causal Direction} & Pairwise & Granger-Based &  \\
 &  & Condition Independent-Based & Classification-Based Measures \\
\multirow{3}{*}{Causal Graph} & I.I.D. Data & State space Dynamic-Based & \\
 &  & Structural Equation Modeling-Based &  \\
 & Time Series & Deep Learning-Based & Graph Distance-Based Measures \\
 &  & Hybrid Methods & \\
\hline
\end{tabular}
}
\caption{Question-Data-Method-Evaluation Template}
\label{tab:QDME}
\end{table}

    
Drawing upon the analysis mentioned above, we have amalgamated and organized the principal research directions, which are visually represented in Table \ref{tab:QDME}. 
By employing the Question-Data-Method-Evaluation (QDME) template, the review aims to provide a clear and structured overview of the diverse areas and subtopics within the field of causal discovery research, enhancing the organization and coherence of the work.

Based on the aforementioned papers, while the algorithms and evaluations of causal discovery have become relatively comprehensive, several unresolved challenges persist, highlighting gaps in the existing body of knowledge as below.
\begin{enumerate}
    \item Many articles employ outdated taxonomies and lack updates on the latest algorithms.
    \item Most surveys emphasize theoretical analyses, often neglecting the systematic experiments necessary for quantitative assessment.
    \item Although some articles have conducted experimental comparisons of algorithms, these studies primarily consider causal structures as their experimental factors, overlooking the characteristics of the data.
\end{enumerate}

\section{Survey Methodology}
\label{chp:research-methods}

This section will introduce how to collect relevant research resources, including literature, codes, metrics, and datasets. 
Moreover, the last section briefly explains the analytical technologies we used.

\begin{figure}[htbp]
    \centering
    \includegraphics[scale=0.77]{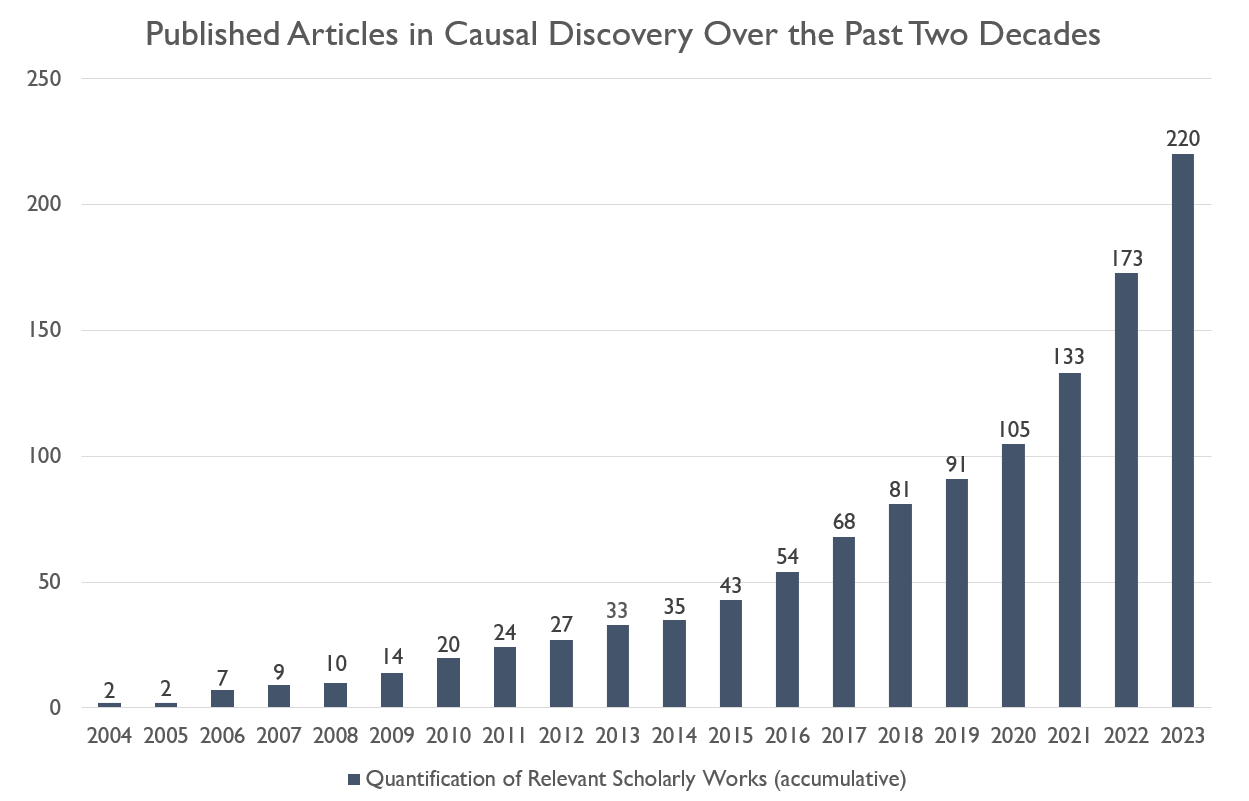}
    \caption{Published articles in ``Causal Discovery'' over the past two decades (accumulative)}
    \label{published_articles}
    \end{figure}

A quantitative research approach is employed to gather and analyze papers from databases systematically. 
This approach facilitates measuring and exploring trends, methods, datasets, and evaluation in the research domain.
Google Scholar was chosen as the literature database due to its extensive coverage of scholarly articles. 
Information acquisition was initiated by employing a keyword search approach. 
Primarily, Figure \ref{published_articles} demonstrates the research trend of causal discovery, underpinned by the number of articles published during the preceding two decades.

Figure \ref{published_articles} illustrates an increasing annual trend in research endeavours about causal discovery. 
This discernible growth can be attributed partly to the robust advancement of AI technologies in contemporary times, which has laid the groundwork for algorithmic developments facilitated by enhanced data processing capabilities.

Furthermore, for preparing the programming underpinnings, our endeavour extends more than literature collection to encompass acquiring requisite algorithmic source code.
To this end, we searched for the source code associated with each algorithm on the GitHub platform, with the integration and revision of original algorithm codes as deemed necessary.

To facilitate a comprehensive and impartial investigative analysis, this study employs both \textbf{\textit{comparative analysis}} and \textbf{\textit{case study}} methodologies.
Specifically, a series of comparative experiments involving diverse algorithms on distinct artificial datasets are conducted.
By analyzing the magnitudes of evaluation metrics, a quantitative assessment of algorithmic performance is executed, yielding overarching insights.
Ultimately, to affirm the robustness and effectiveness of our findings, the conclusions are corroborated through case studies such as real datasets.

\subsection{Data Collection}
\label{chp:Data_Collection}

Here we executed a keyword-based search focusing on aspects within the tree graph above. 
The subsequent table illustrates the collection result derived from the published paper from 2004 to 2023, as sourced from Google Scholar.

To prevent repetitive retrieval and collection of papers, ``Causal Discovery” is used as the basic keyword, with previously retrieved terms excluded when searching for new supplementary terms.
It should be noted that we also searched for keywords related to ``causal inference'' to ensure comprehensive collection of papers within cross fields.
We have identified 220 articles with the highest correlation from over 3000 related articles and have summarized the datasets, algorithms, and evaluation metrics for causal discovery.
The terms ``Hits,” ``Title,” and ``Body” in Table \ref{tab:Paper_collection_result} refer to the number of papers returned by the search, the relevance of the titles to the desired content, and the number of papers that remain after title and body framing, respectively.

\begin{table}[ht]
  \centering
  \caption{Paper Collection Result}
  \scalebox{0.85}{
    \label{tab:Paper_collection_result}
    \begin{tabular}{lccr}
      \toprule
      \textbf{Keywords} & \textbf{Hits} & \textbf{Title} &\textbf{Body}\\
      \midrule
      \multirow{6}{*} 
      {}``Causal Discovery'' + [Survey $|$ Overview $|$ Review] & 980 & 15 & 14\\
    ``Causal Discovery'' + [Bench $|$ Benchmark] & 347 & 7 & 7\\
    ``Causal Discovery'' + Dataset & 619 & 16 & 16\\
    ``Causal Discovery'' + [Evaluation $|$ Comparison] & 240 & 18 & 16\\
    ``Causal Discovery'' + [Method $|$ Algorithm $|$ Approach] & 410 & 138 & 137\\
    ``Causal Inference'' + [Time Series $|$ Cross-sectional $|$ I.I.D.]& 995 & 89 & 30\\
      \midrule
      \multirow{1}{*}
      Overall & - & 283 & 220\\
            \bottomrule
    \end{tabular}
  }
\end{table}

A selection of widely employed datasets for causal discovery has been identified by synthesising diverse review articles. 
These datasets encompass both real-world instances and artificial constructs. 
To streamline ensuing experimental processes, a classification framework has been devised to categorise these datasets into four types of causality relationships.
Additionally, the sources of these datasets are also categorized into real-world and synthetic datasets. 
Specifically, CausalWorld \citep{ahmed2020}, SynTReN \citep{van2006}, LUCAS \citep{guyon2011causality}, ALARM \citep{lauritzen1988local} and ASIA \citep{lauritzen1988local} are synthetic datasets, while Tubingen \citep{mooij2016}, ADNI \citep{petersen2010}, AntiCD3/CD28 \citep{sachs2005}, Abalone \citep{asuncion2007}, fMRI \citep{smith2011}, Causality 4 Climate \citep{runge2020}, Traffic Prediction \citep{pan2018}, OHDNOAA \citep{jangyodsuk2014}, Temperature Ozone \citep{gong2017}, Sachs \citep{sachs2005} and CHILD \citep{spiegelhalter1993bayesian} are real-world datasets.
This categorisation is illustrated in Table \ref{tab:Datasets_collection_result}.

\begin{table}[ht]
  \centering
  \caption{Overview of Datasets Utilised in the Reviewed Papers on Causality Research, Categorised by Causality Type, Year, Application Area, and Source.}
  \scalebox{0.7}{
    \label{tab:Datasets_collection_result}
    \begin{tabular}{llccr}
      \toprule
      \textbf{Causality Type} & \textbf{Dataset} & \textbf{Year} & \textbf{Area} & \textbf{Source} \\
      \midrule
      \multirow{4}{*}{Pairwise} 
      & Cause Effect Pairs Challenge \citep{cause-effect-pairs} & 2013 & Various & Real-world\\
      & Tubingen \citep{mooij2016} & 2016 & Various & Real-world \\
      & CE-Gauss \citep{mooij2016} & 2016 & - & Synthetic\\
      & CE-Multi$|$Net \citep{goudet2018} & 2018 & - & Synthetic\\
      \midrule
      \multirow{3}{*}{Instantaneous} 
      & SynTReN \citep{van2006} & 2006 & Biology & Synthetic\\
      & FLUXNET \citep{pastorello2020} & 2020 & Biogeoscience & Synthetic\\
      & causaLens \citep{dgp_causal_discovery_time_series} & 2021 & Various & Synthetic\\
      \midrule
      \multirow{5}{*}{Time-delay} 
       & fMRI \citep{smith2011} & 2011 & Neuroscience & Real-world\\
       & FinanceCPT \citep{kleinberg2013causality} & 2012 & Economics & Synthetic\\
      & OHDNOAA \citep{jangyodsuk2014} & 2014 & Hydrologic & Real-world\\
      & Traffic Prediction \citep{pan2018} & 2018 & Traffic & Real-world\\
      & Causality 4 Climate \citep{runge2020} & 2020 & Climate & Real-world \\
       \midrule
      \multirow{7}{*}{I.I.D. Data} 
      & Sachs \citep{sachs2005}  & 2005 & Biology & Real-world \\
      & LUCAS \citep{guyon2011causality} & 2011 & Medical & Synthetic\\
      & ALARM \citep{beinlich1989alarm} & 1989 & Belief Networks & Semi-synthetic\\
      & CHILD \citep{spiegelhalter1993bayesian} & 1993 & Medical & Real-world\\
      & ASIA \citep{lauritzen1988local} & 1988 & Medical & Synthetic\\
      & Auto-mpg \citep{driessens2005combining} & 2005 & Engineering & Real-world\\
      \bottomrule
    \end{tabular}
  }
\end{table}

Likewise, we have identified performance metrics for assessing causal discovery techniques. 
These metrics are divided into two overarching families: graph-based metrics \citep{peters2015structural} and classification-based metrics, as expounded in Table \ref{tab:Metrics_collection_result}. 
It is pertinent to highlight that nearly all metric computations necessitate the availability of both estimated DAGs and ground truth DAGs. 
Therefore, the prudent selection of datasets with well-established ground truth becomes imperative.

\begin{table}[h]
  \centering
  \caption{List of Metrics Used in Reviewed Papers.}
  \scalebox{0.75}{
    \label{tab:Metrics_collection_result}
    \begin{tabular}{lcp{6cm}}
      \toprule
      \textbf{Measure Type} & \textbf{Metric} & \textbf{Notions} \\
      \midrule
      \multirow{3}{*}{Graph distance-based measure} 
      & Structural Hamming Distance (SHD) & Calculate the difference between two (binary) adjacency matrices: each edge that is missing or not in the target graph is counted as an error. \\
      \cline{2-3}
      & Frobenius Norm & Compare the similarity between real matrices and estimation matrices. \\
      \cline{2-3}
      & Structural Intervention Distance (SID) & Estimate the count of erroneously deduced intervention distributions. \\
      \midrule
      \multirow{7}{*}{Classification-based measure} 
      & Precision & The quotient of true positives (TP) divided by the sum of TP and false positives (FP). \\
      \cline{2-3}
      & Recall & The proportion of TP in relation to the summation of TP and false negatives (FN). \\
      \cline{2-3}
      & F1 Score & The harmonic mean of precision and recall of the learned structure as compared to true causal structure. \\
      \cline{2-3}
      & FPR & The ratio of the edges that are present in the predicted graph but not present in the ground-truth graph. \\
      \cline{2-3}
      & TPR & The ratio of the common edges between the ground-truth and predicted causal graphs over the number of edges in ground-truth graph. \\
      \cline{2-3}
      & MSE & The sum of square of difference between the predicted and the ground-truth causal graphs divided by the total number of nodes. \\
      \cline{2-3}
      & Area under ROC Curve (AUROC) & Area under ROC curve is the area under the curve of recall versus FPR at different thresholds. \\
      \bottomrule
    \end{tabular}
  }
\end{table}
Additionally, we collected several well-established packages for causal discovery algorithms, including \textit{bnlearn, pcalg, Tetrad, Causal Discovery Toolbox (CDT), CausalNex, gCastle, Tigramite, and causal-learn}.
\textit{bnlearn} \citep{scutari2009learning} is an R package designed for Bayesian network learning and inference. It offers an open-source implementation of various structure learning algorithms, including constraint-based, score-based, and hybrid methods.
Another R package, \textit{pcalg} \citep{kalisch2012causal}, integrates graphical models and causal inference techniques. It provides implementations for several widely-used causal discovery algorithms, including PC, FCI, RFCI, GES, GIES, SIMY, ARGES, and LiNGAM.
\textit{Tetrad} \citep{ramsey2018tetrad} is a Java package designed for generating and simulating data, estimating parameters, testing hypotheses, predicting outcomes, and searching causal models.
\textit{CDT} \citep{kalainathan2019} primarily focuses on discovering causal relationships from observational data, ranging from determining pairwise causal directions to full graph modeling.
\textit{CausalNex} \citep{Beaumont_CausalNex_2021} is a Python library that integrates machine learning and domain expertise for causal inference using Bayesian networks. It enables users to discover structural relationships within data, analyze complex distributions, and evaluate the effects of potential interventions.
\textit{gCastle} \citep{zhang2021gcastlepythontoolboxcausal} is a causal structure learning toolchain developed by Huawei Noah's Ark Laboratory, offering a Python library for mainstream algorithms and emerging gradient-based approaches.
\textit{Tigramite} \citep{tigramite2023} is a Python package designed for time series analysis based on the PCMCI framework. It reconstructs graphical models (conditional independence graphs) from discrete or continuous time series data and generates high-quality graphical representations.
\textit{causal-learn} \citep{zheng2024causal} is a Python library built upon the Java-based \textit{Tetrad} causal discovery platform. The library provides modular code, enabling researchers to implement and extend their own algorithms efficiently.

\subsection{Experimental Data Collection}
\label{chp:Survey_Analysis}

Utilizing a comparative analysis framework, we systematically process the performance metrics. We selected a reference algorithm exhibiting superior performance to enhance result lucidity and then calculated residuals for other algorithms compared with the reference. The overall distribution of these residuals is visually represented through the violin plots.

Nevertheless, two specific algorithms may manifest insignificant disparities between their metric values. Hence, we introduce a significance assessment mechanism to pursue a more methodical treatment of data relationships. Given the constraint that each data size consists of merely five datasets, we opt for the non-parametric Mann-Whitney U test \citep{mann1947test}. This approach circumvents the necessity to assume normality in data distribution. Significance is determined based on a p-value threshold of 0.05; when under this threshold, it signals a notable divergence in the performance of the two algorithms.

In addition, a ranking table is formulated to delineate the evaluation outcomes for enhancing the discernibility of inter-algorithm performance disparities. This table is ordered in a descending manner upon the average values. Meanwhile, we calculate the standard deviation for each algorithm's metric values; reduced standard deviation signifies less dependence of algorithmic performance on data size, which means enhanced stability of algorithms.

\section{Causal Discovery Algorithms}
\label{chp:Causal_Discovery_Algorithms}

\begin{figure}[htbp]
\hypersetup{hidelinks}  
\centering
\begin{forest}
for tree={
    grow'=east,
    s sep=5pt, 
    l sep=0.8cm, 
    anchor=base,
    edge path={
        \noexpand\path[\forestoption{edge}]
        (!u.parent anchor) -- 
        ($( !u.east) + (15pt,0)$) |-  
        (.child anchor)\forestoption{edge label};
    },
    scale=0.28, 
    font=\huge, 
}
[Taxonomy for\\Causal\\Discovery, align=center
    [Granger-Based Methods, align=center
        [\underline{Multivariate Granger Analysis Method} \citep{arize1993}]
        [\underline{Extended Granger Causality} \citep{chen2004}]
        [\underline{Kernel Granger Method} \citep{liao2009}]
        [\underline{Copula Granger Method} \citep{hu2014}]
    ]
    [Condition\\Independence-Based\\Methods, align=center
        [Information Theoretic-Based Approach,  [\underline{oCSE} \citep{sun2015}]
        ]     
        [Causal Network-Based\\Approach, align=center, s sep=2pt, l=0.9cm
        [Constraint-Based,
            [Peter-Clark (PC) \citep{kalisch2007}]
            [CD-NOD \citep{JMLR:v21:19-232}]
            [\underline{PCMCI} \citep{runge2019}]  
            [Fast Causal Inference (FCI) \citep{entner2010causal}][\underline{tsFCI}\citep{entner2010causal}]
        ]
        [Score-Based,
            [Greedy Equivalence Search (GES) \citep{chickering2002}]
            [ES \citep{10.5555/2591248.2591250}]
            [GRaSP \citep{lam2022greedy}]
            [\underline{DYNOTEARS} \citep{pamfil2020}] 
        ]
        ]
    ]
        [State Space\\Dynamic-Based\\Methods, align=center,
        [\underline{CCM} \citep{sugihara2012}]
        [\underline{Cross Map Smoothness (CMS)} \citep{ma2014}]
        [\underline{Inner Composition Alignment (IOTA)} \citep{hempel2011}]
        [\underline{Pairwise Asymmetric Inference (PAI)} \citep{mccracken2014}]
    ]
    [Structural Equation\\Model-Based\\Methods, align=center
        [LiNGAM-Based,
            [ICA-LiNGAM \citep{shimizu2006}]
            [Direct LiNGAM \citep{Shimizu2011}]
            [\underline{VARLiNGAM} \citep{Hyvarinen2010}]
            [RCD \citep{Maeda2020RCDRC}]
            [CAM-UV \citep{Maeda2021CausalAM}]
        ]
        [Additive Noise Models (ANM) \citep{hoyer2008},
            [\underline{TiMINo} \citep{peters2013}]
        ]
        [PNL \citep{zhang2012identifiability}]
        [DAGs with NO TEARS \citep{zheng2018dags}]
        [GOLEM \citep{ng2020role}]
    ]
    [Deep Learning-Based\\Methods, align=center
        [Causal Generative Neural Networks (CGNN) \citep{goudet2018}]
        [DAG-GNN \citep{yu2019daggnn}]
        [\underline{Temporal Causal Discovery Framework (TCDF)} \citep{nauta2019}]
        [\underline{Amortized Causal Discovery (ACD)} \citep{lowe2022amortized}]
        [{Ordering-Based Causal Discovery with Reinforcement Learning} \citep{wang2021ordering}]
        [GraNDAG \citep{lachapelle2019gradient}]
    ]
    [Hybrid Methods, align=center
        [\underline{ARMA-LiNGAM} \citep{kawahara2011}]
        [Scalable Causation Discovery Algorithm (SCDA) \citep{Raghu2018}]
        [\underline{PSDR-TE} \citep{mao2017}]
        [Information Geometric Causal Inference (IGCI) \citep{janzing2012}]
        [Non-linear SEM Estimation using Non-Stationarity (NonSENS) \citep{monti2020causal}]
        [\underline{Neural Granger Causality (NGC)} \citep{Tank2021}]
        [\underline{NBCB} and \underline{CBNB} \citep{bystrova2024causaldiscoverytimeseries}]
    ]
]
\end{forest}
\caption{Taxonomy for Causal Discovery (The underlined algorithms can only be applicable to Time-Series)}
\label{fig:mytree1}
\end{figure}
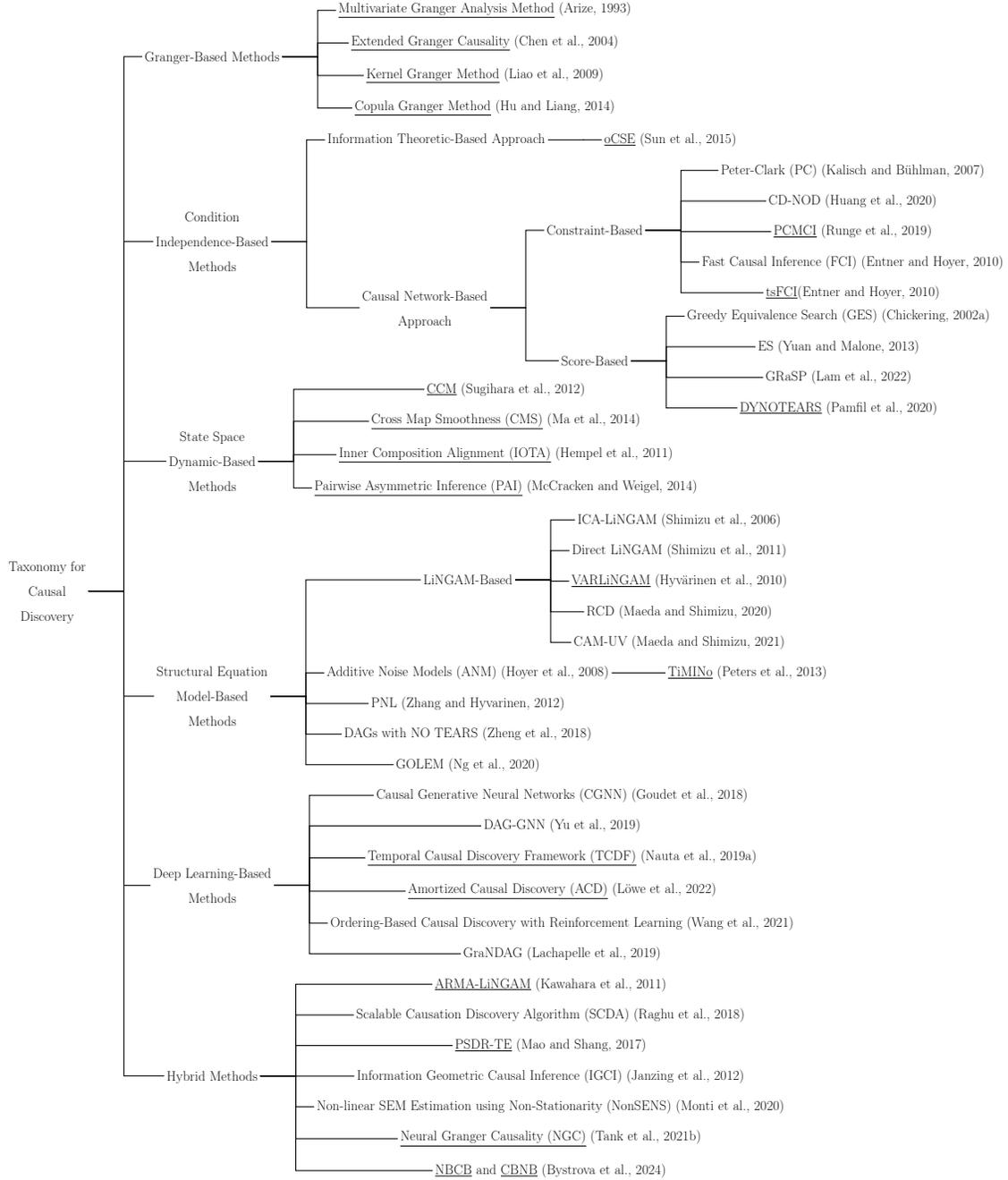

Various research articles present diverse taxonomy for causal discovery, yet a universally accepted classification structure does not currently exist. Specifically, in recent years, the rapid advancements in this domain have resulted in the incompleteness and obsolescence of numerous surveys. It is essential to collect and analyse the taxonomy methodologies proposed in papers critically to establish a systematic categorisation of existing methods. Simultaneously, the devised structure should encompass as many algorithms as possible. Thus, this project diligently compiles and summarises the prevailing methods, effectively partitioning causal discovery into six fundamental categories, as shown in Figure \ref{fig:mytree1}: Granger-Based, Conditional Independence-Based, State Space Dynamics-Based, Structural Equation Modelling-Based, Deep Learning-Based, and Hybrid Method.

\subsection{Granger Based Method}
\label{chp:Granger-Based_Method}

Granger causality (GC) is one of the pioneering measurement methods for analysing time series data. 
Over several decades, undergoing refinement and evolution, it still maintains an irreplaceable position in the contemporary landscape of causal discovery.
The core premise of Granger causality postulates that future events do not affect the present or past, while past events potentially impact both the present and the future.
When the historical information of variables \bm{$x$} and \bm{$y$} are included, leading to better predictions for variable $y$ than predictions based solely on the information of \bm{$y$}, it signifies that variable \bm{$x$} is considered the Granger cause of variable \bm{$y$}.
In mathematical notation, the given statement can be expressed as follows \citep{mccracken2016}:

\begin{equation}
P(y_{n+1}\in \bm{A}\mid \bm{\Omega_n}) \ne P(y_{n+1}\in \bm{A}\mid \bm{\Omega_n} - \bm{x})
\label{equation: equation_for_Granger_Based_Method}
\end{equation}

Equation \ref{equation: equation_for_Granger_Based_Method} illustrates \bm{$x$} Granger causes \bm{$y$}, wherein the variable \bm{$x$} and \bm{$y$} represents two discrete time series, and the subscript $n$ corresponds to the time point $t$. 
The all-encompassing set of information available at all points $t \leq n$ is symbolically denoted as \bm{$\Omega_n$}.
To ascertain the Granger relationship based on the aforementioned formula, the Vector Autoregressive (VAR) model \citep{lutkepohl2005new} stands as the prevailing technique, built upon the premise of data stationarity and equipped to forecast variable values.

Our attention is directed towards several primary algorithms based on GC.
Arize et al. (\citeyear{arize1993}) put forth the Multivariate Granger Causality (MVGC) analysis method to overcome the limitations of Pairwise Granger Causality (PWGC), which can only deal with bivariate data. 
However, ensuring linearity in real-world datasets can present a significant challenge. 
To address this issue, Chen et al. (\citeyear{chen2004}) introduced an approach known as Extended Granger Causality (EGC), specifically designed to handle nonlinear data. 
An alternative model catering to nonlinear time series is the Kernel Granger Causality (KGC) method \citep{marinazzo2008kernel,liao2009,marinazzo2011nonlinear}, showcasing notable attributes such as high accuracy and flexibility.
When confronted with continuous time series data, Hu et al. (\citeyear{hu2014}) proposed the Copula Granger method, which is capable of uncovering nonlinear and higher-order causal relationships \citep{kim2020copula,jang2022vine}.

Although Granger-based causality method has a long history of development, it still struggles to handle complex causal relationships.
Its main drawback is the inability to identify latent confounders and instantaneous causal effects.
Therefore, the Granger-based method is often combined with other methods to achieve mutual development.

\subsection{Condition Independence Based Method}
\label{chp:Condition_Independence_Based_Method}

The conditional independence-based method exhibits a close association with probability.
By quantifying the mutual information \citep{runge2018conditional} between variables, this approach enables the determination of causal relations and causal strength. 
A fundamental concept in this context is the transfer entropy \citep{schreiber2000}, which is defined as follows:

\begin{equation}
T_{\bm{x}\rightarrow \bm{y}} = \sum P(y_{n+1}, y_n^{(k)}, x_n^{(l)}) \log \frac{P(y_{n+1} \mid y_n^{(k)}, x_n^{(l)})}{P(y_{n+1} \mid y_n^{(k)})}
\end{equation}

In contrast to Shannon entropy, the transfer entropy is computed by the Kullback entropy \citep{kullback1997}. In this equation, $x_n$ represents the value of variable \bm{$x$} at the $n$th time point, and likewise for the variable \bm{$y$}, with the superscript indicating the time delay length. When $T_{\bm{x}\rightarrow \bm{y}} - T_{\bm{y}\rightarrow \bm{x}} > 0$, it can be inferred that variable \bm{$x$} is the cause of variable \bm{$y$}; conversely, variable \bm{$y$} is the cause of variable \bm{$x$}.

Causal discovery algorithms based on conditional independence can be categorized into two distinct groups. 
The first category is the information-theoretic-based approach, with Optimal Causation Entropy (oCSE) \citep{sun2014identifying,sun2015} being a representative algorithm. 
oCSE is a two-step discovery algorithm explicitly designed for short time series data. 
The second approach is causal network-based, where the optimal causal graph is determined through statistical testing.
The Peter-Clark (PC) algorithm \citep{kalisch2007} has gained widespread adoption and has proven effective in analyzing high-dimensional time series using causal graphs.
Recognizing the potential interference of latent confounders in causal detection, corresponding approaches have been developed.
A notable example is the Fast Causal Inference (FCI) algorithm \citep{zhang2008completeness,entner2010causal,spirtes2013causal}, a classical method that explicitly accounts for unobserved confounders.
To further enhance control over false positive rates, Runge et al. introduced the PCMCI method and its variants \citep{runge2019,runge2020discovering,gerhardus2020high} by incorporating the MCI test into the PC algorithm. PCMCI is an improvement of PC in time series, which can detect contemporaneous and time-delay effects.

Conditional Distribution Similarity Statistic (CDS) \citep{fonollosa2019conditional} was proposed to detect the causal direction of bivariates. This method measures the statistical characteristics of the joint distribution of marginal variance data after conditioning the bins. This algorithm has been proven to be robust as it has a high AUC in ChaLearn causal pair challenges.

Constraint-based causal Discovery from heterogeneous/NOnstationary Data (CD-NOD) \citep{JMLR:v21:19-232} is another framework designed to discover causal relationships in data where generating processes change over time or across domains. 
It detects changing local mechanisms, recovers causal structures, and estimates the driving force behind nonstationarity. 
This nonparametric method leverages data heterogeneity and connects nonstationarity with soft interventions, demonstrating efficacy on synthetic and real-world datasets like task-fMRI and stock market data.

Here we introduce classic score-based approaches. Compared to the PC algorithm, the Greedy Equivalence Search (GES) algorithm, proposed by Chickering et al. (\citeyear{chickering2002}), shows enhanced robustness when dealing with nonstationary data. 
Greedy Relaxations of Sparsest Permutation (GRaSP) \citep{lam2022greedy,andrews2023fast} is designed to efficiently identify DAGs representing causal structures from observational data. 
It builds on permutation-based reasoning and introduces a novel operation called ``tuck" to relax the assumptions required by previous methods like Triangle Sparsest Permutation (TSP) and Edge Sparsest Permutation (ESP). 
GRaSP consists of three tiers: GRaSP0, GRaSP1, and GRaSP2, each progressively weakening the assumptions and increasing the ability to recover sparser permutations.
GRaSP2, the most relaxed form, outperforms several state-of-the-art algorithms in simulations, demonstrating scalability and accuracy for dense graphs and those with over 100 variables. 
To address causal discovery problems in temporal series, Pamfil et al. (\citeyear{pamfil2020}) introduced DYNOTEARS, a novel Bayesian network learning algorithm that utilizes score constraints to ascertain the edges within the causal structure graph. 
DYNOTEARS is capable of effectively handling both instantaneous and delayed causality, making it a versatile tool for causal inference in time series data.

Compared to Granger-based methods, conditional independence-based methods can handle more complex data scenarios, such as high-dimensional data, instantaneous causality, and latent variables. However, these methods generally require the faithfulness assumption and have limitations in determining causal direction, i.e., some causal links remain unoriented. Despite these drawbacks, they are well-suited for identifying causal skeleton graphs.

\subsection{State Space Dynamic Based Method}
\label{chp:State_Space_Dynamic_Based_Method}

The state space dynamics-based method can be regarded as a complementary approach to address a category of data not encompassed by GC. This method investigates the causality of variables within weakly coupled dynamic systems, significantly enhancing the causal discovery capability in ecological, dynamics, and other relevant domains. This method draws inspiration from the Takens theorem \citep{takens1981} and computes the bidirectional cross-correlation between two variables to establish a cross-mapping. To be specific, variable \bm{$x$} causes \bm{$y$} when $C_{\bm{xy}}>C_{\bm{yx}}$ is satisfied, wherein $C_{\bm{xy}}$ and $C_{\bm{yx}}$ represent the Convergent Cross-Mapping (CCM) correlations from \bm{$x$} to \bm{$y$} and from \bm{$y$} to \bm{$x$}, respectively. The calculation formula for CCM correlation is as follows, where $\rho$ means the Pearson correlation coefficient.

\begin{equation}
C_{\bm{xy}} = [\rho(\bm{y}, \bm{y}|\tilde{\bm{x}})]^2
\end{equation}

\begin{figure}[htbp]
    \centering
    \subfigure[$M_x$ to $M_y$ cross mapping]{
    \label{fig.sub.1}
    \includegraphics[width=0.4\textwidth]{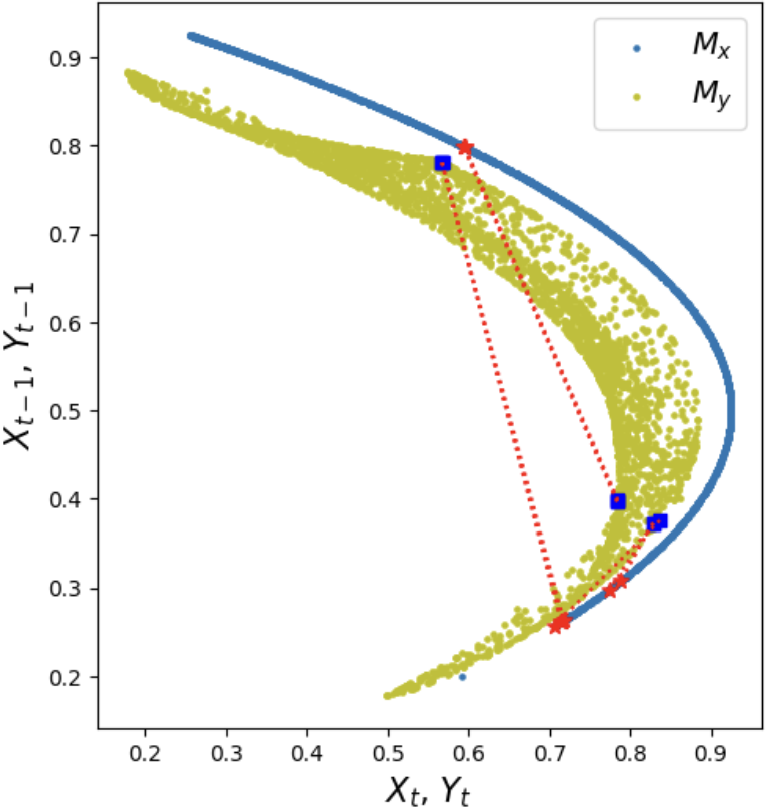}
    }
     \subfigure[$M_y$ to $M_x$ cross mapping]{
    \label{fig.sub.2}
    \includegraphics[width=0.4\textwidth]{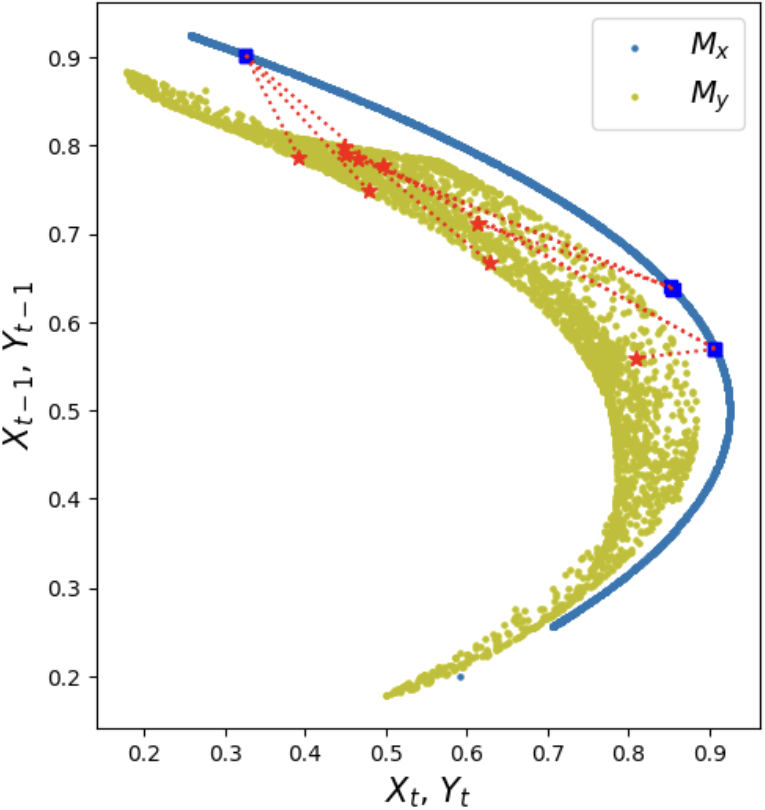}
    }
    \caption{Cross mapping manifold (lag = 1) \citep{ccm2021}}
    \label{fig:Cross_mapping_manifold}
    \end{figure}

We assume a classic example to explain cross-mapping further, using variable \bm{$x$} to construct shadow manifold $M_{\bm{x}}$, and $\bm{y}$ to construct $M_{\bm{y}}$. If \bm{$x$} leads to \bm{$y$}, then using the neighbouring points of a certain point in $M_{\bm{y}}$ should be able to identify better the neighbouring points of the corresponding point in $M_{\bm{x}}$. Supposing a delay of 1 and the shadow manifold graphs from two directions are shown in Figure \ref{fig:Cross_mapping_manifold}, where \bm{$x$} causes \bm{$y$}.

Sugihara et al. (\citeyear{sugihara2012}) introduced the concept of CCM to infer causality. This method proves advantageous for non-separable and weakly connected dynamic systems. An essential feature of CCM is convergence, which means that the longer the time series used (with a larger sample size), the smaller the estimated error of the obtained cross-mapping. To overcome this limitation that necessitates long time series data, Ma et al. (\citeyear{ma2014}) developed Cross Map Smoothness (CMS), specifically designed to handle varying data sizes, particularly short time series. Additionally, Inner Composition Alignment (IOTA) \citep{hempel2011,wang2014segmented} offers an alternative technique for short-time series analysis. McCracken et al. (\citeyear{mccracken2014}) proposed Pairwise Asymmetric Inference (PAI) as an exploratory tool for analysing high-dimensional dynamic systems. PAI aids in uncovering causal relationships within complex systems by evaluating asymmetries in pairwise interactions.

The most significant advantage of state space dynamic based methods is their efficacy in deterministic systems, making them the preferred choice for such specialized data scenarios. However, their primary disadvantage lies in their limited applicability and difficulty in handling datasets with time-variant noises.

\subsection{Structural Equation Model Based Method}
\label{chp:Structural_Equation_Modelling_Based_Method}

The three methods above are well-suited for identifying time delay causality but may not necessarily be adept at detecting instantaneous causality. On the other hand, the Structural Equation Model (SEM) based method represents a significant advancement and revolution in the realm of instantaneous causal discovery. This method ascertains the edges of DAG by establishing a structural equation to solve the coefficient matrix. The most basic form of the structural equation \citep{shimizu2006} is as follows:

\begin{equation}
\mathbf{X} = \mathbf{BX}+\mathbf{E}
\end{equation}

The matrix $\mathbf{B}$ is referred to as the coefficient matrix, with its row and column representing the two dimensions of cause and effect. Upon determining $\mathbf{B}$, the causal relation can be discerned. Additionally, the matrix $\mathbf{E}$ represents the noise matrix in the model, usually non-Gaussian noise.

The Linear Non-Gaussian Acyclic Model (LiNGAM) provides a directed acyclic graph that reveals instantaneous causal relations between variables. 
LiNGAM assumes linear and non-Gaussian independent noise about the data generation method of the system, solving these using Independent Component Analysis (ICA) \citep{lee1998independent,naik2011overview}.
Since ICA algorithms typically employ FastICA and gradient-based algorithms, they may converge to local rather than global optima. 
To address these issues, Shimizu et al. (\citeyear{Shimizu2011}) proposed the Direct LiNGAM algorithm. 
Compared to ICA-LiNGAM, Direct LiNGAM produces more stable and reliable results, though it has some drawbacks.
One drawback is its slower computational efficiency compared to ICA-LiNGAM. 
Additionally, its assumptions are relatively strict; in real-world scenarios, data generation mechanisms are often nonlinear and do not conform to its assumptions.
Hyvärinen et al. (\citeyear{hyvarinen2013pairwise}) proposed a measure for determining pairwise causal direction based on likelihood ratio tests. The objective of the study was to develop a method based on SEM that performs more effectively on real-world data with limited sample sizes. In simulations using brain imaging (fMRI) data, this method demonstrated significantly better performance compared to ICA-LiNGAM and other approaches.

To address the gap in the LiNGAM algorithm family concerning delayed causality, Hyvärinen et al. (\citeyear{Hyvarinen2010}) designed the VARLiNGAM algorithm. 
VARLiNGAM operates in two steps: first, it predicts time lag effects using a Vector Autoregression (VAR) model; second, it estimates instantaneous causality by applying LiNGAM.

Building upon the insights from LiNGAM, Hoyer et al. (\citeyear{hoyer2008}) introduced Additive Noise Models (ANM) to detect nonlinear time series, emphasizing that nonlinearities offer valuable identification power. 
One shortcoming of this algorithm is its high computational cost, as it involves determining the direction between pairs of variables one-on-one, making the algorithm pairwise causality.

The Post-Nonlinear (PNL) causal model \citep{zhang2012identifiability,zhang2015estimation,uemura2022multivariate} addresses the complexities of nonlinear effects, inner noise, and measurement distortions in observed variables for causal discovery. 
Representing each variable as a function of its direct causes, an independent disturbance, and a post-nonlinear distortion, PNL can distinguish between causes and effects, especially in non-Gaussian scenarios. 
The model's identifiability has been extensively studied, revealing that it can generally identify causal directions except in specific conditions. 
Empirical results demonstrate its efficacy in various real-world data sets, making it a robust tool for causal inference in complex systems.

Peters et al. (\citeyear{peters2013}) proposed Time Series Models with Independent Noise (TiMINo) to capture both lagged and instantaneous effects.
This model is based on nonlinear independence tests and can perform well even when the dataset does not satisfy the causal sufficiency assumption.

DAGs with NO TEARS \citep{zheng2018dags} is a causal discovery method that uses continuous optimization schemes to learn the structure of Directed Acyclic Graphs (DAGs) from observational data.
The name ``NO TEARS” stands for ``Nonlinear Optimal Transformations for Efficient and Accurate Recovery of Structure,” emphasizing its focus on tackling issues related to learning nonlinear causal relationships. 
The algorithm transforms the data to make it linear or Gaussian and learns a Structural Equation Model (SEM) linking features in a causal graph.
It optimizes the SEM with a focus on fit and sparsity using methods like gradient descent or ADMM. 
Soft thresholding promotes sparsity by zeroing out edges.
The process iterates until convergence criteria are met, then returns a DAG with NO TEARS, representing causal relationships.
However, Kaiser and Sipos (\citeyear{kaiser2021unsuitability}) analyzed the lack of scale invariance in the NOTEARS algorithm and concluded that this limitation makes NOTEARS unsuitable for identifying true causal relationships from data.

Regression Error based Causal Inference (RECI) \citep{blobaum2018cause} addresses the problem of inferring the causal relationship between two variables by comparing the least-squares errors of predictions in both possible causal directions. Bl{\"o}baum emphasize that RECI can have a significantly lower computational cost than ANM, while delivering comparable or even superior results. Additionally, RECI is straightforward to implement and apply.

GOLEM, introduced by Ng et al. (\citeyear{ng2020role}), is a continuous likelihood-based method for causal discovery. 
It uses a score-based approach with soft sparsity and DAG constraints to maximize the data probability of a linear Gaussian model. 
GOLEM employs two objective functions to account for noise variances and uses an l1 penalty for complexity. 
It formulates an unconstrained optimization problem, ensuring the graph remains a DAG under reasonable assumptions.
The algorithm utilizes gradient-based optimization methods, with Adam optimizer and GPU acceleration. 
A post-processing step removes low-weight edges to enhance performance. 
GOLEM effectively restores DAG structures while managing soft constraints.

Repetitive Causal Discovery (RCD) \citep{Maeda2020RCDRC,maeda2022rcd} is a method for identifying causal structures in data affected by latent confounders. 
It repeatedly infers causal directions between small sets of observed variables, determining if relationships are influenced by latent confounders. 
The resulting causal graph uses bi-directed arrows to indicate variables sharing the same latent confounders and directed arrows for causal directions between variables not affected by the same latent confounder.
Experimental validation with simulated and real-world data shows that RCD effectively identifies latent confounders and causal directions.

Causal Additive Models with Unobserved Variables (CAM-UV) \citep{Maeda2021CausalAM} handle causal discovery in the presence of unobserved variables, particularly for nonlinear causal relationships. 
This model extends causal additive models by accounting for both unobserved common causes and intermediate variables. 
CAM-UV identifies all theoretically possible causal relationships without bias from unobserved variables, avoiding incorrect inferences. 
Empirical results from artificial and simulated fMRI data confirm CAM-UV's effectiveness in inferring causal structures despite the presence of unobserved variables.

The advantage of Structural Equation Model based methods is their applicability to a wide range of data types. LiNGAM and its variant algorithms can handle i.i.d. data, time series, instantaneous causality, hidden confounders, mixed data, and other data types without requiring faithfulness assumptions. However, these methods predominantly rely on linear relationships, and only a few algorithms are capable of handling nonlinear relations.

\subsection{Deep Learning Based Method}
\label{chp:Deep_Learning_Based_Method}

Deep learning-based methods have emerged as powerful tools in causal discovery, closely connected with machine learning. 
These methods offer significant technical advantages, particularly in processing vast amounts of data. 
Notably, deep learning-based causal algorithms can better infer hidden variables using network information.

For instance, Goudet et al. (\citeyear{goudet2018}) designed Causal Generative Neural Networks (CGNN) to address the challenges posed by latent variables in causal analysis.
CGNN is an algorithm that infers the optimal causal direction on a causal skeleton diagram, which belongs to pairwise causality.
Through testing on both artificial and real-world datasets, CGNN has demonstrated advanced performance in handling potential confounders.

DAG-GNN, developed by Yu et al. (\citeyear{yu2019daggnn}), combines Graph Neural Networks (GNNs) with a score-based approach to learn DAGs from data.
It uses GNNs for node embeddings to model feature dependencies and detect causal relationships.
The method starts by embedding features using GNNs, then defines a score to evaluate the causal structure. 
It formulates an optimization problem to maximize this score using gradient-based techniques like stochastic gradient descent or Adam. 
Causal constraints ensure acyclicity. 
The data is split into training and validation sets to optimize the score and train the model. 
After achieving optimal performance on the validation set, the algorithm returns a DAG representing causal relationships.

Another noteworthy algorithm is the Temporal Causal Discovery Framework (TCDF) \citep{nauta2019}, which effectively handles both latent and instantaneous causal effects.
TCDF adopts an attention mechanism in convolutional neural networks.
The attention coefficients of different variables, learned by the network, can be interpreted as the degree of correlation between variables.
If the attention coefficient is below a certain threshold, it indicates no causal relationship between the two variables.

GraNDAG \citep{lachapelle2019gradient} is a novel score-based approach for learning DAGs from observational data.
It adapts a recent continuous constrained optimization formulation to accommodate nonlinear relationships between variables using neural networks.
This method effectively models complex interactions and avoids the combinatorial nature of the problem. 
By comparing GraNDAG to existing continuous optimization methods and nonlinear greedy search methods, it has been demonstrated that GraNDAG outperforms current continuous methods on most tasks and remains competitive with existing greedy search methods on important causal inference metrics.

Ordering-Based Causal Discovery with Reinforcement Learning (CORL), developed by Wang et al. (\citeyear{wang2021ordering}), combines ordering-based causal discovery with reinforcement learning techniques \citep{zhu2019causal} to learn causal relationships by generating and refining an ordering of variables.
The algorithm treats the problem as a sequential decision-making task, where a reinforcement learning agent arranges variables to approximate true causal relationships. 
A reward function provides feedback, incentivizing correct orderings and penalizing incorrect ones.
The task is formulated as a Markov Decision Process (MDP), with states representing the current ordering and actions selecting the next variable. 
The agent is trained using reinforcement learning algorithms like Q-learning or Proximal Policy Optimization to optimize long-term rewards. 
Balancing exploration and exploitation is crucial. A post-processing step, such as local search, refines the ordering to improve accuracy. 
The algorithm ultimately returns an optimal causal graph representing the relationships between features.

More recently, Löwe et al. (\citeyear{lowe2022amortized}) introduced the Amortized Causal Discovery (ACD) algorithm for time series, which is effective with small data sample sizes and demonstrates efficacy in dynamic systems. 
This model utilizes shared information between dynamic system variables to identify confounders in additive noise datasets.

The advantage of deep learning based methods lies in their capacity to handle datasets with large sample sizes and numerous variables. However, their drawbacks include long running times, low efficiency, and suboptimal performance on short time series.

\subsection{Hybrid Method}
\label{chp:Hybrid_Method}

Hybrid methods combine two or more algorithms to complement and optimize each other, enhancing the ability to discover causality. 
These methods leverage the strengths of different approaches to address their individual limitations and improve overall performance.

One illustrative hybrid approach is the Autoregressive Moving Average - Linear Non-Gaussian Acyclic Model (ARMA-LiNGAM) \citep{kawahara2011}, which combines Granger Causality (GC) and Structural Equation Models (SEM). 
This composite method resolves both instantaneous and delayed causality by leveraging the attributes of LiNGAM and ARMA models. ARMA-LiNGAM's integration allows for a more comprehensive analysis of time series data, accommodating both immediate and lagged effects.
Incorporating deep learning techniques with GC models, Neural Granger Causality (NGC) \citep{Tank2021,wang2023detecting} stands out. 
NGC utilizes the Causal Multilayer Perceptron (CMLP) model to train data, thereby enhancing the accuracy of causal inference tasks.
By combining the predictive power of neural networks with the interpretability of Granger causality, NGC offers a robust framework for identifying causal relationships in complex datasets.

Janzing et al. (\citeyear{janzing2012}) introduced Information Geometric Causal Inference (IGCI) to address the nonlinear challenges encountered by the Additive Noise Model (ANM) algorithm. IGCI enhances causal inference by incorporating information entropy, providing a more effective method for dealing with nonlinear data structures. This approach allows for better differentiation between cause and effect in scenarios where traditional linear models fall short.

Mao et al. (\citeyear{mao2017}) extended the application of the Convergent Cross Mapping (CCM) algorithm from bivariate to multivariate analysis by incorporating transfer entropy. This integration, named Phase State Delay Reconstruction - Transfer Entropy (PSDR-TE), effectively addresses the limitation of the CCM algorithm, which was originally designed for detecting bivariate relationships. PSDR-TE expands the capability of causal discovery to more complex multivariate systems, improving the detection of causal links across multiple variables.

Raghu et al. (\citeyear{Raghu2018}) proposed the Scalable Causation Discovery Algorithm (SCDA), which combines structural equation model-based and conditional independence-based methods. SCDA provides a solution for mixed data containing both continuous and discrete sequences. By integrating these two methodologies, SCDA can handle a broader range of data types and improve the robustness of causal inference in heterogeneous datasets.

Monti et al. (\citeyear{monti2020causal}) proposed an algorithm called Non-linear SEM Estimation using Non-Stationarity (NonSENS) for bivariate data. This approach employs a deep learning-based method, Time Contrastive Learning (TCL), within a SEM framework, allowing for arbitrary instantaneous nonlinear relationships without assuming additive noise. Notably, the direction of effect in arbitrary nonlinear SEMs is proved to be identifiable \citep{hyvarinen2024identifiability}.

Bystrova et al. (\citeyear{bystrova2024causaldiscoverytimeseries}) developed two novel algorithms, NBCB and CBNB, which integrate SEM with a constraint-based approach to infer causal graphs from time series. Both approaches are capable of inferring various types of causal graphs including instantaneous and lagged relationships. These algorithms exhibit effectiveness and robustness across both synthetic and real-world datasets.

In summary, hybrid methods in causal discovery leverage the strengths of multiple algorithms to address their respective weaknesses. By combining techniques such as Granger causality, structural equation models, neural networks, and information entropy, these hybrid approaches offer powerful tools for uncovering causal relationships in diverse and complex datasets.

\section{Empirical Study Design}
\label{chp:Empirical_Study_Design}

The aim of this section is to design experiments to answer the three RQs in section \ref{chp:introduction}. RQ1 is \textit{comparison of algorithm performance}, RQ 2 is \textit{real-world applicability}, and RQ3 is \textit{generalization to unknown datasets}.

In light of this, the experimental framework is structured across four distinct phases. 
The inaugural phase involves conducting a comparative assessment of algorithms applied to synthesised datasets with specific features while concurrently evaluating a range of performance metrics. 
Subsequently, the second phase encompasses presenting and analysing outcomes derived from the initial stage to extract meaningful insights. 
Transitioning to the third phase, real-world datasets are engaged for testing, utilising the insights garnered in the preceding phase to ascertain the optimal algorithm. 
This stage aims to verify whether the test results are consistent with the predicted optimal algorithm. 
The fourth and final phase entails deploying diverse data processing and testing methodologies to ascertain the metadata of the time series datasets. 
This, in turn, facilitates the extrapolation of insights from the second phase to previously unexplored datasets.

\subsection{Datasets}
\label{chp:Datasets}

\begin{figure}[ht]
    \centering
    \includegraphics[scale=0.165]{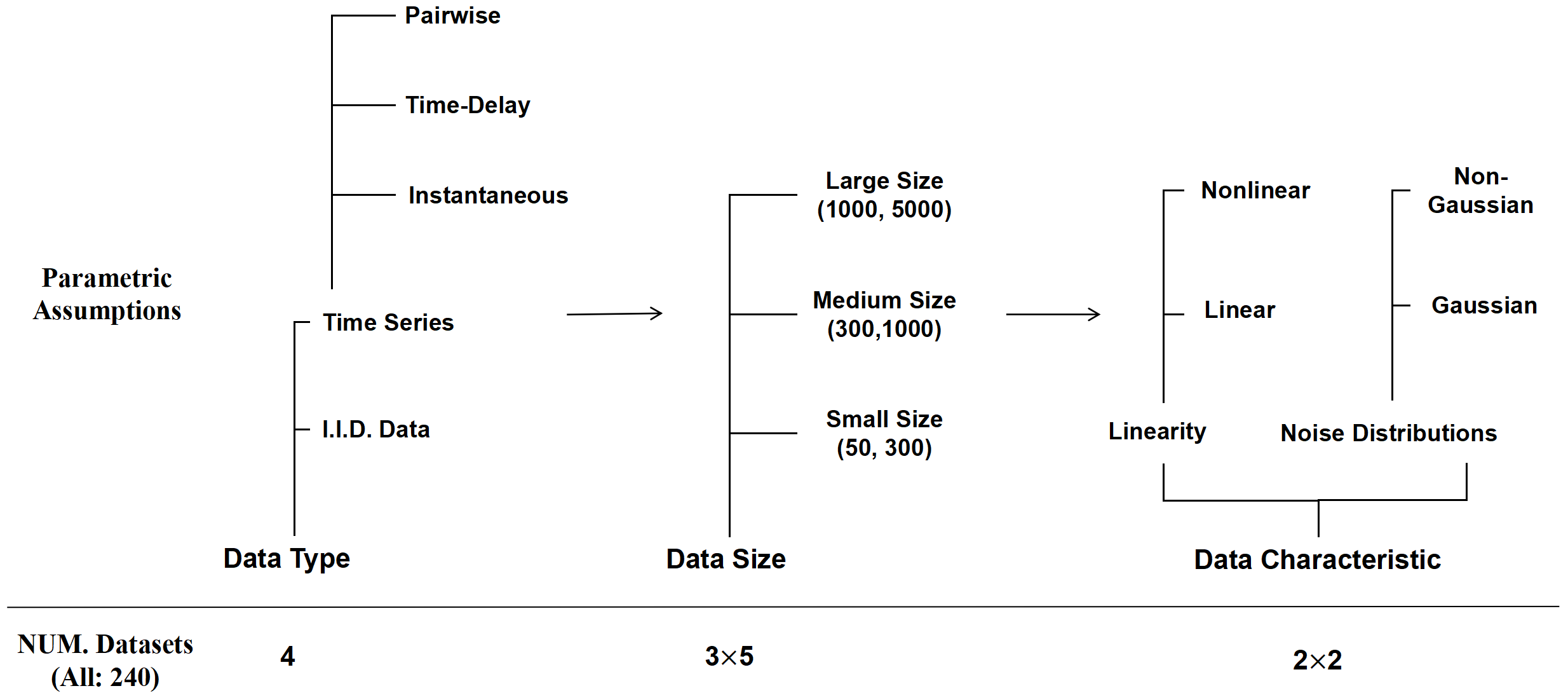}
    \caption{Data Generator Structure}
    \label{fig:data-stru}
    \end{figure}

This experimental inquiry necessitates two dataset categories: synthetic datasets designed to explore underlying patterns and real-world datasets serving the purpose of validation. The data generation structure of the artificial dataset is shown in Figure \ref{fig:data-stru}.

Given the categorization of testing algorithms into three classes, a concomitant preparation of diverse composite datasets becomes imperative. 
The first category entails a data generator tasked with producing causal pairs, whereby the data generation tool within the causal discovery toolbox \citep{kalainathan2019} was employed.
The second and third categories encompass instantaneous and time-delay causality, for which the data generator in Tigramite framework, posited by Runge et al. (\citeyear{runge2023}), was harnessed for data synthesis. 
In the fourth category, characterized by i.i.d. data, we adopted the data generation model established within the gCastle package \citep{zhang2021gcastlepythontoolboxcausal}.

Subsequently, we considered the dataset sizes for both time series and i.i.d. data.
For time series data, the experimental framework established distinct time series lengths ranging from 50 to 300 time points for small-scale datasets, 300 to 1000 for medium-scale datasets, and 1000 to 3000 for large-scale datasets. 
For i.i.d. data, the sizes were similarly categorized into small (50, 100, 150, 200, 250), medium (300, 440, 580, 720, 860), and large (1000, 1400, 1800, 2200, 2600).

The design of dataset attributes necessitates attention to causal relationships and noise distribution types. 
Causal relationships within this context are bifurcated into linear and nonlinear relationships. Linear relationships are generated through polynomial operations on dataset variables, while nonlinear relationships involve trigonometric operations.
The noise distribution types encompass Gaussian noise with parameters defined by a mean of 0 and a standard deviation of 1, and uniform noise spanning the interval (0,1).

Building upon the framework above, five datasets are generated for each data size, each subjected ten times to mitigate runtime-induced biases, resulting in a total of 180 distinct datasets for comprehensive algorithm evaluation. 
All generated datasets adhere to the causal sufficiency assumption and possess stability, prerequisites fundamental to the operation of the algorithmic processes.

Regarding authentic datasets, scarcity in datasets featuring established ground truth, particularly within the domain of time series, is evident. Consequently, this study incorporates two verifiable datasets. 
The first is the “Tuebingen” dataset, which comprises 100 real cause-effect pairs. 
Additionally, the “fMRI” dataset, which aims to investigate the Blood Oxygen Level Dependent (BOLD) signal across 28 distinct intrinsic brain networks, is also integrated into the study.

\subsection{Evaluation Metrics}
\label{chp:Evaluation_Metrics}

In devising this project's evaluation criteria, we tried to encompass multiple aspects as comprehensively as possible. Recognizing that a singular indicator might introduce bias, we select five distinct indicators for our assessment.

The initial metric utilized is the F1 score, which serves as a prevalent evaluation criterion in causal discovery due to its capacity to assess the model's overall performance. The F1 score is computed as follows: 

\begin{equation}
F1 = \frac{2\times precision \times recall}{precision + recall}
\end{equation}

To augment the evaluation of the model's robustness, specifically its capacity to discern TPR and FPR, we employed the Area Under the Receiver Operating Characteristic (AUROC) as a metric, which can be obtained by computing the area under the ROC curve.

The first two indicators signify that higher numerical values correspond to better performance, with a preference to assess the accurate inference. 
For assessing the extent of false causality present in the model, we employ the False Positive Rate (FPR), computed as follows: 

\begin{equation}
FPR = \sum_i \frac{e_i}{E-E_{GT}}, e_i\in E_M/E_{GT}
\end{equation}

Subsequently, we incorporate a graph-based causal discovery metric called Structural Hamming Distance (SHD). 
This metric directly reveals the number of incorrectly inferred edges by comparing the differences between the ground truth and the estimated causal graph. 

Lastly, we meticulously recorded the run time of each algorithm, given that the time cost was considered a significant aspect of this experiment. 
This particular indicator will serve as a crucial criterion for assessing the efficiency of the algorithms.

\subsection{Experimental Algorithms}
\label{chp:Algorithms}

By collecting resources on GitHub, we introduce in this section the algorithms selected for the experiment, as well as their source code and packages.
We mainly use four packages, Causal Discovery Toolbox (CDT) (\citeyear{cdt2019}), gCastle (\citeyear{gCastle2021}), Causal Discovery for Time Series (CD\_TS)(\citeyear{cdts2022}), causal-learn (\citeyear{causallearn2024}) to implement the testing algorithms.

As shown in Table \ref{tab:Algorithm_parameter_configuration_table}, we select MVGC and PWGC, implemented in CD\_TS(\citeyear{cdts2022}), from the repertoire of GC methods, as the subject of experimentation.
It is worth noting that MVGC exclusively addresses delayed causality while PWGC addresses pairwise causality, so their applicability has certain limitations.

In the context of the conditional independence-based method, Runge et al. (\citeyear{tigramite2023}) developed PCMCI and its variant algorithms for time series.
The oCSE, tsFCI and DYNOTEARS algorithms are also included in the experiment for time series.
Considering i.i.d. data, we chooose PC, FCI, GES, GRaSP, ES and CDS. These methods are chosen for their robust performance in different scenarios, providing a comprehensive evaluation of algorithmic capabilities.

Most algorithms rooted in the state space dynamic-based approach primarily concentrate on resolving the directionality of causal pairs.
Among these, we examine the classical CCM and PAI algorithms \citep{ccm2021}, categorizing them as instances of pairwise causality. 
Additionally, IGCI and ANM are incorporated as pairwise causality algorithms.

As for the structural equation model-based approach, our emphasis lies on exploring a variant of the LiNGAM algorithm \citep{lingam2023} tailored for i.i.d. data and time series data, referred to as ICALiNGAM, DirectLiNGAM, RCD, and VARLiNGAM. 
The TiMINO algorithm is excluded from consideration due to its documented inferiority compared to PCMCI and TCDF \citep{nauta2019}.

TCDF algorithm(\citeyear{tcdf2019}) frequently emerges in diverse surveys and holds a pivotal position within the field, thus making it a suitable choice as a representative algorithm for deep learning-based methods.
For i.i.d. data, we include three algorithms from gCastle package: DAG-GNN, CORL and GraNDAG.
Besides, we do not include CGNN in this experiment due to its prolonged running time, which could impede the efficiency of the overall analysis.

Within the hybrid method, we opt to employ both NeuralGC \citep{ngc2021} and IGCI. 
NeuralGC possesses the ability to handle both delay and instantaneous causal relationships, while IGCI is specialized in resolving causal pairs.


In order to present a lucid exposition of the algorithms employed in this experiment, we marked the testing algorithms with * in Table \ref{tab:Algorithm_parameter_configuration_table}.
It is essential to highlight that, in pursuit of optimal performance for each algorithm, a series of individual tests were conducted. 
According to these tests, specific default values of several algorithms were modified to ensure the validation of the results.

\begin{table}[ht]
  \centering
  \caption{Causal Discovery Algorithms}
  \scalebox{0.55}{
    \label{tab:Algorithm_parameter_configuration_table}
    \begin{tabular}{lcccccccr}
      \toprule
      \textbf{Method} & \textbf{Algorithm} & \textbf{Time-series} & \textbf{i.i.d.} & \textbf{Faithfulness} &\textbf{CMC} & \textbf{Sufficiency} & \textbf{Linear} & \textbf{Software}\\
      \midrule
      \multirow{5}{*}{Granger-based} 
      & PWGC* & \checkmark & & & & \checkmark & \checkmark & \href{https://github.com/ckassaad/causal_discovery_for_time_series.git}{CD\_TS}\\
      & MVGC* & \checkmark & & & & \checkmark & \checkmark & \href{https://github.com/ckassaad/causal_discovery_for_time_series.git}{CD\_TS}\\
      & EGC & \checkmark & & & & \checkmark &\\
      & KGC & \checkmark & & & & \checkmark & & \href{https://github.com/danielemarinazzo/KernelGrangerCausality.git}{KernelGrangerCausality}\\
      & CopulaGC & \checkmark & & & & \checkmark & & \href{https://github.com/hualouliang/CopulaGrangerCausality_ContinuousData.git}{CopulaGrangerCausality}\\
      \midrule
      \multirow{10}{*}{Condition independence-based} 
      & oCSE* & \checkmark & & \checkmark & \checkmark & \checkmark & & \href{https://github.com/ckassaad/causal_discovery_for_time_series.git}{CD\_TS}\\
      & PC* & & \checkmark & \checkmark & & \checkmark & & \href{https://github.com/FenTechSolutions/CausalDiscoveryToolbox.git}{CDT}; \href{https://github.com/huawei-noah/trustworthyAI.git}{gCastle}; \href{https://github.com/py-why/causal-learn.git}{causal-learn}\\
      & PCMCI* & \checkmark & & \checkmark & \checkmark & \checkmark & &\href{https://github.com/jakobrunge/tigramite.git}{tigramite}; \href{https://github.com/ckassaad/causal_discovery_for_time_series.git}{CD\_TS}\\
      & FCI* & & \checkmark & \checkmark & \checkmark & & &\href{https://github.com/py-why/causal-learn.git}{causal-learn}\\
      & tsFCI* & \checkmark & & \checkmark & \checkmark & & & \href{https://github.com/ckassaad/causal_discovery_for_time_series.git}{CD\_TS}\\
      & GES* & & \checkmark & \checkmark& \checkmark& & & \href{https://github.com/FenTechSolutions/CausalDiscoveryToolbox.git}{CDT}; \href{https://github.com/huawei-noah/trustworthyAI.git}{gCastle}; \href{https://github.com/py-why/causal-learn.git}{causal-learn}\\
      & GRaSP* & & \checkmark & & & & \checkmark & \href{https://github.com/py-why/causal-learn.git}{causal-learn}\\
      & ES* & & \checkmark & & & & &\href{https://github.com/py-why/causal-learn.git}{causal-learn}\\
      & DYNOTEARS* & \checkmark & & & & \checkmark &
      \checkmark & \href{https://github.com/ckassaad/causal_discovery_for_time_series.git}{CD\_TS}\\
      & CDS* & & \checkmark & & & \checkmark & & \href{https://github.com/FenTechSolutions/CausalDiscoveryToolbox.git}{CDT}\\
      \midrule    
      \multirow{4}{*}{State space dynamics-based} 
      & CCM* & \checkmark & & & & \checkmark & & \href{https://github.com/PrinceJavier/causal_ccm.git}{causal\_ccm}\\
      & CMS & \checkmark & & & & \checkmark &\\
      & IOTA & \checkmark & & & & \checkmark &\\
      & PAI* & \checkmark & & & & \checkmark & & \href{https://github.com/PrinceJavier/causal_ccm.git}{causal\_ccm}\\
      \midrule    
      \multirow{11}{*}{Structural equation model-based} 
      & ICALiNGAM* & & \checkmark & & \checkmark & \checkmark & \checkmark & \href{https://github.com/cdt15/lingam.git}{LiNGAM}; \href{https://github.com/huawei-noah/trustworthyAI.git}{gCastle}; \href{https://github.com/py-why/causal-learn.git}{causal-learn}\\
      & DirectLiNGAM* & & \checkmark & & \checkmark & \checkmark & \checkmark & \href{https://github.com/cdt15/lingam.git}{LiNGAM}; \href{https://github.com/huawei-noah/trustworthyAI.git}{gCastle}; \href{https://github.com/py-why/causal-learn.git}{causal-learn}\\
      & VARLiNGAM* & \checkmark & & & \checkmark & \checkmark & \checkmark & \href{https://github.com/cdt15/lingam.git}{LiNGAM}; \href{https://github.com/ckassaad/causal_discovery_for_time_series.git}{CD\_TS}; \href{https://github.com/py-why/causal-learn.git}{causal-learn}\\
      & RECI* & &\checkmark & & & \checkmark & & \href{https://github.com/FenTechSolutions/CausalDiscoveryToolbox.git}{CDT}\\
      & RCD* & & \checkmark & & & & & \href{https://github.com/cdt15/lingam.git}{LiNGAM}; \href{https://github.com/py-why/causal-learn.git}{causal-learn}\\
      & CAM-UV & & \checkmark & & & & & \href{https://github.com/cdt15/lingam.git}{LiNGAM}; \href{https://github.com/py-why/causal-learn.git}{causal-learn}\\
      & ANM* & \checkmark & \checkmark & & \checkmark & \checkmark & & \href{https://github.com/FenTechSolutions/CausalDiscoveryToolbox.git}{CDT}; \href{https://github.com/huawei-noah/trustworthyAI.git}{gCastle}; \href{https://github.com/py-why/causal-learn.git}{causal-learn}\\
      & TiMINO* & \checkmark & & & \checkmark & \checkmark & & \href{https://github.com/ckassaad/causal_discovery_for_time_series.git}{CD\_TS}\\
      & PNL & & \checkmark & & \checkmark & \checkmark & & \href{https://github.com/huawei-noah/trustworthyAI.git}{gCastle}; \href{https://github.com/py-why/causal-learn.git}{causal-learn}\\
      & NOTEARS* & & \checkmark & & \checkmark & \checkmark & \checkmark & \href{https://github.com/huawei-noah/trustworthyAI.git}{gCastle}\\
      & GOLEM* & & \checkmark & & \checkmark & \checkmark & \checkmark & \href{https://github.com/huawei-noah/trustworthyAI.git}{gCastle}\\
       \midrule    
      \multirow{4}{*}{Deep learning-based} 
      & CGNN & \checkmark & \checkmark & \checkmark & \checkmark & & & \href{https://github.com/FenTechSolutions/CausalDiscoveryToolbox.git}{CDT}\\
      & DAG-GNN* & & \checkmark & \checkmark & & \checkmark & & \href{https://github.com/huawei-noah/trustworthyAI.git}{gCastle}\\
      & TCDF* & \checkmark & & & & & & \href{https://github.com/M-Nauta/TCDF.git}{TCDF}; \href{https://github.com/ckassaad/causal_discovery_for_time_series.git}{CD\_TS}\\
      & ACD & \checkmark & & & & & & \href{https://github.com/loeweX/AmortizedCausalDiscovery.git}{ACD}\\
      & CORL* & & \checkmark & \checkmark & \checkmark & & & \href{https://github.com/huawei-noah/trustworthyAI.git}{gCastle}\\
      & GraNDAG* & & \checkmark & \checkmark & \checkmark & \checkmark & & \href{https://github.com/huawei-noah/trustworthyAI.git}{gCastle}\\
      \midrule
      \multirow{4}{*}{Hybrid} 
      & ARMA-LiNGAM & \checkmark & & & & \checkmark & \checkmark & \href{https://github.com/cdt15/lingam.git}{LiNGAM}\\
      & NeuralGC* & \checkmark & & & & & & \href{https://github.com/iancovert/Neural-GC.git}{Neural-GC}\\
      & IGCI* & \checkmark & \checkmark & \checkmark & \checkmark & \checkmark & & \href{https://github.com/FenTechSolutions/CausalDiscoveryToolbox.git}{CDT}\\
      & SCDA & & \checkmark & & & & \checkmark\\
      & PSDR-TE & \checkmark & & & & &\\      
      
      \bottomrule
    \end{tabular}
  }
\end{table}

\subsection{Environment Settings}
\label{chp:Environment_Settings}

Here we will expound upon the environment's configuration of the entire code architecture, encompassing domains such as software provisioning, hardware parameters, database integration, and related facets.

The instantiation of this project is grounded in the Python programming language, with compilation facilitated through the PyCharm (professional edition) software. 
Upon successful compilation, the resultant code is subsequently uploaded onto the designated server for operational deployment.

The particular details about the server infrastructure are outlined herewith: The server infrastructure is established on the Ubuntu operating system, boasting four Graphics Processing Units (GPUs) that operate in tandem. 
These GPUs are compatible with the 11.4 version of the Compute Unified Device Architecture (CUDA).
Notably, each GPU has a computational prowess of 350 Watts and 24,268 Megabytes of memory.

The source code's comprehensive architecture encompasses five submodules, each fulfilling designated functions. 
The initial submodule, termed the ``dataset”, functionality encompasses the storage of datasets and ground truth sets in CSV format alongside the capacity to generate synthetic datasets.
Moving forward, the second submodule, designated the ``examples”, serves the dual purpose of harmonizing testing algorithms and facilitating the visual representation of obtained outcomes.
The third submodule, ``save”, is dedicated to the archival of causal discovery results derived from algorithmic testing endeavours.
Within the fourth submodule, designated the ``src”, the central objective pertains to integrating external configuration files, thus ensuring seamless operational functionality of the codebase. 
The final submodule, aptly labelled the ``doc”, assumes the role of elucidating practical usage instances of this code library while presenting information regarding the version of the installation package.

\section{Expirical Results and Analyses}
\label{chp:results}

Within this section, the research questions posited in the preceding section will be addressed systematically, resulting in generalizable findings according to experimental plots and tables.

\subsection{Answer to RQ1: Comparison of Algorithm Performance}
\label{chp:Answer_to_RQ1}

Through meticulous experimentation, we have derived comparative graphs and ranking tables for algorithms across four distinct causality types.
Considering that the ranking algorithm table only displays the average of metrics across all data sizes, we still need violin plots in Appendix \ref{appendix_a} to supplement the changes in recommendation algorithms under specific sample lengths.
This section will subsequently present an in-depth analysis of each category.

The causal relationships, whether linear or nonlinear, in conjunction with noise distributions (Gaussian or non-Gaussian), yield four distinct subgraphs, each representing different data modalities. 
Dataset time lengths are categorized as follows: small (50, 300), medium (300, 1000), and large (1000, 5000). 
Each dataset is processed ten times to compute mean values. 

\subsubsection{Pairwise causality}
\label{chp:RQ1_PW}

Figure \ref{pairwise_fig} illustrates the experimental outcomes for the \textbf{pairwise causal discovery algorithms}. 
We identify an algorithm that consistently performs well as the reference algorithm. 
For instance, within linear relationships, the CCM algorithm serves as the benchmark, whereas the PAI algorithm is selected for nonlinear relationships.
The deviations between the algorithms and their respective reference counterparts are depicted using violin plots in Appendix \ref{appendix_a}.
 
Acknowledging the limitations of violin plots in conveying precise numerical results, we supplement our analysis with a ranking table. 
Table \ref{tab:pairwise ranking} quantifies and compares the performance disparities among the algorithms under consideration. 
The algorithms are ordered in descending sequence based on metric values in this table. 
Higher F1 scores and AUROC values indicate superior model performance, while lower FPR and SHD values reflect better algorithmic behavior.

\begin{table}[ht]
  \begin{center}
    \caption{Ranking Table of Pairwise Causality}
    \scalebox{0.7}{
    \label{tab:pairwise ranking}
    \begin{tabular}{lrrrr}
    \toprule
         & \textbf{Linear Gaussian} & \textbf{Linear Non-Gaussian} &\textbf{Non-Linear Gaussian} &\textbf{Non-Linear Non-Gaussian}\\
    \hline
      \textbf{F1$\uparrow$}  & CCM $(0.56\pm0.04)$ &CCM $(0.57\pm0.06)$&PAI $(0.71\pm0.05)$ & PAI $(0.73\pm0.07)$\\
      & PAI $(0.51\pm0.07)$ & PAI $(0.56\pm0.09)$ & CDS $(0.59\pm0.16)$ &  CCM $(0.65\pm0.08)$\\
      & PWGC $(0.50\pm0.03)$ & PWGC $(0.53\pm0.03)$ & CCM $(0.57\pm0.04)$ & CDS $(0.59\pm0.16)$\\
      & CDS $(0.47\pm0.13)$ & CDS $(0.50\pm0.14)$ & PWGC $(0.48\pm0.04)$ & PWGC $(0.50\pm0.03)$\\
      & IGCI $(0.44\pm0.04)$ & IGCI $(0.48\pm0.04)$& ANM $(0.36\pm0.07)$& ANM $(0.35\pm0.07)$\\
      & ANM $(0.33\pm0.09)$ & ANM $(0.31\pm0.09)$ & IGCI $(0.32\pm0.05)$& IGCI $(0.31\pm0.05)$\\
      & RECI $(0.29\pm0.03)$ & RECI $(0.21\pm0.04)$ & RECI $(0.22\pm0.03)$& RECI $(0.21\pm0.03)$\\
    \hline
     \textbf{AUROC$\uparrow$}  & CCM $(0.67\pm0.03)$ &CCM $(0.67\pm0.05)$ &PAI $(0.78\pm0.04)$ & PAI $(0.80\pm0.05)$\\
      & PAI $(0.63\pm0.05)$ &  PAI $(0.67\pm0.07)$ & CDS $(0.72\pm0.04)$ &  CCM $(0.73\pm0.06)$\\
      & CDS $(0.62\pm0.03)$ & CDS $(0.65\pm0.03)$ & CCM $(0.68\pm0.03)$ & CDS $(0.71\pm0.06)$\\
       & PWGC $(0.62\pm0.03)$ & PWGC $(0.64\pm0.05)$ & PWGC $(0.61\pm0.06)$ & PWGC $(0.62\pm0.03)$\\
       & ANM $(0.59\pm0.04)$ & IGCI $(0.61\pm0.03)$ & ANM $(0.59\pm0.04)$ & ANM $(0.59\pm0.04)$ \\
       & IGCI $(0.58\pm0.03)$ & ANM $(0.59\pm0.03)$& IGCI $(0.49\pm0.04)$& IGCI $(0.49\pm0.04)$ \\
       & RECI $(0.47\pm0.02)$ & RECI $(0.45\pm0.03)$& RECI $(0.42\pm0.02)$& RECI $(0.41\pm0.02)$ \\
       \hline
       \textbf{FPR$\downarrow$}  & RECI $(0.71\pm0.03)$ &RECI $(0.73\pm0.04)$ & RECI $(0.78\pm0.03)$ & RECI $(0.79\pm0.03)$\\
      & IGCI $(0.56\pm0.04)$ &  IGCI $(0.52\pm0.04)$ & IGCI $(0.69\pm0.05)$ &  IGCI $(0.69\pm0.05)$\\
      & PWGC $(0.50\pm0.03)$ & PWGC $(0.47\pm0.03)$ & PWGC $(0.52\pm0.04)$ & PWGC $(0.50\pm0.03)$\\
       & PAI $(0.49\pm0.07)$ & PAI $(0.45\pm0.09)$ & CCM $(0.43\pm0.04)$ & CCM $(0.35\pm0.08)$\\
       & CDS $(0.47\pm0.13)$ & CCM $(0.43\pm0.06)$ & ANM $(0.37\pm0.04)$ & CDS $(0.34\pm0.10)$\\
       & CCM $(0.44\pm0.04)$ & CDS $(0.42\pm0.11)$ & CDS $(0.35\pm0.11)$& ANM $(0.34\pm0.07)$\\
       & ANM $(0.31\pm0.05)$ & ANM $(0.28\pm0.08)$ & PAI $(0.29\pm0.05)$& PAI $(0.27\pm0.07)$\\
       \hline
       \textbf{SHD$\downarrow$}  & RECI $(7.13\pm0.33)$ &RECI $(7.34\pm0.42)$ & RECI $(7.78\pm0.26)$ & RECI $(7.88\pm0.26)$\\
      & ANM $(6.68\pm0.86)$ & ANM $(6.87\pm0.89)$ & IGCI $(6.85\pm0.48)$ &  IGCI $(6.86\pm0.56)$\\
      & IGCI $(5.57\pm0.35)$ & IGCI $(5.24\pm0.36)$ & ANM $(6.42\pm0.72)$ & ANM $(6.54\pm0.73)$\\
       & CDS $(5.32\pm1.28)$ & CDS $(4.96\pm1.40)$ & PWGC $(5.22\pm0.41)$ & PWGC $(4.97\pm0.33)$\\
       & PWGC $(4.97\pm0.32)$ & PWGC $(4.66\pm0.34)$& CCM $(4.27\pm0.42)$ & CDS $(4.13\pm1.60)$\\
       & PAI $(4.93\pm0.72)$ & PAI $(4.45\pm0.93)$ & CDS $(4.15\pm1.64)$& CCM $(3.54\pm0.75)$\\
       & CCM $(4.39\pm0.39)$ & CCM $(4.34\pm0.62)$ & PAI $(2.88\pm0.54)$& PAI $(2.71\pm0.68)$\\
       \bottomrule
    \end{tabular}}
  \end{center}
\end{table}

Drawing insights from Figure \ref{pairwise_fig} and Table \ref{tab:pairwise ranking}, we synthesize a comprehensive summary delineating the preeminent algorithmic selections under diverse application scenarios. 
In scenarios where the overall performance of the model is emphasized, the F1 score assumes paramount importance.
For the linear dataset, CCM is advocated as the optimal choice, while the RECI algorithm, deemed the least effective, exhibits an F1 score approximately 50\% lower than that of CCM.
Across other datasets, the PAI algorithm consistently emerges as the pinnacle performer. 
Particularly in nonlinear and Gaussian datasets, PAI outperforms its peers significantly, with its F1 score surpassing that of the second-best algorithm by 20\%.
Note that the optimal algorithm for each data size is always among the top three algorithms in the average ranking, which means that the sample length has little impact on our recommendation for F1.

In the pursuit of heightened system robustness, AUROC is prioritized. 
The recommendations from this scenario align with those derived from the F1 score, reaffirming the advisability of employing either the CCM or PAI algorithms.
However, the PAI algorithm is not competitive in small sample sizes of linear datasets and is only recommended in medium or large sample sizes ($L>300$).

For systems sensitive to false causality, preference should be given to the FPR metric. 
In cases involving linear interrelationships among variables, ANM is the preferred choice, with the worst algorithm's FPR being more than twice as high as that of ANM.
CDS is more recommended in small size datasets.
Conversely, when dealing with nonlinear relationships among variables, the PAI algorithm emerges as the optimal selection.

When dealing with situations characterized by limited error tolerance, SHD takes precedence. 
The distinction between the CCM and PAI algorithms is minimal within linear datasets. 
However, in nonlinear datasets, PAI demonstrates a noteworthy reduction in SHD, exceeding a minimum of 20\% compared to other algorithms.


\subsubsection{Instantaneous causality}
\label{chp:RQ1_inst}

Following this exposition, we elucidate the performance evaluation of \textbf{instantaneous causal discovery algorithms}, as visually depicted in Figure \ref{instantaneous_fig} and quantitatively delineated in Table \ref{tab:instantaneous ranking}. 
Figure \ref{instantaneous_fig} records the performance of seven algorithms, with oCSE serving as the reference algorithm for Gaussian noise datasets and VARLiNGAM as the reference for non-Gaussian noise datasets.

Similar to the analysis of pairwise causality, we consider five scenarios when evaluating instantaneous causality algorithms.
The top 3 best-performing algorithms under each metric can be summarized from Table \ref{tab:instantaneous ranking}. We will present them in Table \ref{tab:recommendation} and not elaborate here.
It is necessary to supplement some insights based on data sample size in Figure \ref{instantaneous_fig}.

Firstly, when prioritizing F1 scores, for non-linear Gaussian datasets, tsFCI performs in the top three in small data sizes, while PCMCI performs better in large data sizes.
For Nonlinear non-Gaussian datasets, VARLiNGAM will only be recommended when the sample length is small ($L<1000$). 
The average ranking of algorithm performance under AUROC metric is basically consistent with the ranking under each data size.

In the context of prioritizing FPR metrics, PCMCI belongs to the recommendation algorithms in all four data types, indicating its superior performance under the FPR metric.
For Linear Gaussian datasets, tsFCI is only recommended when the sample size is small.

Under the SHD metric, oCSE, PCMCI, and VARLiNGAM are recommended for all four data types, proving that these three algorithms have stable performance to find the real causal relations and are not heavily dependent on data features.



\begin{table}[ht]
  \begin{center}
    \caption{Ranking Table of Instantaneous Causality}
    \scalebox{0.65}{
    \label{tab:instantaneous ranking}
    \begin{tabular}{lrrrr}
    \toprule
         & \textbf{Linear Gaussian} & \textbf{Linear Non-Gaussian} &\textbf{Non-Linear Gaussian} &\textbf{Non-Linear Non-Gaussian}\\
    \hline
      \textbf{F1$\uparrow$}  & oCSE$(0.74\pm0.23)$ & VARLiNGAM$(0.88\pm0.13)$& NeuralGC $(0.53\pm0.10)$ & NeuralGC $(0.52\pm0.07)$\\
      & PCMCI $(0.71\pm0.20)$ & oCSE $(0.72\pm0.18)$ & tsFCI $(0.45\pm0.08)$ & VARLiNGAM $(0.50\pm0.15)$\\
      & VARLiNGAM $(0.63\pm0.07)$ & PCMCI $(0.71\pm0.16)$ & PCMCI $(0.44\pm0.20)$ & tsFCI $(0.48\pm0.10)$\\
      & NeuralGC $(0.61\pm0.12)$ & DYNOTEARS $(0.64\pm0.10)$ & oCSE $(0.42\pm0.21)$ & PCMCI $(0.41\pm0.12)$\\
       & DYNOTEARS $(0.56\pm0.12)$ & NeuralGC $(0.61\pm0.12)$ & VARLiNGAM $(0.36\pm0.15)$ & oCSE $(0.40\pm0.14)$\\
     & tsFCI $(0.50\pm0.07)$ & tsFCI $(0.50\pm0.05)$ & DYNOTEARS $(0.31\pm0.15)$ & DYNOTEARS $(0.37\pm0.13)$\\
      & TCDF $(0.32\pm0.16)$ & TCDF $(0.39\pm0.12)$ & TCDF $(0.22\pm0.15)$ & TCDF $(0.24\pm0.11)$\\
    \hline
     \textbf{AUROC$\uparrow$}  & oCSE $(0.81\pm0.14)$ &VARLiNGAM $(0.88\pm0.13)$ &NeuralGC $(0.65\pm0.07)$ & NeuralGC $(0.64\pm0.05)$\\
      & PCMCI $(0.77\pm0.14)$ &  oCSE $(0.79\pm0.12)$ & oCSE $(0.59\pm0.13)$ & tsFCI $(0.58\pm0.10)$ \\
      & NeuralGC $(0.72\pm0.07)$ & PCMCI $(0.78\pm0.11)$ & PCMCI $(0.58\pm0.14)$ & oCSE $(0.58\pm0.08)$\\
       & VARLiNGAM $(0.70\pm0.06)$ & NeuralGC $(0.71\pm0.07)$ & tsFCI $(0.56\pm0.08)$ & VARLiNGAM $(0.57\pm0.14)$\\
       & tsFCI $(0.64\pm0.05)$ & DYNOTEARS $(0.70\pm0.10)$ & VARLiNGAM $(0.45\pm0.14)$ & PCMCI $(0.41\pm0.12)$\\
       & DYNOTEARS $(0.63\pm0.10)$ & tsFCI $(0.63\pm0.04)$ & DYNOTEARS $(0.41\pm0.15)$ & DYNOTEARS $(0.47\pm0.14)$\\
       & TCDF $(0.41\pm0.14)$ & TCDF $(0.48\pm0.09)$ & TCDF $(0.31\pm0.13)$ & TCDF $(0.34\pm0.10)$\\
       \hline
       \textbf{FPR$\downarrow$}  & NeuralGC $(0.79\pm0.27)$ &NeuralGC $(0.82\pm0.22)$ & NeuralGC$(0.95\pm0.26)$ & NeuralGC $(0.92\pm0.28)$\\
      & TCDF $(0.70\pm0.36)$ &  tsFCI $(0.60\pm0.33)$ & TCDF $(0.81\pm0.28)$ &  TCDF $(0.69\pm0.22)$\\
      & DYNOTEARS $(0.63\pm0.42)$ & TCDF $(0.40\pm0.23)$ & VARLiNGAM $(0.76\pm0.35)$ & DYNOTEARS $(0.66\pm0.39)$\\
       & VARLiNGAM $(0.62\pm0.25)$ & DYNOTEARS $(0.37\pm0.32)$ & DYNOTEARS $(0.75\pm0.34)$ & VARLiNGAM$(0.54\pm0.41)$\\
       & tsFCI $(0.61\pm0.35)$ & PCMCI $(0.23\pm0.14)$ & tsFCI $(0.57\pm0.28)$ & tsFCI $(0.48\pm0.30)$\\
       & PCMCI $(0.24\pm0.17)$ & oCSE $(0.22\pm0.16)$ & PCMCI $(029\pm0.15)$ &  PCMCI $(0.29\pm0.17)$\\
       & oCSE $(0.22\pm0.19)$ &  VARLiNGAM $(0.21\pm0.29)$ & oCSE $(0.25\pm0.16)$ & oCSE $(0.22\pm0.19)$\\
       \hline
       \textbf{SHD$\downarrow$}  & tsFCI $(8.41\pm1.42)$ &tsFCI $(8.36\pm1.33)$ &tsFCI$(8.21\pm1.12)$ & tsFCI $(7.78\pm1.13)$\\
      & NeuralGC $(6.37\pm2.60)$ &  NeuralGC $(6.19\pm2.27)$ & NeuralGC $(7.20\pm2.00)$ &  NeuralGC $(7.25\pm1.96)$\\
      & TCDF $(6.02\pm0.98)$ & TCDF $(5.76\pm0.74)$ & DYNOTEARS $(6.80\pm1.44)$ & TCDF $(6.56\pm0.77)$\\
       & DYNOTEARS $(5.67\pm1.59)$ & DYNOTEARS $(4.45\pm1.59)$ & TCDF $(6.77\pm1.29)$ & DYNOTEARS $(5.97\pm1.62)$\\
       & VARLiNGAM $(5.24\pm0.99)$ & PCMCI $(2.88\pm1.18)$ & VARLiNGAM $(6.44\pm1.62)$ & PCMCI $(5.37\pm0.71)$\\
       & PCMCI $(2.92\pm1.53)$ & oCSE $(2.73\pm1.39)$ & oCSE $(4.91\pm1.03)$ & VARLiNGAM $(5.33\pm1.71)$\\
       & oCSE $(2.57\pm1.60)$ & VARLiNGAM $(1.81\pm1.98)$ & PCMCI $(4.91\pm1.10)$ & oCSE $(5.21\pm0.69)$\\
       \bottomrule
    \end{tabular}}
  \end{center}
\end{table}

\subsubsection{Time-delay Causality}
Below, we will analyze the performance comparison of \textbf{time-delay causal algorithms} in detail.
The top 3 best-performing algorithms under each metric can be summarized from Table \ref{tab:time-delay ranking}. We will present them in Table \ref{tab:recommendation} and not elaborate here.

\begin{table}[ht]
  \begin{center}
    \caption{Ranking Table of Time-delay Causality}
    \scalebox{0.65}{
    \label{tab:time-delay ranking}
    \begin{tabular}{lrrrr}
    \toprule
         & \textbf{Linear Gaussian} & \textbf{Linear Non-Gaussian} &\textbf{Non-Linear Gaussian} &\textbf{Non-Linear Non-Gaussian}\\
    \hline
      \textbf{F1$\uparrow$}  & VARLiNGAM $(0.87\pm0.14)$ & PCMCI$(0.88\pm0.11)$&PCMCI $(0.78\pm0.08)$ & PCMCI $(0.79\pm0.07)$\\
      & MVGC $(0.86\pm0.09)$ & VARLiNGAM $(0.87\pm0.13)$ & DYNOTEARS $(0.75\pm0.07)$ &  NeuralGC $(0.77\pm0.09)$\\
      & DYNOTEARS $(0.86\pm0.11)$ & DYNOTEARS $(0.87\pm0.11)$ & VARLiNGAM $(0.74\pm0.07)$ & DYNOTEARS $(0.74\pm0.09)$\\
       & PCMCI $(0.86\pm0.11)$ & MVGC $(0.87\pm0.11)$ & NeuralGC $(0.74\pm0.07)$ & oCSE $(0.73\pm0.09)$\\
     & NeuralGC $(0.83\pm0.13)$ & NeuralGC $(0.83\pm0.14)$ & MVGC $(0.72\pm0.06)$ & VARLiNGAM $(0.73\pm0.07)$\\
      & oCSE $(0.78\pm0.15)$ & oCSE $(0.81\pm0.12)$ & oCSE $(0.71\pm0.10)$ & MVGC $(0.72\pm0.08)$\\
      & TCDF $(0.71\pm0.12)$ & TCDF $(0.73\pm0.10)$& tsFCI $(0.67\pm0.06)$& tsFCI $(0.67\pm0.04)$\\
      & tsFCI $(0.68\pm0.07)$ & tsFCI $(0.69\pm0.05)$& TCDF $(0.60\pm0.07)$ & TCDF $(0.60\pm0.10)$\\
    \hline
     \textbf{AUROC$\uparrow$}  & MVGC $(0.93\pm0.06)$ & MVGC $(0.93\pm0.06)$ &PCMCI $(0.83\pm0.05)$ & PCMCI $(0.84\pm0.04)$\\
      & VARLiNGAM $(0.92\pm0.08)$ & VARLiNGAM $(0.93\pm0.08)$ & MVGC $(0.82\pm0.05)$ & MVGC $(0.84\pm0.05)$ \\
      & DYNOTEARS $(0.90\pm0.09)$ & PCMCI $(0.91\pm0.07)$ & DYNOTEARS $(0.81\pm0.07)$ & NeuralGC $(0.82\pm0.07)$\\
       & PCMCI $(0.90\pm0.07)$ & DYNOTEARS $(0.91\pm0.09)$ & VARLiNGAM $(0.81\pm0.05)$ & oCSE $(0.82\pm0.06)$\\
       & NeuralGC $(0.87\pm0.10)$ & oCSE $(0.87\pm0.07)$ & NeuralGC $(0.80\pm0.06)$ & DYNOTEARS $(0.81\pm0.07)$\\
       & oCSE $(0.86\pm0.08)$ & NeuralGC $(0.87\pm0.10)$ & oCSE $(0.79\pm0.08)$ & VARLiNGAM $(0.80\pm0.05)$\\
       & TCDF $(0.82\pm0.09)$ & TCDF $(0.84\pm0.08)$ & tsFCI $(0.74\pm0.04)$ & tsFCI $(0.75\pm0.02)$\\
       & tsFCI $(0.76\pm0.03)$ & tsFCI $(0.77\pm0.03)$& TCDF $(0.73\pm0.07)$ & TCDF $(0.72\pm0.09)$\\
       \hline
       \textbf{FPR$\downarrow$}  & tsFCI $(0.49\pm0.26)$ & tsFCI $(0.46\pm0.28)$ & NeuralGC $(0.64\pm0.38)$ & NeuralGC $(0.62\pm0.37)$\\
       & NeuralGC $(0.45\pm0.39)$ & NeuralGC $(0.43\pm0.41)$ & TCDF $(0.50\pm0.21)$ & TCDF $(0.54\pm0.26)$\\
      & oCSE $(0.33\pm0.32)$ & oCSE $(0.31\pm0.30)$ & tsFCI $(0.46\pm0.20)$ &  DYNOTEARS $(0.51\pm0.35)$\\
      & DYNOTEARS $(0.26\pm0.29)$ & DYNOTEARS $(0.25\pm0.35)$ & DYNOTEARS $(0.46\pm0.35)$ & tsFCI $(0.49\pm0.23)$\\
       & PCMCI $(0.25\pm0.19)$ & PCMCI $(0.21\pm0.17)$ & PCMCI $(0.45\pm0.16)$ & PCMCI$(0.47\pm0.21)$\\
       & TCDF $(0.24\pm0.25)$ &  TCDF $(0.18\pm0.20)$ & oCSE $(0.41\pm0.23)$ &  VARLiNGAM $(0.45\pm0.21)$\\
       & VARLiNGAM $(0.11\pm0.12)$ & VARLiNGAM $(0.09\pm0.13)$ & VARLiNGAM $(0.36\pm0.24)$ &  oCSE $(0.37\pm0.23)$\\
       & MVGC $(0.07\pm0.24)$ & MVGC $(0.04\pm0.13)$ & MVGC $(0.25\pm0.25)$ &  MVGC $(0.22\pm0.22)$\\
       \hline
       \textbf{SHD$\downarrow$}  & tsFCI $(7.15\pm1.45)$ & tsFCI $(7.15\pm1.51)$ & tsFCI $(7.33\pm1.13)$ & tsFCI $(7.17\pm1.17)$\\
      & TCDF $(4.58\pm1.68)$ & TCDF $(4.35\pm1.33)$ & TCDF $(6.10\pm0.92)$ & TCDF $(6.39\pm1.33)$\\
      & oCSE $(4.15\pm2.22)$ & NeuralGC $(4.12\pm3.59)$ & NeuralGC $(5.86\pm1.90)$ & NeuralGC $(5.38\pm2.41)$\\
      & NeuralGC $(4.14\pm3.33)$ & oCSE  $(3.62\pm1.74)$& oCSE $(5.54\pm1.52)$ & VARLiNGAM $(5.34\pm1.14)$\\
       & DYNOTEARS $(2.91\pm2.47)$ & DYNOTEARS $(2.68\pm2.60)$ & VARLiNGAM $(5.31\pm1.17)$ & DYNOTEARS $(5.25\pm2.02)$\\
       & PCMCI $(2.80\pm1.95)$ & MVGC $(2.43\pm1.97)$ & MVGC $(5.25\pm0.85)$ & oCSE $(5.06\pm1.58)$\\
       & MVGC $(2.49\pm1.60)$ & PCMCI $(2.30\pm1.80)$ & DYNOTEARS $(5.10\pm1.62)$ & MVGC $(4.97\pm1.09)$\\
       & VARLiNGAM $(2.29\pm1.98)$ & VARLiNGAM $(2.18\pm2.06)$ & PCMCI $(4.41\pm1.00)$ & PCMCI $(4.20\pm1.11)$\\
       \bottomrule
    \end{tabular}}
  \end{center}
\end{table}

It is necessary to supplement some insights based on data sample size in Figure \ref{time-delay_fig}.
Firstly, we analyze the algorithms under the F1 metric. 
For Linear Gaussian data, VARLiNGAM performs better in larger sample sizes. On the contrary, MVGC is more suitable for small sample size data.
The top three ranking algorithms perform relatively stably on Linear non-Gaussian and Nonlinear Gaussian data, and their rankings do not change significantly based on changes in sample size.
For Nonlinear non-Gaussian data, DYNOTEARS is recommended only when the sample size is small, since it does not ranked in the top three in large sample size.

The ordering of algorithms according to the AUROC metric exhibits a relatively stable pattern.
Specifically, the top three algorithms in average ranking table reflects the recommended algorithms for AUROC across all data sizes.

In evaluating performance utilizing the FPR metric, for linear data, TCDF ranks in the top three only when the sample length $> 1000$, so it is not a recommended algorithm when the sample size is small.
For non-linear data, VARLiNGAM is not outstanding when the sample length $<300$ and is only recommended for use in large data size.
Note that although the oCSE algorithm does not always in the top three of the mean ranking, it is always the best algorithm in small sample sizes ($L<300$).

Considering SHD, for linear type, VARLiNGAM is more suitable for large sample data, while PCMCI is more suitable for small sample data.
For nonlinear type, MVGC is more suitable for small sample data.
In Nonlinear non-Gaussian data, although NeuralGC is not among the top three algorithms, it is the best algorithm when sample length $>1000$ and can be recommended as a supplementary algorithm.


After comparing the performance of time-delay causal discovery algorithms, it can be concluded that tsFCI are not recommended in any scenario, as they are not competitive across any metric.

\subsubsection{i.i.d. Causality}
Below, we will analyze the performance comparison of \textbf{i.i.d. data causal algorithms} in detail.

\begin{table}[htbp]
  \begin{center}
    \caption{Ranking Table of i.i.d. Causality}
    \scalebox{0.51}{
    \label{tab:iid ranking}
    \begin{tabular}{lrrrr}
    \toprule
         & \textbf{Linear Gaussian} & \textbf{Linear Non-Gaussian} &\textbf{Non-Linear Gaussian} &\textbf{Non-linear Non-Gaussian}\\
    \hline
      \textbf{F1$\uparrow$} & GOLEM $(0.99\pm0.02)$ & DirectLiNGAM$(0.99\pm0.05)$& GraNDAG $(0.51\pm0.24)$ & CORL $(0.51\pm0.14)$\\
      & CORL $(0.98\pm0.04)$ & CORL $(0.98\pm0.04)$ & CORL $(0.49\pm0.07)$ &  GOLEM $(0.46\pm0.11)$\\
      & DAG-GNN $(0.95\pm0.08)$ & GOLEM $(0.95\pm0.07)$ & GES $(0.44\pm0.11)$ & GraNDAG $(0.39\pm0.18)$\\
      & NOTEARS $(0.90\pm0.10)$ & ICALiNGAM $(0.94\pm0.09)$ & GOLEM $(0.44\pm0.05)$ & ICALiNGAM $(0.36\pm0.15)$\\
       & ES $(0.85\pm0.11)$ & DAG-GNN $(0.91\pm0.09)$ & RCD $(0.43\pm0.06)$ & DAG-GNN $(0.35\pm0.12)$\\
     & ICALiNGAM $(0.70\pm0.13)$ & RCD $(0.87\pm0.17)$ & FCI $(0.41\pm0.12)$ & RCD $(0.33\pm0.06)$\\
      & GRaSP $(0.62\pm0.13)$ & NOTEARS $(0.86\pm0.09)$ & PC $(0.40\pm0.12)$ & GES $(0.30\pm0.06)$\\
      & PC $(0.57\pm0.10)$& GRaSP $(0.66\pm0.15)$ & ICALiNGAM $(0.38\pm0.08)$ & PC $(0.27\pm0.09)$\\
      & FCI $(0.51\pm0.16)$& PC $(0.58\pm0.10)$ & ES $(0.35\pm0.12)$ & NOTEARS $(0.25\pm0.11)$\\
      & GES $(0.46\pm0.14)$& FCI $(0.51\pm0.13)$ & NOTEARS $(0.32\pm0.08)$ & FCI $(0.24\pm0.08)$\\
      & DirectLiNGAM $(0.40\pm0.09)$& ES $(0.51\pm0.13)$ & DAG-GNN $(0.31\pm0.06)$ & ES $(0.23\pm0.08)$\\
      & GraNDAG $(0.36\pm0.24)$& GES $(0.49\pm0.13)$ & GRaSP $(0.29\pm0.18)$ & GRaSP $(0.15\pm0.07)$\\
      & RCD $(0.35\pm0.04)$& GraNDAG $(0.18\pm0.16)$ & DirectLiNGAM $(0.11\pm0.06)$ & DirectLiNGAM $(0.12\pm0.08)$\\
    \hline
     \textbf{AUROC$\uparrow$} & GOLEM $(0.99\pm0.02)$ &DirectLiNGAM $(0.99\pm0.04)$ & CORL $(0.66\pm0.06)$ & CORL $(0.61\pm0.11)$\\
      & CORL $(0.98\pm0.03)$ &  CORL $(0.98\pm0.04)$ & NOTEARS $(0.63\pm0.04)$ & GOLEM $(0.58\pm0.09)$ \\
      & DAG-GNN $(0.96\pm0.07)$ & GOLEM $(0.96\pm0.06)$ & GraNDAG $(0.62\pm0.23)$ & DAG-GNN $(0.53\pm0.09)$\\
       & NOTEARS $(0.91\pm0.09)$ & ICALiNGAM $(0.95\pm0.08)$ & ICALiNGAM $(0.60\pm0.05)$ & NOTEARS $(0.52\pm0.08)$\\
       & ES $(0.86\pm0.09)$ & DAG-GNN $(0.92\pm0.08)$ & DAG-GNN $(0.59\pm0.05)$ & ICALiNGAM $(0.48\pm0.12)$\\
       & ICALiNGAM $(0.73\pm0.12)$ & RCD $(0.89\pm0.14)$ & GOLEM $(0.56\pm0.07)$ & GraNDAG $(0.47\pm0.19)$\\
       & GRaSP $(0.67\pm0.13)$ & NOTEARS $(0.88\pm0.09)$ & RCD $(0.52\pm0.07)$ & RCD $(0.46\pm0.06)$\\
       & FCI $(0.61\pm0.12)$& GRaSP $(0.71\pm0.13)$ & GES $(0.49\pm0.10)$ & GES $(0.35\pm0.06)$ \\
       & PC $(0.61\pm0.09)$& FCI $(0.63\pm0.08)$ & FCI $(0.49\pm0.09)$ & GRaSP $(0.33\pm0.10)$\\
       & RCD $(0.60\pm0.02)$& PC $(0.61\pm0.09)$ & GRaSP $(0.48\pm0.11)$ & PC $(0.31\pm0.08)$\\
       & GES $(0.52\pm0.13)$& ES $(0.56\pm0.12)$ & PC $(0.46\pm0.11)$ & FCI $(0.30\pm0.08)$\\
       & GraNDAG $(0.47\pm0.23)$& GES $(0.54\pm0.12)$ & ES $(0.40\pm0.11)$ & ES $(0.28\pm0.08)$\\
       & DirectLiNGAM $(0.45\pm0.08)$& GraNDAG $(0.29\pm0.19)$ & DirectLiNGAM $(0.27\pm0.08)$ & DirectLiNGAM $(0.25\pm0.12)$\\
       \hline
       \textbf{FPR$\downarrow$}  & RCD $(0.75\pm0.21)$ & ES $(0.46\pm0.11)$ & ES $(0.38\pm0.12)$ & ES $(0.29\pm0.17)$\\
      & GES $(0.50\pm0.19)$ &  GES $(0.42\pm0.17)$ & GraNDAG $(0.24\pm0.34)$ & GraNDAG $(0.28\pm0.34)$\\
      & DirectLiNGAM $(0.41\pm0.10)$ & PC $(0.18\pm0.06)$ & PC $(0.22\pm0.06)$ & PC $(0.22\pm0.13)$\\
       & ICALiNGAM $(0.23\pm0.11)$ & GRaSP $(0.12\pm0.09)$ & GES $(0.21\pm0.06)$ & GES$(0.22\pm0.15)$\\
       & PC $(0.17\pm0.05)$ &  GraNDAG $(0.10\pm0.06)$ & RCD $(0.19\pm0.14)$ &  RCD $(0.21\pm0.16)$\\
       & GRaSP $(0.16\pm0.12)$ &  RCD $(0.09\pm0.13)$ & FCI $(0.12\pm0.05)$ &  FCI $(0.16\pm0.09)$\\
       & GraNDAG $(0.12\pm0.11)$ &  DAG-GNN $(0.06\pm0.07)$ & GOLEM $(0.11\pm0.07)$ & GRaSP $(0.09\pm0.09)$\\
       & ES $(0.09\pm0.06)$& FCI $(0.05\pm0.04)$ & GRaSP $(0.09\pm0.06)$ & GOLEM $(0.06\pm0.05)$\\
       & FCI $(0.06\pm0.04)$& NOTEARS $(0.05\pm0.05)$ & DirectLiNGAM $(0.07\pm0.03)$ & DirectLiNGAM $(0.05\pm0.02)$\\
       & DAG-GNN $(0.05\pm0.09)$& GOLEM $(0.04\pm0.07)$ & CORL $(0.04\pm0.05)$ & CORL $(0.05\pm0.04)$\\
       & NOTEARS $(0.04\pm0.05)$& ICALiNGAM $(0.04\pm0.05)$ & ICALiNGAM $(0.04\pm0.02)$ & ICALiNGAM $(0.05\pm0.03)$\\
       & CORL $(0.02\pm0.05)$& CORL $(0.02\pm0.06)$ & DAG-GNN $(0.03\pm0.03)$ & DAG-GNN $(0.04\pm0.03)$\\
       & GOLEM $(0.01\pm0.02)$& DirectLiNGAM $(0.01\pm0.03)$ & NOTEARS $(0.01\pm0.01)$ & NOTEARS $(0.02\pm0.02)$\\
       \hline
       \textbf{SHD$\downarrow$}  & RCD $(37.6\pm5.95)$ & GES $(16.79\pm4.62)$ & RCD $(21.51\pm4.16)$ & RCD $(16.53\pm5.91)$\\
      & GES $(18.72\pm5.41)$ &  GraNDAG $(15.75\pm2.16)$ & ES $(17.13\pm3.39)$ & GraNDAG $(13.95\pm10.57)$\\
      & DirectLiNGAM $(17.45\pm3.51)$ & ES $(15.35\pm3.93)$ & DirectLiNGAM $(15.20\pm0.60)$ & ES $(13.93\pm4.61)$\\
       & GraNDAG $(14.20\pm4.60)$ & FCI $(11.03\pm1.75)$ & FCI $(14.81\pm1.11)$ & GES $(12.97\pm4.03)$\\
       & FCI $(10.96\pm2.02)$ & PC $(10.84\pm2.21)$ & GraNDAG $(13.91\pm9.08)$ & PC $(12.63\pm3.56)$\\
       & PC $(10.76\pm2.05)$ & GRaSP $(8.48\pm3.33)$ & GES $(13.55\pm2.26)$ & FCI $(11.79\pm3.41)$\\
       & GRaSP $(9.83\pm3.35)$ & RCD $(4.43\pm6.60)$ & GRaSP $(13.33\pm2.11)$ & GRaSP $(9.65\pm2.21)$\\
       & ICALiNGAM $(8.27\pm4.09)$& NOTEARS $(3.35\pm2.54)$ & PC $(13.24\pm2.22)$ & DirectLiNGAM $(9.13\pm1.91)$\\
       & ES $(3.40\pm2.35)$& DAG-GNN $(2.72\pm0.2.77)$ & DAG-GNN $(13.21\pm0.75)$& NOTEARS $(8.31\pm1.67)$\\
       & NOTEARS $(2.56\pm2.70)$& GOLEM $(1.63\pm2.16)$ & GOLEM $(13.19\pm1.76)$ & DAG-GNN $(7.95\pm1.73)$\\
       & DAG-GNN $(1.56\pm2.79)$& ICALiNGAM $(1.43\pm2.25)$ & NOTEARS $(12.93\pm0.90)$ & ICALiNGAM $(7.77\pm1.91)$\\
       & CORL $(0.56\pm1.39)$& CORL $(0.67\pm1.61)$ & ICALiNGAM $(12.24\pm1.05)$ & GOLEM $(7.64\pm1.85)$\\
       & GOLEM $(0.33\pm0.71)$& DirectLiNGAM $(0.36\pm1.35)$ & CORL $(10.95\pm1.39)$ & CORL $(6.84\pm1.99)$\\      
       \bottomrule
    \end{tabular}}
  \end{center}
\end{table}

Based on Figure \ref{iid_fig} and Table \ref{tab:iid ranking}, we can derive insights into the performance of recommendation algorithms across different i.i.d. data types. 
Analyzing the algorithms using the F1 metric, Table 8 reveals that the optimal algorithms vary according to specific data characteristics, with the CORL algorithm consistently ranking among the top three performers.

When considering the AUROC metric, the CORL algorithm shows superior performance on nonlinear datasets, although the improvement over the second-best algorithm is less than 10\%. 
In contrast, GOLEM or DirectLiNGAM exhibit a slight advantage over CORL when applied to linear datasets.

Assessing performance using the FPR metric, NOTEARS emerges as the most proficient algorithm for nonlinear datasets. The GraNDAG algorithm performs best when the sample size is large, so we also include it in Table \ref{tab:iid ranking}.
For linear datasets, GOLEM performs best under Gaussian noise distribution, while DirectLiNGAM excels with non-Gaussian noise. Note that we need to supplement the DAG-GNN algorithm on large datasets, although it performs poorly on small-sized datasets.

Evaluated using the SHD metric, CORL demonstrates optimal performance on nonlinear datasets. 
Conversely, GOLEM achieves the lowest SHD values on linear datasets with Gaussian noise, while DirectLiNGAM performs best with non-Gaussian noise.


\subsubsection{Discussion on Algorithm Efficiency}
It is known that effectiveness does not equal efficiency.
Even if some algorithms have good causal discovery performance, they may not be suitable for users because of long running time.
So here we will specifically discuss the running time of the algorithms we tested in the first four sections, as shown in Table \ref{tab:runtime ranking}, to help users make a trade-off between effectiveness and efficiency.
\begin{table}[ht]
  \begin{center}
    \caption{Ranking Table of Run Time (s)$\downarrow$}
    \scalebox{0.55}{
    \label{tab:runtime ranking}
    \begin{tabular}{lrrrr}
    \toprule
         & \textbf{Linear Gaussian} & \textbf{Linear Non-Gaussian} &\textbf{Non-Linear Gaussian} &\textbf{Non-Linear Non-Gaussian}\\
    \hline
        \textbf{Pairwise}  & ANM $(13.19\pm24.70)$ &ANM $(13.22\pm24.77)$ &ANM $(13.34\pm25.09)$ & ANM $(13.20\pm24.85)$\\
      & CCM $(6.03\pm10.49)$ &  CCM $(6.03\pm10.50)$ & CCM $(5.99\pm10.40)$ &  CCM $(6.00\pm10.45)$\\
      & PAI $(6.02\pm10.48)$ & PAI $(6.02\pm10.48)$ & PAI $(5.98\pm10.40)$ & PAI $(5.98\pm10.40)$\\
       & CDS $(0.68\pm0.39)$ & CDS $(0.68\pm0.39)$ & CDS $(0.75\pm0.48)$ & CDS $(0.75\pm0.47)$\\
       & IGCI $(0.06\pm0.06)$ & IGCI $(0.06\pm0.06)$ & IGCI $(0.06\pm0.06)$ & IGCI $(0.06\pm0.06)$\\
       & PWGC $(0.05\pm0.02)$ & PWGC $(0.05\pm0.02)$& PWGC $(0.05\pm0.02)$& PWGC $(0.05\pm0.02)$\\
       & RECI $(0.04\pm0.01)$ & RECI $(0.04\pm0.01)$ & RECI $(0.04\pm0.01)$& RECI $(0.04\pm0.01)$\\
        \hline
        \textbf{Instantaneous}  & NeuralGC $(88.38\pm35.99)$ &NeuralGC $(88.10\pm30.65)$ &NeuralGC $(86.15\pm33.63)$ & NeuralGC $(88.51\pm31.86)$\\
      & TCDF $(4.97\pm0.15)$ & TCDF $(4.92\pm0.12)$ & TCDF $(4.93\pm0.11)$ &  TCDF $(4.96\pm0.11)$\\
      & tsFCI $(0.66\pm0.12)$ & tsFCI $(0.67\pm0.14)$ & tsFCI $(0.66\pm0.15)$ & tsFCI $(0.66\pm0.15)$\\
       & oCSE $(0.51\pm0.29)$ & oCSE $(0.49\pm0.29)$ & oCSE $(0.43\pm0.31)$ & oCSE $(0.41\pm0.28)$\\
       & DYNOTEARS $(0.32\pm0.14)$ & DYNOTEARS $(0.28\pm0.10)$ & DYNOTEARS $(0.28\pm0.15)$ & DYNOTEARS $(0.27\pm0.18)$\\
       & PCMCI $(0.10\pm0.04)$ & PCMCI $(0.10\pm0.04)$ & PCMCI $(0.09\pm0.03)$ & PCMCI $(0.09\pm0.03)$\\
       & VARLiNGAM $(0.09\pm0.10)$ & VARLiNGAM $(0.09\pm0.10)$ & VARLiNGAM $(0.09\pm0.01)$ & VARLiNGAM $(0.09\pm0.10)$\\
        \hline
        \textbf{Time-delay}  & NeuralGC $(112.34\pm55.35)$ &NeuralGC $(113.11\pm50.86)$ &NeuralGC $(105.62\pm48.50)$ & NeuralGC $(101.19\pm42.87)$\\
      & TCDF $(4.97\pm0.15)$ & TCDF $(4.90\pm0.15)$ & TCDF $(4.85\pm0.14)$ &  TCDF $(4.94\pm0.14)$\\
      & MVGC $(1.22\pm1.39)$ & MVGC $(1.20\pm1.37)$ & MVGC $(1.24\pm1.43)$ & MVGC $(1.21\pm1.37)$\\
       & tsFCI $(1.05\pm0.71)$ & tsFCI $(0.86\pm0.29)$ & tsFCI $(1.04\pm0.30)$ & tsFCI $(0.99\pm0.47)$\\
       & oCSE $(0.60\pm0.42)$ & oCSE $(0.59\pm0.38)$ & oCSE $(0.58\pm0.35)$ & oCSE $(0.56\pm0.39)$\\
       & PCMCI $(0.52\pm0.37)$ & PCMCI $(0.46\pm0.21)$ & PCMCI $(0.49\pm0.24)$ & PCMCI $(0.53\pm0.36)$\\
       & VARLiNGAM $(0.13\pm0.14)$ & VARLiNGAM $(0.13\pm0.14)$ & DYNOTEARS $(0.16\pm0.16)$ & DYNOTEARS $(0.18\pm0.16)$\\
       & DYNOTEARS $(0.08\pm0.08)$ & DYNOTEARS $(0.06\pm0.05)$& VARLiNGAM $(0.13\pm0.14)$& VARLiNGAM $(0.13\pm0.14)$\\
        \hline
        \textbf{i.i.d}  & GraNDAG $(211.26\pm5.44)$ &GraNDAG $(213.19\pm19.10)$ & CORL $(248.73\pm31.10)$ & CORL $(237.54\pm1.79)$\\
      & CORL $(194.95\pm4.56)$ &  CORL $(186.32\pm1.91)$ & GraNDAG $(224.22\pm28.42)$ &  GraNDAG $(214.67\pm2.71)$\\
      & DAG-GNN $(92.73\pm102.88)$ & RCD $(115.50\pm320.54)$ & DAG-GNN $(95.26\pm98.78)$ & DAG-GNN $(95.89\pm106.63)$\\
       & GOLEM $(28.24\pm2.61)$ & DAG-GNN $(97.33\pm112.08)$ & GOLEM $(27.69\pm3.23)$ & GOLEM $(26.79\pm1.62)$\\
       & NOTEARS $(3.20\pm1.62)$ & GOLEM $(26.57\pm1.69)$ & RCD $(4.24\pm10.83)$ & RCD $(11.68\pm24.02)$\\
       & ES $(0.61\pm0.27)$ & NOTEARS $(4.34\pm4.49)$ & NOTEARS $(0.66\pm0.28)$ & NOTEARS $(0.70\pm0.47)$\\
       & GES $(0.54\pm0.16)$ & ES $(0.70\pm0.33)$ & GES $(0.39\pm0.17)$ & GES $(0.33\pm0.17)$\\
       & ICALiNGAM $(0.30\pm0.16)$& GES $(0.49\pm0.16)$ & ES $(0.35\pm0.21)$ & ES $(0.29\pm0.20)$\\
       & DirectLiNGAM $(0.10\pm0.05)$& DirectLiNGAM $(0.10\pm0.05)$ & ICALiNGAM $(0.22\pm0.09)$ & ICALiNGAM $(0.28\pm0.13)$ \\
       & GRaSP $(0.09\pm0.04)$& GRaSP $(0.10\pm0.05)$ & DirectLiNGAM $(0.11\pm0.05)$ & DirectLiNGAM $(0.10\pm0.05)$\\
       & PC $(0.09\pm0.05)$& PC $(0.09\pm0.07)$ & PC $(0.05\pm0.04)$ & PC $(0.06\pm0.08)$\\
       & RCD $(0.08\pm0.07)$& FCI $(0.08\pm0.02)$ & FCI $(0.05\pm0.03)$ & FCI $(0.06\pm0.08)$\\
       & FCI $(0.07\pm0.02)$& ICALiNGAM $(0.04\pm0.03)$ & GRaSP $(0.04\pm0.03)$& GRaSP $(0.04\pm0.04)$\\
       \bottomrule
    \end{tabular}}
  \end{center}
\end{table}

Considering time cost of pairwise algorithms, RECI, PWGC, and IGCI stand out as the exemplars of efficiency across all datasets, showcasing a runtime nearly one order of magnitude lower than that of other algorithms.

For instantaneous and time-delay causality, DYNOTEARS, PCMCI, VARLiNGAM algorithm are preferred. 
In contrast, the NeuralGC algorithm exhibits the most prolonged computational execution time, exceeding that of the best algorithm by nearly three orders of magnitude.

When considering i.i.d. data, FCI, ICALiNGAM, and GRaSP are the preferred choices.
In contrast, the least efficient algorithms, GraNDAG and CORL, have runtime that are thousands of times longer than the most efficient ones.

\subsubsection{Recommendation Algorithms}
It is important to note that the ranking table for each data category is obtained by calculating the average value of the results run on 15 datasets. 
\textbf{To answer RQ 1, by sorting out the experimental findings of these causal discovery algorithms, we select the top three with the best average value under each evaluation metric as our recommended algorithms, as shown in Table \ref{tab:recommendation}.}

\begin{table}[htbp]
  \begin{center}
    \caption{Recommendation Algorithms Table}
    \scalebox{0.55}{
    \label{tab:recommendation}
    \begin{tabular}{lcccccr}
    \toprule
        & &\textbf{F1 score}& \textbf{AUROC} & \textbf{FPR} & \textbf{SHD} & \textbf{Run Time (s)}\\
        \hline
        \textbf{Pairwise} &\textbf{linear} & CCM & CCM & ANM & CCM & RECI\\
        & & PAI & PAI (L) & CCM & PAI & PWGC\\
    & & PWGC & CDS & CDS (S) & PWGC & IGCI\\
        \cline{2-7}
        &\textbf{nonlinear} & PAI & PAI & PAI & PAI & RECI\\
        & & CDS & CDS & CDS & CDS & PWGC\\
    & & CCM & CCM & ANM & CCM & IGCI\\
    \hline
    \textbf{Instantaneous} &\textbf{linear + Gaussian} & oCSE & oCSE & oCSE & oCSE & VARLiNGAM\\
    & & \underline{PCMCI} & \underline{PCMCI} & \underline{PCMCI} & \underline{PCMCI} & \underline{PCMCI}\\
    & & VARLiNGAM & NeuralGC & tsFCI (S) & VARLiNGAM (S) & DYNOTEARS\\
    \cline{2-7}
        & \textbf{linear + non-Gaussian} & \underline{VARLiNGAM} & \underline{VARLiNGAM} & \underline{VARLiNGAM} & \underline{VARLiNGAM} & \underline{VARLiNGAM}\\
        & & oCSE & oCSE & oCSE & oCSE & \underline{PCMCI}\\
    & & \underline{PCMCI} & \underline{PCMCI} & \underline{PCMCI} & \underline{PCMCI} & DYNOTEARS\\
        \cline{2-7}
        &\textbf{nonlinear + Gaussian} & NeuralGC & NeuralGC& oCSE& \underline{PCMCI} & VARLiNGAM\\
        & & tsFCI (S) & oCSE & \underline{PCMCI} & oCSE & \underline{PCMCI}\\
    & & \underline{PCMCI} (L) & \underline{PCMCI} & tsFCI & VARLiNGAM (S) & DYNOTEARS\\
        \cline{2-7}
        &\textbf{nonlinear + non-Gaussian} & NeuralGC & NeuralGC & oCSE & oCSE & VARLiNGAM\\
        & & VARLiNGAM (S) & tsFCI (L) & PCMCI & VARLiNGAM & PCMCI\\
    & & tsFCI & oCSE & tsFCI & PCMCI & DYNOTEARS\\
    \hline
    \textbf{Time-delay} &\textbf{linear + Gaussian} & \underline{VARLiNGAM} (L) & MVGC & MVGC & \underline{VARLiNGAM} (L) & DYNOTEARS\\
    & & MVGC (S) & \underline{VARLiNGAM} & \underline{VARLiNGAM} & MVGC & \underline{VARLiNGAM}\\
    & & DYNOTEARS & DYNOTEARS & TCDF (L) & PCMCI (S) & PCMCI\\
    & & & & \textcolor{gray}{oCSE (S)} & & \\
    \cline{2-7}
        & \textbf{linear + non-Gaussian} & PCMCI& MVGC& MVGC & \underline{VARLiNGAM} & DYNOTEARS\\
        & & \underline{VARLiNGAM} & \underline{VARLiNGAM} & \underline{VARLiNGAM} & PCMCI & \underline{VARLiNGAM}\\
    & & DYNOTEARS & PCMCI & TCDF (L) & MVGC (S) & PCMCI\\
    & & & & \textcolor{gray}{oCSE (S)} & & \\
        \cline{2-7}
        &\textbf{nonlinear + Gaussian} & PCMCI& PCMCI& MVGC & PCMCI & VARLiNGAM\\
        & & DYNOTEARS & MVGC & VARLiNGAM (L) & DYNOTEARS & DYNOTEARS\\
    & & VARLiNGAM & DYNOTEARS & oCSE (S) & MVGC (S) & PCMCI\\
        \cline{2-7}
        &\textbf{nonlinear + non-Gaussian} & PCMCI & PCMCI & MVGC & PCMCI & VARLiNGAM\\
        & & NeuralGC & MVGC & oCSE (S) & MVGC & DYNOTEARS\\
    & & DYNOTEARS (S) & NeuralGC & VARLiNGAM (L) & oCSE & PCMCI\\
    & & & & & \textcolor{gray}{NeuralGC (L)} & \\
    \hline
    \textbf{i.i.d.} &\textbf{linear + Gaussian} & GOLEM & GOLEM & GOLEM & GOLEM & FCI\\
    & & CORL & CORL & CORL & CORL & RCD\\
    & & DAG-GNN & DAG-GNN & NOTEARS & DAG-GNN & PC\\
    & & & & \textcolor{gray}{DAG-GNN (L)} & \textcolor{gray}{DAG-GNN (L)}&\\
    \cline{2-7}
        & \textbf{linear + non-Gaussian} & DirectLiNGAM & DirectLiNGAM & DirectLiNGAM & DirectLiNGAM & ICALiNGAM\\
        & & CORL & CORL & CORL & CORL & FCI\\
    & & GOLEM & GOLEM & ICALiNGAM & ICALiNGAM & PC\\
        \cline{2-7}
        &\textbf{nonlinear + Gaussian} & GraNDAG (L) & CORL & NOTEARS & CORL & GRaSP\\
        & & CORL & NOTEARS & DAG-GNN & ICALiNGAM & FCI\\
    & & GES & GraNDAG (L) & ICALiNGAM & NOTEARS & PC\\
    & & & & \textcolor{gray}{GRaNDAG (L)} & \textcolor{gray}{GRaNDAG (L)} & \\
        \cline{2-7}
        &\textbf{nonlinear + non-Gaussian} & CORL & CORL & NOTEARS & CORL & GRaSP\\
        & & GOLEM & GOLEM & DAG-GNN & GOLEM & FCI\\
    & & GraNDAG & DAG-GNN & ICALiNGAM & ICALiNGAM & PC\\
    \bottomrule
\end{tabular}}
  \end{center}
\end{table}

Note that if the optimal algorithm for a specific data size is not among the top three in average ranking, we will supplement it with gray font to ensure that Table \ref{tab:recommendation} covers all possible scenarios as much as possible.
``S” indicates that the algorithm is more suitable on small sample length ($L<1000$), while ``L” indicates that the algorithm is more suitable on large sample length ($L>1000$).

\subsection{Answer to RQ2: Real-World Applicability}
\label{chp:Answer_to_RQ2}

\begin{figure}[htbp]
    \centering
    \includegraphics[scale=0.45]{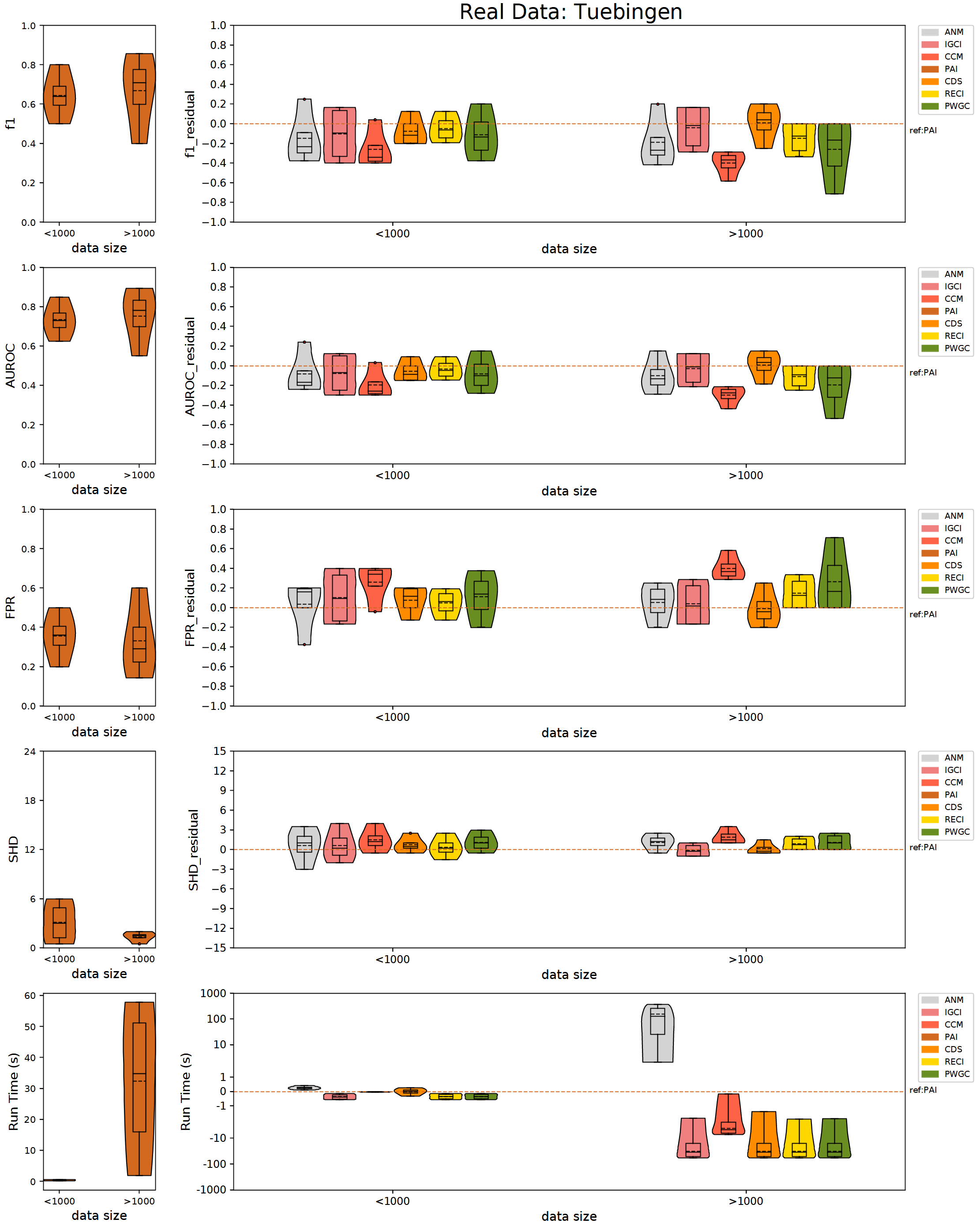}
    \caption{Evaluation of algorithms (ANM, IGCI, CCM, PAI, CDS, RECI, PWGC) on real datasets: Tuebingen}
    \label{fig:real_tuebingen}
    \end{figure}

We first tested the real-world dataset, Tuebingen, which comprises 100 pairs of causal relationships within a stationary time series framework characterized by nonlinearity and non-Gaussian attributes.
The time length spans from 94 to 16,382 time points. 
Leveraging the insights posited in Section \ref{chp:Answer_to_RQ1}, it is deduced that the PAI, CDS, CCM algorithms are the preeminent choices under metrics such as F1 score, AUROC, and SHD; PAI, CDS, ANM are recommended for FPR metric.
Additionally, RECI, PWGC, IGCI algorithm are identified as the most efficient. 
The dataset consists of one hundred instances of causal pairs, which we divided based on their temporal extent: those exceeding 1000 time points were classified as ``large datasets'' and those below 1000 time points were categorized as ``small datasets''. 
Subsequently, an exhaustive execution of all algorithms on this real dataset was conducted, resulting in Figure \ref{fig:real_tuebingen}.

Observation of the graph reveals a clear pattern: using the PAI algorithm as the benchmark, except for CDS, the average F1 and AUROC metrics of the other algorithms consistently reside beneath the horizontal baseline, while the metrics of FPR and SHD exhibit values surpassing those of PAI. 
This collective trend signifies that PAI demonstrates superiority as the optimal algorithm across these four evaluative metrics on ``small datasets'', while CDS performs better on ``large datasets''.

Moreover, when temporal considerations are factored in, the violin plot corresponding to PECI, IGCI, PWGC, CCM algorithms are prominently positioned beneath the horizontal reference line. 
This distinctive placement underscores that RECI holds the lowest time complexity.

These findings align with the algorithmic recommendations derived from experiments on the authentic dataset and corroborate the deductions drawn based on the outcomes expounded in Section \ref{chp:Answer_to_RQ3}. 
This congruence augments our confidence in extending the theoretical framework to real-world datasets, thereby validating our theoretical assertions and demonstrating their practical applicability.

The second real dataset pertains to functional Magnetic Resonance Imaging (fMRI), comprising 28 sets of multivariate time series.
A subset of data that failed to meet criteria associated with causal sufficiency was omitted, resulting in the examination of 27 datasets. 
This dataset is characterized by nonlinearity and Gaussian attributes, emblematic of time-delay causal causality with $lag=1$. 
Among the dataset constituents, 21 sets comprise fewer than 1000 data points, while the remaining six sets exceed this threshold.
Each dataset includes 5, 10, or 15 time series variables. 
Guided by these salient attributes, we predict that one of PCMCI, DYNOTEARS, VARLiNGAM algorithms will exhibit optimal performance under F1 score, whereas PCMCI, MVGC, DYNOTEARS algorithms will attain primacy in terms of AUROC.
For FPR, MVGC is the most recommended algorithm since it performs well on datasets of all sizes, while VARLiNGAM is only recommended on large sample sizes and oCSE is only recommended on small sample sizes.
Considering SHD, PCMCI, DYNOTEARS, and MVGC are recommended algorithms, with MVGC being more suitable for small sample sizes.
When taking into account the time cost, we recommend PCMCI, VARLiNGAM, and DYNOTEARS algorithms.

\begin{figure}[htbp]
    \centering
    \includegraphics[scale=0.42]{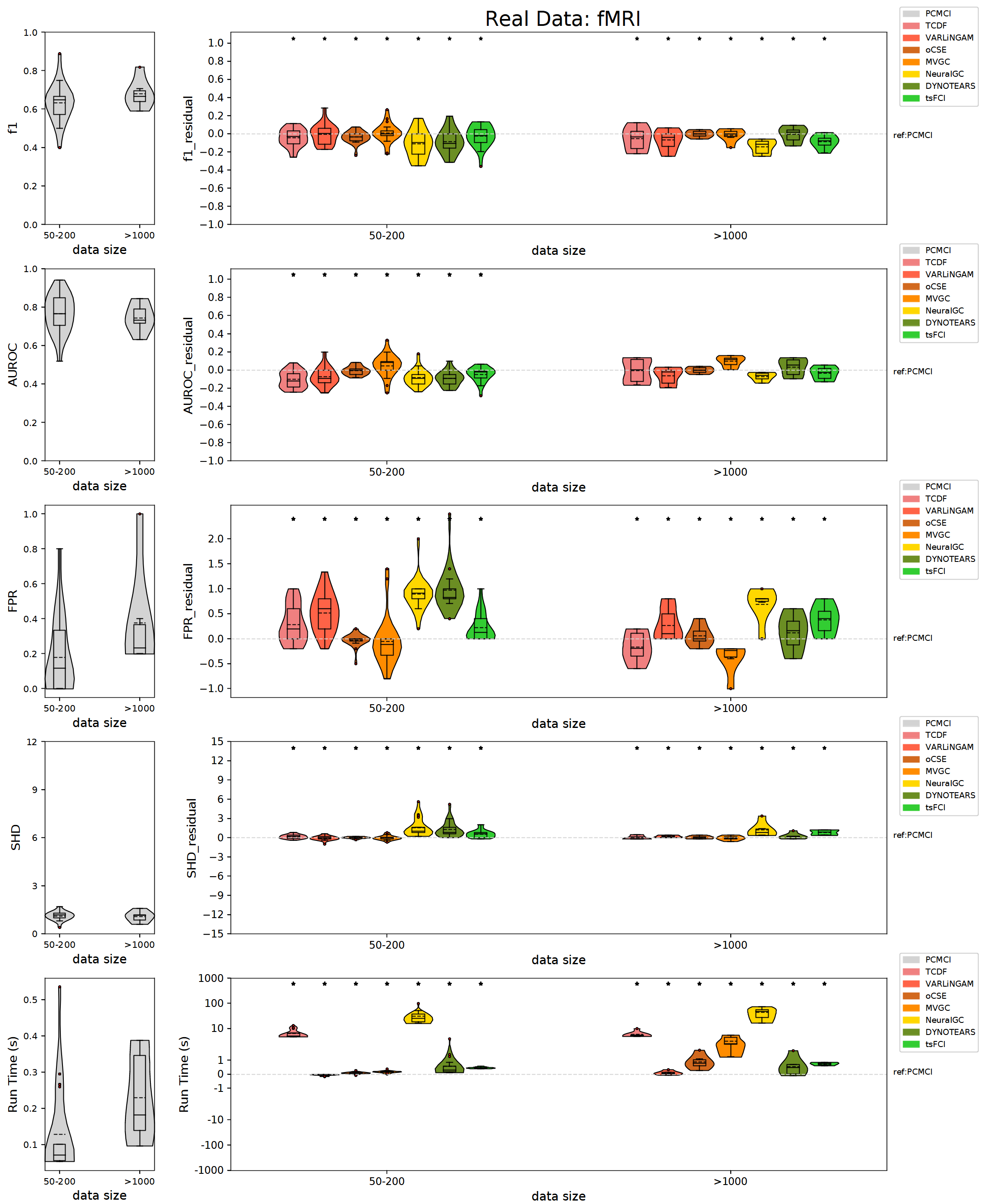}
    \caption{Evaluation of algorithms (PCMCI, TCDF, VARLiNGAM, oCSE, MVGC, NeuralGC, DYNOTEARS, tsFCI) on real datasets: fMRI}
    \label{real_fMRI}
    \end{figure} 
    
To ensure a comprehensive assessment of algorithmic efficacy, we evaluated both instantaneous causal discovery algorithms and time-delay algorithms on the fMRI dataset. 
A total of nine distinct algorithms were compared. 
Analysis of Figure \ref{real_fMRI} reveals that the reference algorithm, PCMCI, achieves the highest values under the F1 metric and the lowest values for Run time. 
Regarding AUROC, FPR and SHD, it can be clearly seen from the graph that MVGC is the best performing one, which is included in our recommendations. This means that MVGC has good stability and low error rate.
This alignment with our earlier assessments further bolsters the robustness of our conclusions.\\

\textbf{To answer RQ 2, the optimal algorithm identified through experimental testing on the two datasets aligns with the one determined according to the insights in Section \ref{chp:Answer_to_RQ1}. 
This consistency underscores the reliability of the theory derived from both synthetic and real-world datasets.}

\subsection{Answer to RQ3: Generalization to Unknown Datasets}
\label{chp:Answer_to_RQ3}

In Section \ref{chp:Answer_to_RQ3.1}, a metadata detection program was designed to extract data features. 
Subsequently, in Section \ref{chp:Answer_to_RQ3.2}, our recommendation program was tested on various datasets to verify its consistency with the algorithm test results.

\subsubsection{Answer to RQ 3.1: Metadata Recognition for Algorithm Selection}
\label{chp:Answer_to_RQ3.1}

Given that our prior analyses focused on causality types, linearity among series, and noise distribution, capturing these pivotal attributes within an unknown dataset is crucial for the project's universality and practical applicability.

The first task of metadata detection is identifying temporal lags within variables. We employ the Time Lag Cross Correlation (TLCC) technique to accomplish this.
TLCC is measured by gradually shifting a time series vector and repeatedly calculating the correlation between two signals. 
Identifying correlation maxima facilitates the ascertainment of inter-variable temporal lag. 
Specifically, a zero lag denotes an instantaneous causal association, whereas a non-zero lag signifies a time-delay causality. If no lag is detected, the dataset is classified as i.i.d. data.

Subsequently, identifying the noise distribution is requisite. 
We employ concurrent evaluative methodologies encompassing the Shapiro-Wilk, Kolmogorov-Smirnov, and Anderson-Darling tests. 
These tests collectively serve to discern the presence of Gaussian noise in the data. The following criteria serve as benchmarks:

1. The Shapiro-Wilk test's computed p-value surpasses the significance threshold of 0.05.

2. The p-value resulting from the Kolmogorov-Smirnov test exceeds 0.05.

3. The p-value derived from the Anderson-Darling test remains below its critical threshold.

Fulfillment of these conditions collectively allows for the inference of Gaussian noise as the prevailing noise type characterizing the dataset.

Lastly, a crucial inquiry involves ascertaining potential linear interdependence among variables. 
To address this, a linear regression framework is applied to every pair of variables.
Subsequently, the coefficient of determination (R-squared) is derived to gauge the efficacy of the model fit. 
A predetermined threshold of 0.5 is set for assessment. 
If the computed R-squared value surpasses this threshold, it signifies the presence of a discernible linear relationship between the variables. 
Conversely, an R-squared value below the threshold implies suboptimal alignment with the linear regression framework, indicating the absence of a linear relationship between the implicated variables.

To comprehensively appraise the previously delineated feature extraction procedures, we conducted metadata detection experiments on 100 datasets, with causality types, linear relations, noise distributions, and time lengths randomly generated. 
This evaluation was accomplished by computing the accuracy of judgments for the three metadata categories. 
We conducted ten trials, yielding comprehensive average and standard deviation metrics, as detailed in Table \ref{tab:metadata}. 

\begin{table}[ht]
  \begin{center}
    \caption{Accuracy of Metadata Extraction}
    \scalebox{0.9}{
    \label{tab:metadata}
    \begin{tabular}{lccc}
    \toprule
        & \textbf{Causality Type} & \textbf{Linear Relation} &\textbf{Gaussian Noise}\\
    \hline
   \textbf{Accuracy} &$0.86\pm0.01$  &$0.81\pm0.07$ &$0.82\pm0.09$ \\
   \bottomrule
    \end{tabular}}
  \end{center}
\end{table}
\textbf{To answer RQ 3.1, Table \ref{tab:metadata} shows that these three metadata can be extracted precisely, with accuracies over 75\% and standard deviations of no more than 0.1, indicating that the judgment program is also relatively stable.}

\subsubsection{Answer to RQ 3.2: Practical Recommendations for Users}
\label{chp:Answer_to_RQ3.2}

LUCAS and Sachs was selected as the test dataset. The extracted metadata, based on the program described in the previous section, along with the corresponding recommendation algorithms provided by Table \ref{tab:recommendation}, are listed in Table \ref{tab: meta_lucas_sachs}.

\begin{table}[ht]
  \begin{center}
    \caption{Extracted Metadata of LUCAS and Sachs}
    \scalebox{0.6}{
    \label{tab: meta_lucas_sachs}
    \begin{tabular}{lccr}
    \toprule
     & & \textbf{LUCAS} & \textbf{Sachs}\\
     \cline{2-4}
        \textbf{Metadata} & \textbf{Data size} & (500, 11) & (7466, 11)\\
        & \textbf{Dependency funcs} & nonlinear & linear\\
        & \textbf{Noises distributions} & Gaussian & non-Gaussian\\
    \cline{2-4}
   \textbf{Recommendation algorithms} & \textbf{F1} & GraNDAG; CORL; GES & DirectLiNGAM; CORL; GOLEM\\
   & \textbf{AUROC} & CORL; NOTEARS; GraNDAG & DirectLiNGAM; CORL; GOLEM\\
   & \textbf{FPR} & NOTEARS; DAG-GNN; ICALiNGAM & DirectLiNGAM; CORL; ICALiNGAM\\
   & \textbf{SHD} & CORL; ICALiNGAM; NOTEARS & DirectLiNGAM; CORL; ICALiNGAM\\
   & \textbf{Runtime} & GRaSP; FCI; PC & ICALiNGAM; FCI; PC\\
   \bottomrule
    \end{tabular}}
  \end{center}
\end{table}

13 algorithms for i.i.d. data were tested on the LUCAS dataset and 12 algorithms were tested on Sachs. The RCD algorithm is not considered for Sachs due to its long running time (run time $>15 minutes$).
The results, presented in Table \ref{tab:lucas} illustrates that the optimal algorithm for each metric is within previous recommendation range, which verifies the effectiveness of our recommendation program.
To illustrate clearly, the algorithm we recommend is underlined, and the optimal algorithm obtained from horizontal testing is highlighted in bold.

\begin{table}[h!]
  \begin{center}
    \caption{Test Result of LUCAS and Sachs}
    \scalebox{0.6}{
    \label{tab:lucas}
    \begin{tabular}{lrrrrr}
    \toprule
 & \textbf{F1$\uparrow$} & \textbf{AUROC$\uparrow$} & \textbf{FPR$\downarrow$} & \textbf{SHD$\downarrow$} & \textbf{Runtime$\downarrow$}\\
    \hline
       \textbf{LUCAS} &\textbf{\underline{GraNDAG (0.70)}} & \textbf{\underline{NOTEARS (0.75)}}  & RCD (0.47) & RCD (30.0) & CORL (257.80)\\
       & NOTEARS (0.70) & \textbf{\underline{GraNDAG (0.75)}} & DirectLiNGAM (0.33) & DirectLiNGAM (16.0) & GraNDAG (233.52) \\
      &  \underline{GES (0.69)} & GES (0.72) & ES (0.26) & \underline{CORL (13.0)} & DAG-GNN (44.29)\\
      &  DAG-GNN (0.69) & DAG-GNN (0.72) & GOLEM (0.23) & ES (13.0) & GOLEM (32.61)\\
      &  FCI (0.67) & FCI (0.68) & CORL (0.23) & GOLEM (11.0) & RCD (12.19)\\
      &  GRaSP (0.67) & GRaSP (0.68) & PC (0.21) & PC (11.0) & NOTEARS (0.83)\\
      &  GOLEM (0.63) & GOLEM (0.67) & GES (0.16) & \underline{ICALiNGAM (10.0)} & ES (0.71)\\
      & ICALiNGAM (0.59) & ICALiNGAM (0.62) & \underline{ICALiNGAM (0.16)} & GES (8.0) & GES (0.42)\\
      &  ES (0.56) & ES (0.61) & \underline{DAG-GNN (0.16)} & GRaSP (8.0) & DirectLiNGAM (0.10)\\
      &  \underline{CORL (0.53)} & \underline{CORL (0.57)} & GRaSP (0.09) & DAG-GNN (8.0) & \underline{FCI (0.05)}\\
      &  PC (0.48) & RCD (0.52)& FCI (0.09) & FCI (7.0) & ICALiNGAM (0.05)\\
      &  RCD (0.31) & PC (0.52) & GraNDAG (0.02) & GraNDAG (6.0) & \underline{GRaSP (0.05)}\\
      &  DirectLiNGAM (0.21) & DirectLiNGAM (0.25) & \textbf{\underline{NOTEARS (0.02)}} & \textbf{\underline{NOTEARS (5.0)}} & \textbf{\underline{PC (0.05)}}\\
       \hline
      \textbf{Sachs} &  \textbf{\underline{CORL (0.30)}} & \textbf{\underline{CORL (0.35)}} & ES (0.84) & ES (37.0) & DAG-GNN (814.25)\\
       & NOTEARS (0.26) & \underline{DirectLiNGAM (0.32)} & DAG-GNN (0.70) & DAG-GNN (32.0) & GraNDAG (236.29) \\
      &  GRaSP (0.25) & GRaSP (0.32) & GRaSP (0.62) & FCI (31.0) & CORL (95.95)\\
      &  DAG-GNN (0.24) & NOTEARS (0.32) & PC (0.57) & PC (31.0) & GOLEM (68.19)\\
      &  \underline{GOLEM (0.24)} & DAG-GNN (0.31) & FCI (0.49) & GRaSP (30.0) & NOTEARS (25.67)\\
      &  ICALiNGAM (0.23) & \underline{GOLEM (0.29)} & \underline{CORL (0.43)} & \underline{CORL (25.0)} & ES (4.76)\\
      &  PC (0.22) & ICALiNGAM (0.29) & NOTEARS (0.41) & GES (25.0) & GES (1.14)\\
      &  FCI (0.19) & PC (0.29) & \underline{ICALiNGAM (0.35)} & NOTEARS (25.0) & \underline{PC (1.10)}\\
      &  ES (0.19) & ES (0.26) & GOLEM (0.32) & GOLEM (24.0) & \underline{FCI (0.68)}\\
       & \underline{DirectLiNGAM (0.17)} & FCI (0.26) & GES (0.19) & \underline{ICALiNGAM (24.0)} & DirectLiNGAM (0.38)\\
      &  GES (0.16) & GES(0.22)& GraNDAG (0.14) & GraNDAG (20.0) & GRaSP (0.35)\\
      &  GraNDAG (0.08) & GraNDAG (0.18) & \textbf{\underline{DirectLiNGAM (0.08)}} & \textbf{\underline{DirectLiNGAM (17.0)}} & \textbf{\underline{ICALiNGAM (0.09)}}\\
    \bottomrule
    \end{tabular}}
  \end{center}
\end{table}

\textbf{To answer RQ 3.2, the case study of these two real datasets indicates that the optimal algorithms obtained from the experiment are included in our recommended algorithms. This means that users can quickly find the most suitable algorithm based on extracted metadatas and save computing power.}

\section{Threats to Validity}
\label{chp:Threats_to_Validity}

\subsection{Threats to Internal Validity}

\subsubsection{Correctness of The Codes}

A potential threat to the study revolves around the accuracy of the employed codebase. Despite rigorous testing and validation procedures, the complexity of algorithmic implementations and potential oversights during the coding process may give rise to errors. Variations in the code may inadvertently influence the outcomes, thereby introducing a threat to the study's internal validity.

To address this, the study highlights the ongoing commitment to code review and validation, emphasizing the significance of code quality in ensuring the reliability of the study's outcomes. Specifically, we conducted validation tests on each algorithm to ensure its correctness.

\subsubsection{Implementation of Control Variables}

In light of the comparative analytical methodology employed within the experimental framework of this project, the precision in governing variables assumes paramount importance due to its direct influence on the ensuing output. If multiple testing algorithms are administered across disparate datasets, the potential for consequential impact stemming from inherent dataset dissimilarities becomes salient. Such variations could undermine the cogency of the ultimate conclusions drawn from the study.

To mitigate this, distinct algorithms were executed on a singular dataset, thereby facilitating establishing a controlled environment wherein the dataset variable maintains consistency. Notably, our program was executed holistically on a singular server infrastructure, ensuring uniformity in the controlling variable of the computing capacity and thereby rendering temporal comparisons feasible. In the interest of upholding the validation of the comparative analysis, synthesized data types of specific time lengths were all derived from one DAG structure. This strategic alignment eradicated any potential ambiguities stemming from disparate ground truth, consequently enhancing the reliability and interpretability of the results.

\subsection{Threats to external validity}

\subsubsection{Metadata selection bias}

The external validity of this study could be threatened by metadata selection bias. In light of our methodology, which exclusively constructs testing data predicated upon established causal linkages, noise distribution characteristics, and dataset dimensions, it is prudent to acknowledge the potential bias that crucial time series' metadata might not be exhaustively contained. Consequently, there remains a prospect for more influential data attributes that could threaten the comprehensiveness and external validity of the experimental outcomes.

In order to mitigate the potential influence of this factor on the experimental validity, two real-world datasets were examined, and it was ascertained that the empirical findings obtained from these real-world datasets exhibit congruence with those derived from the synthetic datasets. This correspondence serves as an indirect validation of the judicious selection of metadata, affirming its representativeness.

\subsubsection{Temporal Validity}

Considering the incessant evolution characterizing algorithms, it is plausible that our research might not encompass forthcoming optimizations, thereby engendering temporal constraints on the experiment's conclusions. This scenario presents an external threat to the generalizability of research endeavours.

To attenuate this influence to the greatest extent possible, a mitigation strategy has been adopted wherein algorithms are encoded following a standardized input-output paradigm, thereby creating an extensible code library. This design contributes to the seamless integration of forthcoming algorithmic enhancements into the repository, facilitating their inclusion within the testing framework. This holistic program design underscores commendable operational feasibility and assimilation capacity.

\section{Conclusions}
\label{chp:conclusion}

This section will summarize the entire project and analyze potential future work from three aspects.

\subsection{Conclusions}
\label{chp:Conclusions_and_Summary}

This work conducted comprehensive research on causal discovery in time series, introducing the research topic's relevant background, importance, and literature system. We introduced six types of methods and sorted out over 20 algorithms.

In terms of experiments, our task is to explore the optimal algorithms in different application scenarios. We successfully summarised recommendation algorithms corresponding to 16 data types through comparative experiments and verified the applicability of these insights through testing on two real-world datasets. Furthermore, we extended the practicality of this discovery to the unknown dataset through metadata extraction technology, with the testing accuracy up to $80\%$.

Finally, we discussed the experimental results and related them to existing literature, and listed the threats that could affect the validity of the results. By discussing these two aspects, the findings of this project are made more reliable and vivid.

Given the above content, this project has systematically completed all research objectives, thoroughly analyzing causal discovery for time series from theoretical and practical aspects. We provided practical and effective algorithm recommendations from users' perspectives, filling the research gaps in previous articles and providing specific reference values for future algorithm research.

\subsection{Discussion on Future Research Directions}
\label{chp:Recommendations_for_Further_Work}

Opportunities for enhancement and progress persist in this study. 
Based on the discussion section, a trajectory for future investigations can be charted across three principal dimensions to improve the algorithm guideline system. 
The initial facet involves considering a broader range of data types, such as high-dimensional datasets or time series incorporating latent variables.

The second avenue of progression entails the extension of algorithmic testing. 
Due to temporal constraints in this project, certain time-intensive algorithms were omitted from the assessment. 
It is advisable to incorporate these additional algorithms in future endeavors to achieve a more exhaustive and comprehensive comparative analysis of algorithmic efficacy.

The final and pivotal facet pertains to enhancing the precision of metadata identification. 
To achieve this objective, the adoption of more sophisticated data processing techniques is warranted. 
Notably, the integration of machine learning methodologies should be considered.
This approach would yield more precise determinations of unknown data characteristics, particularly in distinguishing between time-delay and instantaneous causality.

Upon the successful execution of the proposed endeavors, users will be furnished with a meticulous causal discovery algorithm recommendation service. 
This service would facilitate the expedient identification of optimal algorithms for arbitrary datasets, thereby markedly curtailing the duration of trial-and-error procedures and minimizing computational resource consumption. 
This work effectively addresses the intricate conundrum associated with algorithm selection for causal discovery, which holds significant research and practical value.

Besides, it is important to discuss research directions related to existing research gaps in the field, mainly focusing on three aspects:
\begin{itemize}
  \item \textit{Datasets and algorithms}:
One significant issue is the limited availability of real datasets with ground truth, which restricts algorithm experiments to only a few real datasets.
Therefore, developing a broader range of real datasets is crucial for benchmarking causal discovery algorihthms.
Additionally, in the development of new algorithms, two-step approaches show significant potential. For instance, integrating causal discovery methods with neural networks in different stages could overcome the limitations relying on a single method.
  \item \textit{Algorithm selector}:
A critical research direction is the joint analysis of data features and causal graph structures.
This effort aims to develop a comprehensive algorithm selector \citep{runge2023} capable of addressing all possible data scenarios, thereby assisting users in handling various complex situations.
  \item \textit{Causality or Association?}
Understanding the fundamental differences between causality and regression is crucial.
Conducting empirical research to clarify the quantitative differences between causal algorithms and association-based methods will provide valuable insights for advancing causal research and statistical techniques.
This understanding is also essential for evaluating the impact that causal learning could bring to the field of explainable AI \citep{montavon2018methods}.
\end{itemize}

Additionally, future work could explore the application of causal discovery methods to verify the results of data augmentation \citep{gao2023data}.
For example, these methods could be used to ascertain whether newly generated data retains the causality of the original data.
This verification process would enhance the reliability and validity of augmented datasets, ensuring that the fundamental causal relationships are preserved.



\bibliography{sample}

\begin{thebibliography}{179}
\providecommand{\natexlab}[1]{#1}
\providecommand{\url}[1]{\texttt{#1}}
\expandafter\ifx\csname urlstyle\endcsname\relax
  \providecommand{\doi}[1]{doi: #1}\else
  \providecommand{\doi}{doi: \begingroup \urlstyle{rm}\Url}\fi

\bibitem[Ahmed et~al.(2020)Ahmed, Tr{\"a}uble, Goyal, Neitz, Bengio, Sch{\"o}lkopf, W{\"u}thrich, and Bauer]{ahmed2020}
Ossama Ahmed, Frederik Tr{\"a}uble, Anirudh Goyal, Alexander Neitz, Yoshua Bengio, Bernhard Sch{\"o}lkopf, Manuel W{\"u}thrich, and Stefan Bauer.
\newblock Causalworld: A robotic manipulation benchmark for causal structure and transfer learning.
\newblock \emph{arXiv preprint arXiv:2010.04296}, 2020.

\bibitem[Aliprantis(2015)]{aliprantis2015distinction}
Dionissi Aliprantis.
\newblock A distinction between causal effects in structural and {R}ubin causal models.
\newblock \emph{SSRN Electronic Journal}, 15-05, 2015.

\bibitem[Andrews et~al.(2023)Andrews, Ramsey, and Sanchez~Romero]{andrews2023fast}
Bryan Andrews, Joseph Ramsey, and Ruben Sanchez~Romero.
\newblock Fast scalable and accurate discovery of dags using the best order score search and grow shrink trees.
\newblock \emph{Advances in Neural Information Processing Systems}, 36:\penalty0 63945--63956, 2023.

\bibitem[Arize(1993)]{arize1993}
Augustine~C Arize.
\newblock Determinants of income velocity in the united kingdom: multivariate granger causality.
\newblock \emph{The American Economist}, 37\penalty0 (2):\penalty0 40--45, 1993.

\bibitem[Assaad(2022)]{cdts2022}
Charles~K Assaad.
\newblock {causal\_discovery\_for\_time\_series}, 2022.
\newblock URL \url{https://github.com/ckassaad/causal_discovery_for_time_series.git}.

\bibitem[Assaad et~al.(2022)Assaad, Devijver, and Gaussier]{Assaad2022}
Charles~K Assaad, Emilie Devijver, and Eric Gaussier.
\newblock Survey and evaluation of causal discovery methods for time series.
\newblock \emph{Journal of Artificial Intelligence Research}, 73:\penalty0 767--819, 2022.

\bibitem[Assaad et~al.(2023)Assaad, Ez-Zejjari, and Zan]{assaad2023root}
Charles~K Assaad, Imad Ez-Zejjari, and Lei Zan.
\newblock Root cause identification for collective anomalies in time series given an acyclic summary causal graph with loops.
\newblock In \emph{International Conference on Artificial Intelligence and Statistics}, pages 8395--8404. PMLR, 2023.

\bibitem[Assaad et~al.(2021)Assaad, Devijver, Gaussier, and Ait-Bachir]{assaad2021mixed}
Karim Assaad, Emilie Devijver, Eric Gaussier, and Ali Ait-Bachir.
\newblock A mixed noise and constraint-based approach to causal inference in time series.
\newblock In \emph{Machine Learning and Knowledge Discovery in Databases. Research Track: European Conference, ECML PKDD 2021, Bilbao, Spain, September 13--17, 2021, Proceedings, Part I 21}, pages 453--468. Springer, 2021.

\bibitem[Asuncion and Newman(2007)]{asuncion2007}
Arthur Asuncion and David Newman.
\newblock Uci machine learning repository.
\newblock \emph{Irvine, CA, USA}, 2007.

\bibitem[Bareinboim and Pearl(2016)]{bareinboim2016causal}
Elias Bareinboim and Judea Pearl.
\newblock Causal inference and the data-fusion problem.
\newblock \emph{Proceedings of the National Academy of Sciences}, 113\penalty0 (27):\penalty0 7345--7352, 2016.

\bibitem[Barrera and Miljkovic(2022)]{barrera2022link}
Emiliano~Lopez Barrera and Dragan Miljkovic.
\newblock The link between the two epidemics provides an opportunity to remedy obesity while dealing with covid-19.
\newblock \emph{Journal of Policy Modeling}, 44\penalty0 (2):\penalty0 280--297, 2022.

\bibitem[Barrett et~al.(2010)Barrett, Barnett, and Seth]{barrett2010multivariate}
Adam~B Barrett, Lionel Barnett, and Anil~K Seth.
\newblock Multivariate granger causality and generalized variance.
\newblock \emph{Physical Review E—Statistical, Nonlinear, and Soft Matter Physics}, 81\penalty0 (4):\penalty0 041907, 2010.

\bibitem[Beaumont et~al.(2021)Beaumont, Horsburgh, Pilgerstorfer, Droth, Oentaryo, Ler, Nguyen, Ferreira, Patel, and Leong]{Beaumont_CausalNex_2021}
Paul Beaumont, Ben Horsburgh, Philip Pilgerstorfer, Angel Droth, Richard Oentaryo, Steven Ler, Hiep Nguyen, Gabriel~Azevedo Ferreira, Zain Patel, and Wesley Leong.
\newblock {CausalNex (Version 0.12.1)}, October 2021.
\newblock URL \url{https://github.com/quantumblacklabs/causalnex}.

\bibitem[Beinlich et~al.(1989)Beinlich, Suermondt, Chavez, and Cooper]{beinlich1989alarm}
Ingo~A Beinlich, Henri~Jacques Suermondt, R~Martin Chavez, and Gregory~F Cooper.
\newblock The alarm monitoring system: A case study with two probabilistic inference techniques for belief networks.
\newblock In \emph{AIME 89: Second European Conference on Artificial Intelligence in Medicine, London, August 29th--31st 1989. Proceedings}, pages 247--256. Springer, 1989.

\bibitem[Biswas and Mukherjee(2024)]{biswas2024consistent}
Rahul Biswas and Somabha Mukherjee.
\newblock Consistent causal inference from time series with pc algorithm and its time-aware extension.
\newblock \emph{Statistics and Computing}, 34\penalty0 (1):\penalty0 14, 2024.

\bibitem[Bl{\"o}baum et~al.(2018)Bl{\"o}baum, Janzing, Washio, Shimizu, and Sch{\"o}lkopf]{blobaum2018cause}
Patrick Bl{\"o}baum, Dominik Janzing, Takashi Washio, Shohei Shimizu, and Bernhard Sch{\"o}lkopf.
\newblock Cause-effect inference by comparing regression errors.
\newblock In \emph{International Conference on Artificial Intelligence and Statistics}, pages 900--909. PMLR, 2018.

\bibitem[Bystrova et~al.(2024)Bystrova, Assaad, Arbel, Devijver, Gaussier, and Thuiller]{bystrova2024causaldiscoverytimeseries}
Daria Bystrova, Charles Assaad, Julyan Arbel, Emilie Devijver, {\'E}ric Gaussier, and Wilfried Thuiller.
\newblock Causal discovery from time series with hybrids of constraint-based and noise-based algorithms.
\newblock \emph{Transactions on Machine Learning Research Journal}, 2024.

\bibitem[Chen et~al.(2004)Chen, Rangarajan, Feng, and Ding]{chen2004}
Yonghong Chen, Govindan Rangarajan, Jianfeng Feng, and Mingzhou Ding.
\newblock Analyzing multiple nonlinear time series with extended granger causality.
\newblock \emph{Physics letters A}, 324\penalty0 (1):\penalty0 26--35, 2004.

\bibitem[Cheng et~al.(2019)Cheng, Guo, and Liu]{cheng2019robust}
Lu~Cheng, Ruocheng Guo, and Huan Liu.
\newblock Robust cyberbullying detection with causal interpretation.
\newblock In \emph{Companion Proceedings of The 2019 World Wide Web Conference}, pages 169--175, 2019.

\bibitem[Cheng et~al.(2022)Cheng, Guo, Moraffah, Sheth, Candan, and Liu]{Cheng2022}
Lu~Cheng, Ruocheng Guo, Raha Moraffah, Paras Sheth, K~Sel{\c{c}}uk Candan, and Huan Liu.
\newblock Evaluation methods and measures for causal learning algorithms.
\newblock \emph{IEEE Transactions on Artificial Intelligence}, 3\penalty0 (6):\penalty0 924--943, 2022.

\bibitem[Chickering(2002{\natexlab{a}})]{chickering2002}
David~Maxwell Chickering.
\newblock Learning equivalence classes of bayesian-network structures.
\newblock \emph{The Journal of Machine Learning Research}, 2:\penalty0 445--498, 2002{\natexlab{a}}.

\bibitem[Chickering(2002{\natexlab{b}})]{chickering2002optimal}
David~Maxwell Chickering.
\newblock Optimal structure identification with greedy search.
\newblock \emph{Journal of machine learning research}, 3\penalty0 (Nov):\penalty0 507--554, 2002{\natexlab{b}}.

\bibitem[Chickering(2020)]{chickering2020statistically}
Max Chickering.
\newblock Statistically efficient greedy equivalence search.
\newblock In \emph{Conference on Uncertainty in Artificial Intelligence}, pages 241--249. Pmlr, 2020.

\bibitem[Chiyohara et~al.(2023)Chiyohara, Furukawa, Noda, Morimoto, and Imamizu]{chiyohara2023proprioceptive}
Shinya Chiyohara, Jun-ichiro Furukawa, Tomoyuki Noda, Jun Morimoto, and Hiroshi Imamizu.
\newblock Proprioceptive short-term memory in passive motor learning.
\newblock \emph{Scientific Reports}, 13\penalty0 (1):\penalty0 20826, 2023.

\bibitem[Colombo et~al.(2012)Colombo, Maathuis, Kalisch, and Richardson]{colombo2012learning}
Diego Colombo, Marloes~H Maathuis, Markus Kalisch, and Thomas~S Richardson.
\newblock Learning high-dimensional directed acyclic graphs with latent and selection variables.
\newblock \emph{The Annals of Statistics}, pages 294--321, 2012.

\bibitem[Colombo et~al.(2014)Colombo, Maathuis, et~al.]{colombo2014order}
Diego Colombo, Marloes~H Maathuis, et~al.
\newblock Order-independent constraint-based causal structure learning.
\newblock \emph{J. Mach. Learn. Res.}, 15\penalty0 (1):\penalty0 3741--3782, 2014.

\bibitem[Driessens and D{\v{z}}eroski(2005)]{driessens2005combining}
Kurt Driessens and Sa{\v{s}}o D{\v{z}}eroski.
\newblock Combining model-based and instance-based learning for first order regression.
\newblock In \emph{Proceedings of the 22nd international conference on Machine learning}, pages 193--200, 2005.

\bibitem[Edinburgh et~al.(2021)Edinburgh, Eglen, and Ercole]{edinburgh2021}
Tom Edinburgh, Stephen~J. Eglen, and Ari Ercole.
\newblock {Causality indices for bivariate time series data: A comparative review of performance}.
\newblock \emph{Chaos: An Interdisciplinary Journal of Nonlinear Science}, 31\penalty0 (8):\penalty0 083111, 08 2021.
\newblock ISSN 1054-1500.
\newblock \doi{10.1063/5.0053519}.
\newblock URL \url{https://doi.org/10.1063/5.0053519}.

\bibitem[Eichler(2012)]{eichler2012causal}
Michael Eichler.
\newblock Causal inference in time series analysis.
\newblock \emph{Causality: Statistical perspectives and applications}, pages 327--354, 2012.

\bibitem[Entner and Hoyer(2010)]{entner2010causal}
Doris Entner and Patrik~O Hoyer.
\newblock On causal discovery from time series data using fci.
\newblock \emph{Probabilistic graphical models}, 16, 2010.

\bibitem[Fang et~al.(2023)Fang, Zhu, Zhang, Liu, Chen, and He]{fang2023low}
Zhuangyan Fang, Shengyu Zhu, Jiji Zhang, Yue Liu, Zhitang Chen, and Yangbo He.
\newblock On low-rank directed acyclic graphs and causal structure learning.
\newblock \emph{IEEE Transactions on Neural Networks and Learning Systems}, 35\penalty0 (4):\penalty0 4924--4937, 2023.

\bibitem[Fonollosa(2019)]{fonollosa2019conditional}
Jos{\'e}~AR Fonollosa.
\newblock Conditional distribution variability measures for causality detection.
\newblock \emph{Cause Effect Pairs in Machine Learning}, pages 339--347, 2019.

\bibitem[Ganguly et~al.(2023)Ganguly, Fazlija, Badar, Fisichella, Sikdar, Schrader, Wallat, Rudra, Koubarakis, Patro, et~al.]{ganguly2023review}
Niloy Ganguly, Dren Fazlija, Maryam Badar, Marco Fisichella, Sandipan Sikdar, Johanna Schrader, Jonas Wallat, Koustav Rudra, Manolis Koubarakis, Gourab~K Patro, et~al.
\newblock A review of the role of causality in developing trustworthy ai systems.
\newblock \emph{arXiv preprint arXiv:2302.06975}, 2023.

\bibitem[Gao et~al.(2023)Gao, Li, and Xu]{gao2023data}
Zijun Gao, Lingbo Li, and Tianhua Xu.
\newblock Data augmentation for time-series classification: An extensive empirical study and comprehensive survey.
\newblock \emph{arXiv preprint arXiv:2310.10060}, 2023.

\bibitem[Garc{\'\i}a-Vel{\'a}zquez et~al.(2020)Garc{\'\i}a-Vel{\'a}zquez, Jokela, and Rosenstr{\"o}m]{garcia2020direction}
Regina Garc{\'\i}a-Vel{\'a}zquez, Markus Jokela, and Tom~Henrik Rosenstr{\"o}m.
\newblock Direction of dependence between specific symptoms of depression: A non-gaussian approach.
\newblock \emph{Clinical Psychological Science}, 8\penalty0 (2):\penalty0 240--251, 2020.

\bibitem[Gelman(2011)]{gelman2011causality}
Andrew Gelman.
\newblock Causality and statistical learning, 2011.

\bibitem[Gerhardus and Runge(2020)]{gerhardus2020high}
Andreas Gerhardus and Jakob Runge.
\newblock High-recall causal discovery for autocorrelated time series with latent confounders.
\newblock \emph{Advances in Neural Information Processing Systems}, 33:\penalty0 12615--12625, 2020.

\bibitem[Geweke(1982)]{geweke1982}
John Geweke.
\newblock Measurement of linear dependence and feedback between multiple time series.
\newblock \emph{Journal of the American statistical association}, 77\penalty0 (378):\penalty0 304--313, 1982.

\bibitem[Gong et~al.(2017)Gong, Zhang, Sch{\"o}lkopf, Glymour, and Tao]{gong2017}
Mingming Gong, Kun Zhang, Bernhard Sch{\"o}lkopf, Clark Glymour, and Dacheng Tao.
\newblock Causal discovery from temporally aggregated time series.
\newblock In \emph{Uncertainty in artificial intelligence: proceedings of the... conference. Conference on Uncertainty in Artificial Intelligence}, volume 2017. NIH Public Access, 2017.

\bibitem[Goudet et~al.(2018)Goudet, Kalainathan, and Caillou]{goudet2018}
Olivier Goudet, Diviyan Kalainathan, and Philippe Caillou.
\newblock Learning functional causal models with generative neural networks.
\newblock \emph{Explainable and interpretable models in computer vision and machine learning}, pages 39--80, 2018.

\bibitem[Granger(1969)]{granger1969}
Clive~WJ Granger.
\newblock Investigating causal relations by econometric models and cross-spectral methods.
\newblock \emph{Econometrica: journal of the Econometric Society}, pages 424--438, 1969.

\bibitem[Guo et~al.(2020)Guo, Cheng, Li, Hahn, and Liu]{Guo2020}
Ruocheng Guo, Lu~Cheng, Jundong Li, P.~Richard Hahn, and Huan Liu.
\newblock A survey of learning causality with data: Problems and methods.
\newblock \emph{ACM Comput. Surv.}, 53\penalty0 (4), jul 2020.
\newblock ISSN 0360-0300.
\newblock \doi{10.1145/3397269}.
\newblock URL \url{https://doi.org/10.1145/3397269}.

\bibitem[Guyon et~al.(2011)Guyon, Aliferis, Cooper, Elisseeff, Pellet, Spirtes, and Statnikov]{guyon2011causality}
Isabelle Guyon, Constantin Aliferis, Gregory Cooper, Andr{\'e} Elisseeff, Jean~Philippe Pellet, Peter Spirtes, and Alexander Statnikov.
\newblock Causality workbench.
\newblock In \emph{Causality in the sciences}. Oxford University Press, 2011.

\bibitem[Hasan et~al.(2023)Hasan, Hossain, and Gani]{hasan2023survey}
Uzma Hasan, Emam Hossain, and Md~Osman Gani.
\newblock A survey on causal discovery methods for iid and time series data.
\newblock \emph{Transactions on Machine Learning Research}, 2023.

\bibitem[Hempel et~al.(2011)Hempel, Koseska, Kurths, and Nikoloski]{hempel2011}
Stefan Hempel, Aneta Koseska, J{\"u}rgen Kurths, and Zora Nikoloski.
\newblock Inner composition alignment for inferring directed networks from short time series.
\newblock \emph{Physical review letters}, 107\penalty0 (5):\penalty0 054101, 2011.

\bibitem[Henckel et~al.(2022)Henckel, Perkovi{\'c}, and Maathuis]{henckel2022graphical}
Leonard Henckel, Emilija Perkovi{\'c}, and Marloes~H Maathuis.
\newblock Graphical criteria for efficient total effect estimation via adjustment in causal linear models.
\newblock \emph{Journal of the Royal Statistical Society Series B: Statistical Methodology}, 84\penalty0 (2):\penalty0 579--599, 2022.

\bibitem[Hoyer et~al.(2008)Hoyer, Janzing, Mooij, Peters, and Sch{\"o}lkopf]{hoyer2008}
Patrik Hoyer, Dominik Janzing, Joris~M Mooij, Jonas Peters, and Bernhard Sch{\"o}lkopf.
\newblock Nonlinear causal discovery with additive noise models.
\newblock \emph{Advances in neural information processing systems}, 21, 2008.

\bibitem[Hoyer et~al.(2012)Hoyer, Hyvarinen, Scheines, Spirtes, Ramsey, Lacerda, and Shimizu]{hoyer2012causal}
Patrik~O Hoyer, Aapo Hyvarinen, Richard Scheines, Peter~L Spirtes, Joseph Ramsey, Gustavo Lacerda, and Shohei Shimizu.
\newblock Causal discovery of linear acyclic models with arbitrary distributions.
\newblock \emph{arXiv preprint arXiv:1206.3260}, 2012.

\bibitem[Hu and Liang(2014)]{hu2014}
Meng Hu and Hualou Liang.
\newblock A copula approach to assessing granger causality.
\newblock \emph{NeuroImage}, 100:\penalty0 125--134, 2014.

\bibitem[Hu et~al.(2015)Hu, Wang, Zhang, Kong, Cao, and Kozma]{hu2015comparison}
Sanqing Hu, Hui Wang, Jianhai Zhang, Wanzeng Kong, Yu~Cao, and Robert Kozma.
\newblock Comparison analysis: Granger causality and new causality and their applications to motor imagery.
\newblock \emph{IEEE transactions on neural networks and learning systems}, 27\penalty0 (7):\penalty0 1429--1444, 2015.

\bibitem[Huang et~al.(2020)Huang, Zhang, Zhang, Ramsey, Sanchez-Romero, Glymour, and Schlkopf]{JMLR:v21:19-232}
Biwei Huang, Kun Zhang, Jiji Zhang, Joseph Ramsey, Ruben Sanchez-Romero, Clark Glymour, and Bernhard Schlkopf.
\newblock Causal discovery from heterogeneous/nonstationary data.
\newblock \emph{Journal of Machine Learning Research}, 21\penalty0 (89):\penalty0 1--53, 2020.
\newblock URL \url{http://jmlr.org/papers/v21/19-232.html}.

\bibitem[Hyv{\"a}rinen and Smith(2013)]{hyvarinen2013pairwise}
Aapo Hyv{\"a}rinen and Stephen~M Smith.
\newblock Pairwise likelihood ratios for estimation of non-gaussian structural equation models.
\newblock \emph{The Journal of Machine Learning Research}, 14\penalty0 (1):\penalty0 111--152, 2013.

\bibitem[Hyv\"{a}rinen et~al.(2010)Hyv\"{a}rinen, Zhang, Shimizu, and Hoyer]{Hyvarinen2010}
Aapo Hyv\"{a}rinen, Kun Zhang, Shohei Shimizu, and Patrik~O. Hoyer.
\newblock Estimation of a structural vector autoregression model using non-gaussianity.
\newblock \emph{J. Mach. Learn. Res.}, 11:\penalty0 1709–1731, aug 2010.
\newblock ISSN 1532-4435.

\bibitem[Hyv{\"a}rinen et~al.(2024)Hyv{\"a}rinen, Khemakhem, and Monti]{hyvarinen2024identifiability}
Aapo Hyv{\"a}rinen, Ilyes Khemakhem, and Ricardo Monti.
\newblock Identifiability of latent-variable and structural-equation models: from linear to nonlinear.
\newblock \emph{Annals of the Institute of Statistical Mathematics}, 76\penalty0 (1):\penalty0 1--33, 2024.

\bibitem[Ikeuchi et~al.(2023)Ikeuchi, Ide, Zeng, Maeda, and Shimizu]{lingam2023}
Takashi Ikeuchi, Mayumi Ide, Yan Zeng, Takashi~Nicholas Maeda, and Shohei Shimizu.
\newblock {LiNGAM - Discovery of non-gaussian linear causal models (Version 1.9.0)}, 2023.
\newblock URL \url{https://github.com/cdt15/lingam.git}.

\bibitem[Imbens(2004)]{imbens2004nonparametric}
Guido~W Imbens.
\newblock Nonparametric estimation of average treatment effects under exogeneity: A review.
\newblock \emph{Review of Economics and statistics}, 86\penalty0 (1):\penalty0 4--29, 2004.

\bibitem[Jang et~al.(2022)Jang, Kim, and Noh]{jang2022vine}
Hyuna Jang, Jong-Min Kim, and Hohsuk Noh.
\newblock Vine copula granger causality in mean.
\newblock \emph{Economic Modelling}, 109:\penalty0 105798, 2022.

\bibitem[Jangyodsuk et~al.(2014)Jangyodsuk, Seo, and Gao]{jangyodsuk2014}
Piraporn Jangyodsuk, Dong-Jun Seo, and Jean Gao.
\newblock Causal graph discovery for hydrological time series knowledge discovery.
\newblock \emph{International Conference on Hydroinformatics}, 2014.

\bibitem[Janzing et~al.(2012)Janzing, Mooij, Zhang, Lemeire, Zscheischler, Daniu{\v{s}}is, Steudel, and Sch{\"o}lkopf]{janzing2012}
Dominik Janzing, Joris Mooij, Kun Zhang, Jan Lemeire, Jakob Zscheischler, Povilas Daniu{\v{s}}is, Bastian Steudel, and Bernhard Sch{\"o}lkopf.
\newblock Information-geometric approach to inferring causal directions.
\newblock \emph{Artificial Intelligence}, 182:\penalty0 1--31, 2012.

\bibitem[Javier(2021)]{ccm2021}
Prince Joseph~Erneszer Javier.
\newblock {causal-ccm a Python implementation of Convergent Cross Mapping (Version 0.3.3)}, June 2021.
\newblock URL \url{https://github.com/PrinceJavier/causal_ccm.git}.

\bibitem[Ji et~al.(2024)Ji, Zhang, Han, and Liu]{ji2024metacae}
Junzhong Ji, Zuozhen Zhang, Lu~Han, and Jinduo Liu.
\newblock Metacae: Causal autoencoder with meta-knowledge transfer for brain effective connectivity estimation.
\newblock \emph{Computers in Biology and Medicine}, 170:\penalty0 107940, 2024.

\bibitem[Jin and Xu(2024)]{jin2024contemporaneous}
Bingzi Jin and Xiaojie Xu.
\newblock Contemporaneous causality among price indices of ten major steel products.
\newblock \emph{Ironmaking \& Steelmaking}, page 03019233241249361, 2024.

\bibitem[K{\"a}ding and Runge(2021)]{kading2021benchmark}
Christoph K{\"a}ding and Jakob Runge.
\newblock A benchmark for bivariate causal discovery methods.
\newblock In \emph{EGU General Assembly Conference Abstracts}, pages EGU21--8584, 2021.

\bibitem[Kaiser and Sipos(2021)]{kaiser2021unsuitability}
Marcus Kaiser and Maksim Sipos.
\newblock Unsuitability of notears for causal graph discovery.
\newblock \emph{arXiv preprint arXiv:2104.05441}, 2021.

\bibitem[Kalainathan and Goudet(2019)]{cdt2019}
Diviyan Kalainathan and Olivier Goudet.
\newblock {CausalDiscoveryToolbox (Version 0.6.0)}, 2019.
\newblock URL \url{https://github.com/FenTechSolutions/CausalDiscoveryToolbox.git}.

\bibitem[Kalainathan et~al.(2020)Kalainathan, Goudet, and Dutta]{kalainathan2019}
Diviyan Kalainathan, Olivier Goudet, and Ritik Dutta.
\newblock Causal discovery toolbox: Uncovering causal relationships in python.
\newblock \emph{Journal of Machine Learning Research}, 21\penalty0 (37):\penalty0 1--5, 2020.

\bibitem[Kalisch and B{\"u}hlman(2007)]{kalisch2007}
Markus Kalisch and Peter B{\"u}hlman.
\newblock Estimating high-dimensional directed acyclic graphs with the pc-algorithm.
\newblock \emph{Journal of Machine Learning Research}, 8\penalty0 (3), 2007.

\bibitem[Kalisch et~al.(2012)Kalisch, M{\"a}chler, Colombo, Maathuis, and B{\"u}hlmann]{kalisch2012causal}
Markus Kalisch, Martin M{\"a}chler, Diego Colombo, Marloes~H Maathuis, and Peter B{\"u}hlmann.
\newblock Causal inference using graphical models with the r package pcalg.
\newblock \emph{Journal of statistical software}, 47:\penalty0 1--26, 2012.

\bibitem[Kawahara et~al.(2011)Kawahara, Shimizu, and Washio]{kawahara2011}
Yoshinobu Kawahara, Shohei Shimizu, and Takashi Washio.
\newblock Analyzing relationships among arma processes based on non-gaussianity of external influences.
\newblock \emph{Neurocomputing}, 74\penalty0 (12-13):\penalty0 2212--2221, 2011.

\bibitem[Kim et~al.(2020)Kim, Lee, and Hwang]{kim2020copula}
Jong-Min Kim, Namgil Lee, and Sun~Young Hwang.
\newblock A copula nonlinear granger causality.
\newblock \emph{Economic Modelling}, 88:\penalty0 420--430, 2020.

\bibitem[Kleinberg(2013)]{kleinberg2013causality}
Samantha Kleinberg.
\newblock \emph{Causality, probability, and time}.
\newblock Cambridge University Press, 2013.

\bibitem[Ko et~al.(2018)Ko, Lim, Ko, and Kim]{ko2018experimental}
Song Ko, Hyunki Lim, Hoon Ko, and Dae-Won Kim.
\newblock Experimental comparisons with respect to the usage of the promising relations in eda-based causal discovery.
\newblock \emph{Annals of Operations Research}, 265:\penalty0 241--255, 2018.

\bibitem[Kullback(1997)]{kullback1997}
Solomon Kullback.
\newblock Information theory and statistics.
\newblock \emph{Courier Corporation}, 1997.

\bibitem[Lachapelle et~al.(2019)Lachapelle, Brouillard, Deleu, and Lacoste-Julien]{lachapelle2019gradient}
S{\'e}bastien Lachapelle, Philippe Brouillard, Tristan Deleu, and Simon Lacoste-Julien.
\newblock Gradient-based neural dag learning.
\newblock \emph{arXiv preprint arXiv:1906.02226}, 2019.

\bibitem[Lam et~al.(2022)Lam, Andrews, and Ramsey]{lam2022greedy}
Wai-Yin Lam, Bryan Andrews, and Joseph Ramsey.
\newblock Greedy relaxations of the sparsest permutation algorithm.
\newblock In \emph{Uncertainty in Artificial Intelligence}, pages 1052--1062. PMLR, 2022.

\bibitem[Lauritzen and Spiegelhalter(1988)]{lauritzen1988local}
Steffen~L Lauritzen and David~J Spiegelhalter.
\newblock Local computations with probabilities on graphical structures and their application to expert systems.
\newblock \emph{Journal of the Royal Statistical Society: Series B (Methodological)}, 50\penalty0 (2):\penalty0 157--194, 1988.

\bibitem[Lawrence et~al.(2020)Lawrence, Kaiser, Sampaio, and Sipos]{dgp_causal_discovery_time_series}
Andrew~R. Lawrence, Marcus Kaiser, Rui Sampaio, and Maksim Sipos.
\newblock Data generating process to evaluate causal discovery techniques for time series data.
\newblock \emph{Causal Discovery \& Causality-Inspired Machine Learning Workshop at Neural Information Processing Systems}, 2020.

\bibitem[Lee and Lee(1998)]{lee1998independent}
Te-Won Lee and Te-Won Lee.
\newblock \emph{Independent component analysis}.
\newblock Springer, 1998.

\bibitem[Li et~al.(2014)Li, Za{\"\i}ane, and Osornio-Vargas]{li2014discovering}
Jundong Li, Osmar~R Za{\"\i}ane, and Alvaro Osornio-Vargas.
\newblock Discovering statistically significant co-location rules in datasets with extended spatial objects.
\newblock In \emph{Data Warehousing and Knowledge Discovery: 16th International Conference, DaWaK 2014, Munich, Germany, September 2-4, 2014. Proceedings 16}, pages 124--135. Springer, 2014.

\bibitem[Liao et~al.(2009)Liao, Marinazzo, Pan, Gong, and Chen]{liao2009}
Wei Liao, Daniele Marinazzo, Zhengyong Pan, Qiyong Gong, and Huafu Chen.
\newblock Kernel granger causality mapping effective connectivity on fmri data.
\newblock \emph{IEEE transactions on medical imaging}, 28\penalty0 (11):\penalty0 1825--1835, 2009.

\bibitem[L{\"o}we et~al.(2022)L{\"o}we, Madras, Zemel, and Welling]{lowe2022amortized}
Sindy L{\"o}we, David Madras, Richard Zemel, and Max Welling.
\newblock Amortized causal discovery: Learning to infer causal graphs from time-series data.
\newblock In \emph{Conference on Causal Learning and Reasoning}, pages 509--525. PMLR, 2022.

\bibitem[Luo et~al.(2024)Luo, Jin, Jin, Li, Ji, and Dai]{luo2024causal}
Jiaojiao Luo, Zhehao Jin, Heping Jin, Qian Li, Xu~Ji, and Yiyang Dai.
\newblock Causal temporal graph attention network for fault diagnosis of chemical processes.
\newblock \emph{Chinese Journal of Chemical Engineering}, 70:\penalty0 20--32, 2024.

\bibitem[L{\"u}tkepohl(2005)]{lutkepohl2005new}
Helmut L{\"u}tkepohl.
\newblock \emph{New introduction to multiple time series analysis}.
\newblock Springer Science \& Business Media, 2005.

\bibitem[Ma et~al.(2014)Ma, Aihara, and Chen]{ma2014}
Huanfei Ma, Kazuyuki Aihara, and Luonan Chen.
\newblock Detecting causality from nonlinear dynamics with short-term time series.
\newblock \emph{Scientific reports}, 4\penalty0 (1):\penalty0 7464, 2014.

\bibitem[Ma et~al.(2023)Ma, Wang, Bieganek, Tourani, and Aliferis]{ma2023local}
Sisi Ma, Jinhua Wang, Cameron Bieganek, Roshan Tourani, and Constantin Aliferis.
\newblock Local causal pathway discovery for single-cell rna sequencing count data: a benchmark study.
\newblock \emph{Journal of Translational Genetics and Genomics}, 7\penalty0 (1):\penalty0 50--65, 2023.

\bibitem[Maathuis and Colombo(2015)]{maathuis2015generalized}
Marloes~H Maathuis and Diego Colombo.
\newblock A generalized back-door criterion.
\newblock \emph{The Annals of Statistics}, 43\penalty0 (3):\penalty0 1060–1088, 2015.

\bibitem[Maeda(2022)]{maeda2022rcd}
Takashi~Nicholas Maeda.
\newblock I-rcd: an improved algorithm of repetitive causal discovery from data with latent confounders.
\newblock \emph{Behaviormetrika}, 49\penalty0 (2):\penalty0 329--341, 2022.

\bibitem[Maeda and Shimizu(2020)]{Maeda2020RCDRC}
Takashi~Nicholas Maeda and Shohei Shimizu.
\newblock Rcd: Repetitive causal discovery of linear non-gaussian acyclic models with latent confounders.
\newblock In \emph{International Conference on Artificial Intelligence and Statistics}, 2020.
\newblock URL \url{https://api.semanticscholar.org/CorpusID:210164913}.

\bibitem[Maeda and Shimizu(2021)]{Maeda2021CausalAM}
Takashi~Nicholas Maeda and Shohei Shimizu.
\newblock Causal additive models with unobserved variables.
\newblock In \emph{Conference on Uncertainty in Artificial Intelligence}, 2021.
\newblock URL \url{https://api.semanticscholar.org/CorpusID:237511555}.

\bibitem[Makhlouf et~al.(2020)Makhlouf, Zhioua, and Palamidessi]{makhlouf2020survey}
Karima Makhlouf, Sami Zhioua, and Catuscia Palamidessi.
\newblock Survey on causal-based machine learning fairness notions.
\newblock \emph{arXiv preprint arXiv:2010.09553}, 2020.

\bibitem[Mani and Cooper(2000)]{mani2000causal}
Subramani Mani and Gregory~F Cooper.
\newblock Causal discovery from medical textual data.
\newblock In \emph{Proceedings of the AMIA Symposium}, page 542. American Medical Informatics Association, 2000.

\bibitem[Mann and Whitney(1947)]{mann1947test}
Henry~B Mann and Donald~R Whitney.
\newblock On a test of whether one of two random variables is stochastically larger than the other.
\newblock \emph{The annals of mathematical statistics}, pages 50--60, 1947.

\bibitem[Mao and Shang(2017)]{mao2017}
Xuegeng Mao and Pengjian Shang.
\newblock Transfer entropy between multivariate time series.
\newblock \emph{Communications in Nonlinear Science and Numerical Simulation}, 47:\penalty0 338--347, 2017.

\bibitem[Marinazzo et~al.(2021)Marinazzo, M.Pellicoro, and Stramaglia]{kgc2021}
D.~Marinazzo, M.Pellicoro, and S.~Stramaglia.
\newblock {KernelGrangerCausality}, 2021.
\newblock URL \url{https://github.com/danielemarinazzo/KernelGrangerCausality.git}.

\bibitem[Marinazzo et~al.(2008)Marinazzo, Pellicoro, and Stramaglia]{marinazzo2008kernel}
Daniele Marinazzo, Mario Pellicoro, and Sebastiano Stramaglia.
\newblock Kernel-granger causality and the analysis of dynamical networks.
\newblock \emph{Physical Review E—Statistical, Nonlinear, and Soft Matter Physics}, 77\penalty0 (5):\penalty0 056215, 2008.

\bibitem[Marinazzo et~al.(2011)Marinazzo, Liao, Chen, and Stramaglia]{marinazzo2011nonlinear}
Daniele Marinazzo, Wei Liao, Huafu Chen, and Sebastiano Stramaglia.
\newblock Nonlinear connectivity by granger causality.
\newblock \emph{Neuroimage}, 58\penalty0 (2):\penalty0 330--338, 2011.

\bibitem[Masson-Delmotte et~al.(2021)Masson-Delmotte, Zhai, Pirani, Connors, P{\'e}an, Berger, Caud, Chen, Goldfarb, Gomis, et~al.]{masson2021climate}
Val{\'e}rie Masson-Delmotte, Panmao Zhai, Anna Pirani, Sarah~L Connors, Clotilde P{\'e}an, Sophie Berger, Nada Caud, Y~Chen, L~Goldfarb, MI~Gomis, et~al.
\newblock Climate change 2021: the physical science basis.
\newblock \emph{Contribution of working group I to the sixth assessment report of the intergovernmental panel on climate change}, 2\penalty0 (1):\penalty0 2391, 2021.

\bibitem[McCracken(2016)]{mccracken2016}
James~M McCracken.
\newblock Exploratory causal analysis with time series data.
\newblock \emph{Springer International Publishing}, 2016.
\newblock \doi{10.1007/978-3-031-01909-8_3}.
\newblock URL \url{https://doi.org/10.1007/978-3-031-01909-8_3}.

\bibitem[McCracken and Weigel(2014)]{mccracken2014}
James~M McCracken and Robert~S Weigel.
\newblock Convergent cross-mapping and pairwise asymmetric inference.
\newblock \emph{Physical Review E}, 90\penalty0 (6):\penalty0 062903, 2014.

\bibitem[Menegozzo et~al.(2021)Menegozzo, Dall’Alba, and Fiorini]{menegozzo2021industrial}
Giovanni Menegozzo, Diego Dall’Alba, and Paolo Fiorini.
\newblock Industrial time series modeling with causal precursors and separable temporal convolutions.
\newblock \emph{IEEE Robotics and Automation Letters}, 6\penalty0 (4):\penalty0 6939--6946, 2021.

\bibitem[Menegozzo et~al.(2022)Menegozzo, Dall’Alba, and Fiorini]{menegozzo2022cipcad}
Giovanni Menegozzo, Diego Dall’Alba, and Paolo Fiorini.
\newblock Cipcad-bench: Continuous industrial process datasets for benchmarking causal discovery methods.
\newblock In \emph{2022 IEEE 18th International Conference on Automation Science and Engineering (CASE)}, pages 2124--2131. IEEE, 2022.

\bibitem[Mojtabai(2024)]{mojtabai2024problematic}
Ramin Mojtabai.
\newblock Problematic social media use and psychological symptoms in adolescents.
\newblock \emph{Social psychiatry and psychiatric epidemiology}, pages 1--8, 2024.

\bibitem[Montavon et~al.(2018)Montavon, Samek, and M{\"u}ller]{montavon2018methods}
Gr{\'e}goire Montavon, Wojciech Samek, and Klaus-Robert M{\"u}ller.
\newblock Methods for interpreting and understanding deep neural networks.
\newblock \emph{Digital signal processing}, 73:\penalty0 1--15, 2018.

\bibitem[Monti et~al.(2020)Monti, Zhang, and Hyv{\"a}rinen]{monti2020causal}
Ricardo~Pio Monti, Kun Zhang, and Aapo Hyv{\"a}rinen.
\newblock Causal discovery with general non-linear relationships using non-linear ica.
\newblock In \emph{Uncertainty in artificial intelligence}, pages 186--195. PMLR, 2020.

\bibitem[Mooij et~al.(2016)Mooij, Peters, Janzing, Zscheischler, and Sch{\"o}lkopf]{mooij2016}
Joris~M Mooij, Jonas Peters, Dominik Janzing, Jakob Zscheischler, and Bernhard Sch{\"o}lkopf.
\newblock Distinguishing cause from effect using observational data: methods and benchmarks.
\newblock \emph{The Journal of Machine Learning Research}, 17\penalty0 (1):\penalty0 1103--1204, 2016.

\bibitem[Moraffah et~al.(2021)Moraffah, Sheth, Karami, Bhattacharya, Wang, Tahir, Raglin, and Liu]{Moraffah2021}
Raha Moraffah, Paras Sheth, Mansooreh Karami, Anchit Bhattacharya, Qianru Wang, Anique Tahir, Adrienne Raglin, and Huan Liu.
\newblock Causal inference for time series analysis: Problems, methods and evaluation.
\newblock \emph{Knowledge and Information Systems}, 63:\penalty0 3041--3085, 2021.

\bibitem[Naik and Kumar(2011)]{naik2011overview}
Ganesh~R Naik and Dinesh~K Kumar.
\newblock An overview of independent component analysis and its applications.
\newblock \emph{Informatica}, 35\penalty0 (1), 2011.

\bibitem[Nandy et~al.(2017)Nandy, Maathuis, and Richardson]{nandy2017estimating}
Preetam Nandy, Marloes~H Maathuis, and Thomas~S Richardson.
\newblock Estimating the effect of joint interventions from observational data in sparse high-dimensional settings.
\newblock \emph{The Annals of Statistics}, 45\penalty0 (2):\penalty0 647--674, 2017.

\bibitem[Nauta et~al.(2019{\natexlab{a}})Nauta, Bucur, and Seifert]{nauta2019}
Meike Nauta, Doina Bucur, and Christin Seifert.
\newblock Causal discovery with attention-based convolutional neural networks.
\newblock \emph{Machine Learning and Knowledge Extraction}, 1\penalty0 (1):\penalty0 19, 2019{\natexlab{a}}.

\bibitem[Nauta et~al.(2019{\natexlab{b}})Nauta, Bucur, and Seifert]{tcdf2019}
Meike Nauta, Doina Bucur, and Christin Seifert.
\newblock {TCDF-Temporal Causal Discovery Framework (PyTorch)}, 2019{\natexlab{b}}.
\newblock URL \url{https://github.com/M-Nauta/TCDF.git}.

\bibitem[Ng et~al.(2020)Ng, Ghassami, and Zhang]{ng2020role}
Ignavier Ng, AmirEmad Ghassami, and Kun Zhang.
\newblock On the role of sparsity and dag constraints for learning linear dags.
\newblock \emph{Advances in Neural Information Processing Systems}, 33:\penalty0 17943--17954, 2020.

\bibitem[Nogueira et~al.(2021)Nogueira, Gama, and Ferreira]{nogueira2021causal}
Ana~Rita Nogueira, Jo{\~a}o Gama, and Carlos~Abreu Ferreira.
\newblock Causal discovery in machine learning: Theories and applications.
\newblock \emph{Journal of Dynamics \& Games}, 8\penalty0 (3), 2021.

\bibitem[Nogueira et~al.(2022)Nogueira, Pugnana, Ruggieri, Pedreschi, and Gama]{Nogueira2022}
Ana~Rita Nogueira, Andrea Pugnana, Salvatore Ruggieri, Dino Pedreschi, and João Gama.
\newblock Methods and tools for causal discovery and causal inference.
\newblock \emph{WIREs Data Mining and Knowledge Discovery}, 12\penalty0 (2):\penalty0 e1449, 2022.
\newblock \doi{https://doi.org/10.1002/widm.1449}.
\newblock URL \url{https://wires.onlinelibrary.wiley.com/doi/abs/10.1002/widm.1449}.

\bibitem[Ombadi et~al.(2020)Ombadi, Nguyen, Sorooshian, and Hsu]{ombadi2020evaluation}
Mohammed Ombadi, Phu Nguyen, Soroosh Sorooshian, and Kuo-lin Hsu.
\newblock Evaluation of methods for causal discovery in hydrometeorological systems.
\newblock \emph{Water Resources Research}, 56\penalty0 (7):\penalty0 e2020WR027251, 2020.

\bibitem[Pamfil et~al.(2020)Pamfil, Sriwattanaworachai, Desai, Pilgerstorfer, Georgatzis, Beaumont, and Aragam]{pamfil2020}
Roxana Pamfil, Nisara Sriwattanaworachai, Shaan Desai, Philip Pilgerstorfer, Konstantinos Georgatzis, Paul Beaumont, and Bryon Aragam.
\newblock Dynotears: Structure learning from time-series data.
\newblock \emph{PMLR}, pages 1595--1605, 2020.

\bibitem[Pan et~al.(2018)Pan, Liang, Zhang, Yi, Yu, and Zheng]{pan2018}
Zheyi Pan, Yuxuan Liang, Junbo Zhang, Xiuwen Yi, Yong Yu, and Yu~Zheng.
\newblock Hyperst-net: Hypernetworks for spatio-temporal forecasting.
\newblock \emph{arXiv preprint arXiv:1809.10889}, 2018.

\bibitem[Pastorello et~al.(2020)Pastorello, Trotta, and Canfora]{pastorello2020}
Gilberto Pastorello, Carlo Trotta, and Canfora.
\newblock The fluxnet2015 dataset and the oneflux processing pipeline for eddy covariance data.
\newblock \emph{Scientific data}, 7\penalty0 (1):\penalty0 225, 2020.

\bibitem[Pearl(1985)]{pearl1985bayesian}
Judea Pearl.
\newblock Bayesian networks: A model of self-activated memory for evidential reasoning.
\newblock In \emph{Proceedings of the 7th conference of the Cognitive Science Society, University of California, Irvine, CA, USA}, pages 15--17, 1985.

\bibitem[Pearl(2009)]{pearl2009causality}
Judea Pearl.
\newblock \emph{Causality}.
\newblock Cambridge university press, 2009.

\bibitem[Pearl et~al.(2000)]{pearl2000}
Judea Pearl et~al.
\newblock Models, reasoning and inference.
\newblock \emph{Cambridge, UK: CambridgeUniversityPress}, 19\penalty0 (2):\penalty0 3, 2000.

\bibitem[Peters and B{\"u}hlmann(2015)]{peters2015structural}
Jonas Peters and Peter B{\"u}hlmann.
\newblock Structural intervention distance for evaluating causal graphs.
\newblock \emph{Neural computation}, 27\penalty0 (3):\penalty0 771--799, 2015.

\bibitem[Peters et~al.(2013)Peters, Janzing, and Sch{\"o}lkopf]{peters2013}
Jonas Peters, Dominik Janzing, and Bernhard Sch{\"o}lkopf.
\newblock Causal inference on time series using restricted structural equation models.
\newblock \emph{Advances in neural information processing systems}, 26, 2013.

\bibitem[Peters et~al.(2017)Peters, Janzing, and Sch{\"o}lkopf]{peters2017elements}
Jonas Peters, Dominik Janzing, and Bernhard Sch{\"o}lkopf.
\newblock \emph{Elements of causal inference: foundations and learning algorithms}.
\newblock The MIT Press, 2017.

\bibitem[Petersen et~al.(2010)Petersen, Aisen, Beckett, Donohue, Gamst, Harvey, Jack, Jagust, Shaw, Toga, et~al.]{petersen2010}
Ronald~Carl Petersen, Paul~S Aisen, Laurel~A Beckett, Michael~C Donohue, Anthony~Collins Gamst, Danielle~J Harvey, Clifford~R Jack, William~J Jagust, Leslie~M Shaw, Arthur~W Toga, et~al.
\newblock Alzheimer's disease neuroimaging initiative (adni): clinical characterization.
\newblock \emph{Neurology}, 74\penalty0 (3):\penalty0 201--209, 2010.

\bibitem[Raghu et~al.(2018)Raghu, Ramsey, Morris, Manatakis, Sprites, Chrysanthis, Glymour, and Benos]{Raghu2018}
Vineet Raghu, Joseph Ramsey, Alison Morris, Dimitris Manatakis, Peter Sprites, Panos Chrysanthis, Clark Glymour, and Panayiotis Benos.
\newblock Comparison of strategies for scalable causal discovery of latent variable models from mixed data.
\newblock \emph{International Journal of Data Science and Analytics}, 6, 08 2018.
\newblock \doi{10.1007/s41060-018-0104-3}.

\bibitem[Ramsey et~al.(2018)Ramsey, Zhang, Glymour, Romero, Huang, Ebert-Uphoff, Samarasinghe, Barnes, and Glymour]{ramsey2018tetrad}
Joseph~D Ramsey, Kun Zhang, Madelyn Glymour, Ruben~Sanchez Romero, Biwei Huang, Imme Ebert-Uphoff, Savini Samarasinghe, Elizabeth~A Barnes, and Clark Glymour.
\newblock Tetrad—a toolbox for causal discovery.
\newblock In \emph{8th international workshop on climate informatics}, pages 1--4, 2018.

\bibitem[Richardson and Spirtes(2002)]{richardson2002ancestral}
Thomas Richardson and Peter Spirtes.
\newblock Ancestral graph markov models.
\newblock \emph{The Annals of Statistics}, 30\penalty0 (4):\penalty0 962--1030, 2002.

\bibitem[Rosenstr{\"o}m et~al.(2023)Rosenstr{\"o}m, Czajkowski, Solbakken, and Saarni]{rosenstrom2023direction}
Tom~H Rosenstr{\"o}m, Nikolai~O Czajkowski, Ole~Andr{\'e} Solbakken, and Suoma~E Saarni.
\newblock Direction of dependence analysis for pre-post assessments using non-gaussian methods: a tutorial.
\newblock \emph{Psychotherapy Research}, 33\penalty0 (8):\penalty0 1058--1075, 2023.

\bibitem[Rubin(1974)]{rubin1974estimating}
Donald~B Rubin.
\newblock Estimating causal effects of treatments in randomized and nonrandomized studies.
\newblock \emph{Journal of educational Psychology}, 66\penalty0 (5):\penalty0 688, 1974.

\bibitem[Runge(2018)]{runge2018conditional}
Jakob Runge.
\newblock Conditional independence testing based on a nearest-neighbor estimator of conditional mutual information.
\newblock In \emph{International Conference on Artificial Intelligence and Statistics}, pages 938--947. Pmlr, 2018.

\bibitem[Runge(2020)]{runge2020discovering}
Jakob Runge.
\newblock Discovering contemporaneous and lagged causal relations in autocorrelated nonlinear time series datasets.
\newblock In \emph{Conference on Uncertainty in Artificial Intelligence}, pages 1388--1397. Pmlr, 2020.

\bibitem[Runge(2021)]{runge2021necessary}
Jakob Runge.
\newblock Necessary and sufficient graphical conditions for optimal adjustment sets in causal graphical models with hidden variables.
\newblock \emph{Advances in Neural Information Processing Systems}, 34:\penalty0 15762--15773, 2021.

\bibitem[Runge et~al.(2019)Runge, Nowack, Kretschmer, Flaxman, and Sejdinovic]{runge2019}
Jakob Runge, Peer Nowack, Marlene Kretschmer, Seth Flaxman, and Dino Sejdinovic.
\newblock Detecting and quantifying causal associations in large nonlinear time series datasets.
\newblock \emph{Science advances}, 5\penalty0 (11):\penalty0 eaau4996, 2019.

\bibitem[Runge et~al.(2020)Runge, Tibau, Bruhns, Mu{\~n}oz-Mar{\'\i}, and Camps-Valls]{runge2020}
Jakob Runge, Xavier-Andoni Tibau, Matthias Bruhns, Jordi Mu{\~n}oz-Mar{\'\i}, and Gustau Camps-Valls.
\newblock The causality for climate competition.
\newblock In \emph{NeurIPS 2019 Competition and Demonstration Track}, pages 110--120. PMLR, 2020.

\bibitem[Runge et~al.(2023{\natexlab{a}})Runge, Gerhardus, Varando, Eyring, and Camps-Valls]{runge2023}
Jakob Runge, Andreas Gerhardus, Gherardo Varando, Veronika Eyring, and Gustau Camps-Valls.
\newblock Causal inference for time series.
\newblock \emph{Nature Reviews Earth \& Environment}, pages 1--19, 2023{\natexlab{a}}.

\bibitem[Runge et~al.(2023{\natexlab{b}})Runge, Gerhardus, Varando, Eyring, and Camps-Valls]{tigramite2023}
Jakob Runge, Andreas Gerhardus, Gherardo Varando, Veronika Eyring, and Gustau Camps-Valls.
\newblock {tigramite (Version 5.2)}, 2023{\natexlab{b}}.
\newblock URL \url{https://github.com/jakobrunge/tigramite.git}.

\bibitem[Sachs et~al.(2005)Sachs, Perez, Pe'er, Lauffenburger, and Nolan]{sachs2005}
Karen Sachs, Omar Perez, Dana Pe'er, Douglas~A Lauffenburger, and Garry~P Nolan.
\newblock Causal protein-signaling networks derived from multiparameter single-cell data.
\newblock \emph{Science}, 308\penalty0 (5721):\penalty0 523--529, 2005.

\bibitem[Scholkopf(2019)]{Scholkopf2019}
Bernhard Scholkopf.
\newblock Causality for machine learning.
\newblock \emph{Probabilistic and Causal Inference}, 2019.
\newblock URL \url{https://api.semanticscholar.org/CorpusID:208267600}.

\bibitem[Schreiber(2000)]{schreiber2000}
Thomas Schreiber.
\newblock Measuring information transfer.
\newblock \emph{Physical review letters}, 85\penalty0 (2):\penalty0 461, 2000.

\bibitem[Scutari(2009)]{scutari2009learning}
Marco Scutari.
\newblock Learning bayesian networks with the bnlearn r package.
\newblock \emph{arXiv preprint arXiv:0908.3817}, 2009.

\bibitem[Sekhon(2008)]{sekhon2008multivariate}
Jasjeet~S Sekhon.
\newblock Multivariate and propensity score matching software with automated balance optimization: the matching package for r.
\newblock \emph{Journal of Statistical Software, Forthcoming}, 2008.

\bibitem[Shimizu et~al.(2006)Shimizu, Hoyer, Hyv{\"a}rinen, Kerminen, and Jordan]{shimizu2006}
Shohei Shimizu, Patrik~O Hoyer, Aapo Hyv{\"a}rinen, Antti Kerminen, and Michael Jordan.
\newblock A linear non-gaussian acyclic model for causal discovery.
\newblock \emph{Journal of Machine Learning Research}, 7\penalty0 (10), 2006.

\bibitem[Shimizu et~al.(2011)Shimizu, Inazumi, Sogawa, Hyv\"{a}rinen, Kawahara, Washio, Hoyer, and Bollen]{Shimizu2011}
Shohei Shimizu, Takanori Inazumi, Yasuhiro Sogawa, Aapo Hyv\"{a}rinen, Yoshinobu Kawahara, Takashi Washio, Patrik~O. Hoyer, and Kenneth Bollen.
\newblock Directlingam: A direct method for learning a linear non-gaussian structural equation model.
\newblock \emph{J. Mach. Learn. Res.}, 12\penalty0 (null):\penalty0 1225–1248, jul 2011.
\newblock ISSN 1532-4435.

\bibitem[Smith et~al.(2011)Smith, Miller, Salimi-Khorshidi, Webster, Beckmann, Nichols, Ramsey, and Woolrich]{smith2011}
Stephen~M Smith, Karla~L Miller, Gholamreza Salimi-Khorshidi, Matthew Webster, Christian~F Beckmann, Thomas~E Nichols, Joseph~D Ramsey, and Mark~W Woolrich.
\newblock Network modelling methods for fmri.
\newblock \emph{Neuroimage}, 54\penalty0 (2):\penalty0 875--891, 2011.

\bibitem[Sogawa et~al.(2010)Sogawa, Shimizu, Kawahara, and Washio]{sogawa2010experimental}
Yasuhiro Sogawa, Shohei Shimizu, Yoshinobu Kawahara, and Takashi Washio.
\newblock An experimental comparison of linear non-gaussian causal discovery methods and their variants.
\newblock In \emph{The 2010 International Joint Conference on Neural Networks (IJCNN)}, pages 1--8. IEEE, 2010.

\bibitem[Song et~al.(2016)Song, Oyama, Sato, and Kurihara]{song2016evaluation}
Jing Song, Satoshi Oyama, Haruhiko Sato, and Masahito Kurihara.
\newblock Evaluation of causal discovery models in bivariate case using real world data.
\newblock In \emph{Proceedings of the International MultiConference of Engineers and Computer Scientists}, volume~1, 2016.

\bibitem[Spiegelhalter et~al.(1993)Spiegelhalter, Dawid, Lauritzen, and Cowell]{spiegelhalter1993bayesian}
David~J Spiegelhalter, A~Philip Dawid, Steffen~L Lauritzen, and Robert~G Cowell.
\newblock Bayesian analysis in expert systems.
\newblock \emph{Statistical science}, pages 219--247, 1993.

\bibitem[Spirtes and Zhang(2016)]{spirtes2016causal}
Peter Spirtes and Kun Zhang.
\newblock Causal discovery and inference: concepts and recent methodological advances.
\newblock In \emph{Applied informatics}, volume~3, pages 1--28. Springer, 2016.

\bibitem[Spirtes et~al.(2001)Spirtes, Glymour, and Scheines]{spirtes2001causation}
Peter Spirtes, Clark Glymour, and Richard Scheines.
\newblock \emph{Causation, prediction, and search}.
\newblock MIT press, 2001.

\bibitem[Spirtes et~al.(2013)Spirtes, Meek, and Richardson]{spirtes2013causal}
Peter~L Spirtes, Christopher Meek, and Thomas~S Richardson.
\newblock Causal inference in the presence of latent variables and selection bias.
\newblock \emph{arXiv preprint arXiv:1302.4983}, 2013.

\bibitem[Statnikov et~al.(2013)Statnikov, Hamner, Escalante, Isabelle, and Saeed]{cause-effect-pairs}
Alexander Statnikov, Ben Hamner, Hugo~Jair Escalante, Isabelle, and Mehreen Saeed.
\newblock Cause-effect pairs, 2013.
\newblock URL \url{https://kaggle.com/competitions/cause-effect-pairs}.

\bibitem[Stekhoven et~al.(2012)Stekhoven, Moraes, Sveinbj{\"o}rnsson, Hennig, Maathuis, and B{\"u}hlmann]{stekhoven2012causal}
Daniel~J Stekhoven, Izabel Moraes, Gardar Sveinbj{\"o}rnsson, Lars Hennig, Marloes~H Maathuis, and Peter B{\"u}hlmann.
\newblock Causal stability ranking.
\newblock \emph{Bioinformatics}, 28\penalty0 (21):\penalty0 2819--2823, 2012.

\bibitem[Sugihara et~al.(2012)Sugihara, May, Ye, Hsieh, Deyle, Fogarty, and Munch]{sugihara2012}
George Sugihara, Robert May, Hao Ye, Chih-hao Hsieh, Ethan Deyle, Michael Fogarty, and Stephan Munch.
\newblock Detecting causality in complex ecosystems.
\newblock \emph{science}, 338\penalty0 (6106):\penalty0 496--500, 2012.

\bibitem[Sun et~al.(2014)Sun, Cafaro, and Bollt]{sun2014identifying}
Jie Sun, Carlo Cafaro, and Erik~M Bollt.
\newblock Identifying the coupling structure in complex systems through the optimal causation entropy principle.
\newblock \emph{Entropy}, 16\penalty0 (6):\penalty0 3416--3433, 2014.

\bibitem[Sun et~al.(2015)Sun, Taylor, and Bollt]{sun2015}
Jie Sun, Dane Taylor, and Erik~M Bollt.
\newblock Causal network inference by optimal causation entropy.
\newblock \emph{SIAM Journal on Applied Dynamical Systems}, 14\penalty0 (1):\penalty0 73--106, 2015.

\bibitem[Takens(1981)]{takens1981}
Floris Takens.
\newblock Dynamical systems and turbulence.
\newblock \emph{Warwick, 1980}, pages 366--381, 1981.

\bibitem[Tank et~al.(2021{\natexlab{a}})Tank, Covert, Foti, Shojaie, and Fox]{ngc2021}
Alex Tank, Ian Covert, Nicholas Foti, Ali Shojaie, and Emily Fox.
\newblock {Neural-GC}, 2021{\natexlab{a}}.
\newblock URL \url{https://github.com/iancovert/Neural-GC.git}.

\bibitem[Tank et~al.(2021{\natexlab{b}})Tank, Covert, Foti, Shojaie, and Fox]{Tank2021}
Alex Tank, Ian Covert, Nicholas Foti, Ali Shojaie, and Emily~B Fox.
\newblock Neural granger causality.
\newblock \emph{{IEEE} Transactions on Pattern Analysis and Machine Intelligence}, pages 1--1, 2021{\natexlab{b}}.
\newblock \doi{10.1109/tpami.2021.3065601}.
\newblock URL \url{https://doi.org/10.1109%2Ftpami.2021.3065601}.

\bibitem[Tu et~al.(2019)Tu, Zhang, Bertilson, Kjellstrom, and Zhang]{tu2019neuropathic}
Ruibo Tu, Kun Zhang, Bo~Bertilson, Hedvig Kjellstrom, and Cheng Zhang.
\newblock Neuropathic pain diagnosis simulator for causal discovery algorithm evaluation.
\newblock \emph{Advances in Neural Information Processing Systems}, 32, 2019.

\bibitem[Uemura et~al.(2022)Uemura, Takagi, Takayuki, Yoshida, and Shimizu]{uemura2022multivariate}
Kento Uemura, Takuya Takagi, Kambayashi Takayuki, Hiroyuki Yoshida, and Shohei Shimizu.
\newblock A multivariate causal discovery based on post-nonlinear model.
\newblock In \emph{Conference on Causal Learning and Reasoning}, pages 826--839. PMLR, 2022.

\bibitem[Van~den Bulcke et~al.(2006)Van~den Bulcke, Van~Leemput, Naudts, van Remortel, Ma, Verschoren, De~Moor, and Marchal]{van2006}
Tim Van~den Bulcke, Koenraad Van~Leemput, Bart Naudts, Piet van Remortel, Hongwu Ma, Alain Verschoren, Bart De~Moor, and Kathleen Marchal.
\newblock Syntren: a generator of synthetic gene expression data for design and analysis of structure learning algorithms.
\newblock \emph{BMC bioinformatics}, 7:\penalty0 1--12, 2006.

\bibitem[Wang et~al.(2014)Wang, Shang, Lin, and Chen]{wang2014segmented}
Jing Wang, Pengjian Shang, Aijin Lin, and Yuechen Chen.
\newblock Segmented inner composition alignment to detect coupling of different subsystems.
\newblock \emph{Nonlinear Dynamics}, 76:\penalty0 1821--1828, 2014.

\bibitem[Wang et~al.(2023)Wang, Ruan, Hong, and Luo]{wang2023detecting}
Lu~Wang, Hang Ruan, Yanran Hong, and Keyu Luo.
\newblock Detecting the hidden asymmetric relationship between crude oil and the us dollar: A novel neural granger causality method.
\newblock \emph{Research in International Business and Finance}, 64:\penalty0 101899, 2023.

\bibitem[Wang et~al.(2021)Wang, Du, Zhu, Ke, Chen, Hao, and Wang]{wang2021ordering}
Xiaoqiang Wang, Yali Du, Shengyu Zhu, Liangjun Ke, Zhitang Chen, Jianye Hao, and Jun Wang.
\newblock Ordering-based causal discovery with reinforcement learning.
\newblock \emph{arXiv preprint arXiv:2105.06631}, 2021.

\bibitem[Xie et~al.(2019)Xie, Cai, Zeng, Gao, and Hao]{xie2019efficient}
Feng Xie, Ruichu Cai, Yan Zeng, Jiantao Gao, and Zhifeng Hao.
\newblock An efficient entropy-based causal discovery method for linear structural equation models with iid noise variables.
\newblock \emph{IEEE transactions on neural networks and learning systems}, 31\penalty0 (5):\penalty0 1667--1680, 2019.

\bibitem[Yao et~al.(2021)Yao, Chu, Li, Li, Gao, and Zhang]{Yao2021}
Liuyi Yao, Zhixuan Chu, Sheng Li, Yaliang Li, Jing Gao, and Aidong Zhang.
\newblock A survey on causal inference.
\newblock \emph{ACM Trans. Knowl. Discov. Data}, 15\penalty0 (5), may 2021.
\newblock ISSN 1556-4681.
\newblock \doi{10.1145/3444944}.
\newblock URL \url{https://doi.org/10.1145/3444944}.

\bibitem[Ye et~al.(2015)Ye, Deyle, Gilarranz, and Sugihara]{ye2015distinguishing}
Hao Ye, Ethan~R Deyle, Luis~J Gilarranz, and George Sugihara.
\newblock Distinguishing time-delayed causal interactions using convergent cross mapping.
\newblock \emph{Scientific reports}, 5\penalty0 (1):\penalty0 14750, 2015.

\bibitem[Yu et~al.(2019)Yu, Chen, Gao, and Yu]{yu2019daggnn}
Yue Yu, Jie Chen, Tian Gao, and Mo~Yu.
\newblock Dag-gnn: Dag structure learning with graph neural networks.
\newblock In \emph{International conference on machine learning}, pages 7154--7163. PMLR, 2019.

\bibitem[Yuan and Malone(2013)]{10.5555/2591248.2591250}
Changhe Yuan and Brandon Malone.
\newblock Learning optimal bayesian networks: a shortest path perspective.
\newblock \emph{J. Artif. Int. Res.}, 48\penalty0 (1):\penalty0 23–65, oct 2013.
\newblock ISSN 1076-9757.

\bibitem[Zhang(2008)]{zhang2008completeness}
Jiji Zhang.
\newblock On the completeness of orientation rules for causal discovery in the presence of latent confounders and selection bias.
\newblock \emph{Artificial Intelligence}, 172\penalty0 (16-17):\penalty0 1873--1896, 2008.

\bibitem[Zhang et~al.(2021{\natexlab{a}})Zhang, Zhu, Kalander, Ng, Ye, Chen, and Pan]{gCastle2021}
Keli Zhang, Shengyu Zhu, Marcus Kalander, Ignavier Ng, Junjian Ye, Zhitang Chen, and Lujia Pan.
\newblock {gCastle (Version 1.0.4)}, 2021{\natexlab{a}}.
\newblock URL \url{https://github.com/huawei-noah/trustworthyAI.git}.

\bibitem[Zhang et~al.(2021{\natexlab{b}})Zhang, Zhu, Kalander, Ng, Ye, Chen, and Pan]{zhang2021gcastlepythontoolboxcausal}
Keli Zhang, Shengyu Zhu, Marcus Kalander, Ignavier Ng, Junjian Ye, Zhitang Chen, and Lujia Pan.
\newblock gcastle: A python toolbox for causal discovery.
\newblock \emph{arXiv preprint arXiv:2111.15155}, 2021{\natexlab{b}}.

\bibitem[Zhang and Hyvarinen(2012)]{zhang2012identifiability}
Kun Zhang and Aapo Hyvarinen.
\newblock On the identifiability of the post-nonlinear causal model.
\newblock \emph{arXiv preprint arXiv:1205.2599}, 2012.

\bibitem[Zhang et~al.(2015)Zhang, Wang, Zhang, and Sch{\"o}lkopf]{zhang2015estimation}
Kun Zhang, Zhikun Wang, Jiji Zhang, and Bernhard Sch{\"o}lkopf.
\newblock On estimation of functional causal models: general results and application to the post-nonlinear causal model.
\newblock \emph{ACM Transactions on Intelligent Systems and Technology (TIST)}, 7\penalty0 (2):\penalty0 1--22, 2015.

\bibitem[Zheng et~al.(2018)Zheng, Aragam, Ravikumar, and Xing]{zheng2018dags}
Xun Zheng, Bryon Aragam, Pradeep~K Ravikumar, and Eric~P Xing.
\newblock Dags with no tears: Continuous optimization for structure learning.
\newblock \emph{Advances in neural information processing systems}, 31, 2018.

\bibitem[Zheng et~al.(2020)Zheng, Dan, Aragam, Ravikumar, and Xing]{zheng2020learning}
Xun Zheng, Chen Dan, Bryon Aragam, Pradeep Ravikumar, and Eric Xing.
\newblock Learning sparse nonparametric dags.
\newblock In \emph{International Conference on Artificial Intelligence and Statistics}, pages 3414--3425. Pmlr, 2020.

\bibitem[Zheng et~al.(2024{\natexlab{a}})Zheng, Huang, Chen, Ramsey, Gong, Cai, Shimizu, Spirtes, and Zhang]{causallearn2024}
Yujia Zheng, Biwei Huang, Wei Chen, Joseph Ramsey, Mingming Gong, Ruichu Cai, Shohei Shimizu, Peter Spirtes, and Kun Zhang.
\newblock {causal-learn (Version 0.1.3.8)}, 2024{\natexlab{a}}.
\newblock URL \url{https://github.com/py-why/causal-learn.git}.

\bibitem[Zheng et~al.(2024{\natexlab{b}})Zheng, Huang, Chen, Ramsey, Gong, Cai, Shimizu, Spirtes, and Zhang]{zheng2024causal}
Yujia Zheng, Biwei Huang, Wei Chen, Joseph Ramsey, Mingming Gong, Ruichu Cai, Shohei Shimizu, Peter Spirtes, and Kun Zhang.
\newblock Causal-learn: Causal discovery in python.
\newblock \emph{Journal of Machine Learning Research}, 25\penalty0 (60):\penalty0 1--8, 2024{\natexlab{b}}.

\bibitem[Zhu et~al.(2019)Zhu, Ng, and Chen]{zhu2019causal}
Shengyu Zhu, Ignavier Ng, and Zhitang Chen.
\newblock Causal discovery with reinforcement learning.
\newblock \emph{arXiv preprint arXiv:1906.04477}, 2019.

\end{thebibliography}

\newpage

\begin{appendix}
\section{Residual Plot}
\label{appendix_a}
\setcounter{figure}{0}
\renewcommand{\thefigure}{A\arabic{figure}}
\begin{figure}[htbp]
\label{pairwise_fig}
    \centering
    \subfigure[Linear and Gaussian Datasets (pairwise)]{
    \label{fig.sub.1}
    \includegraphics[width=0.9\textwidth]{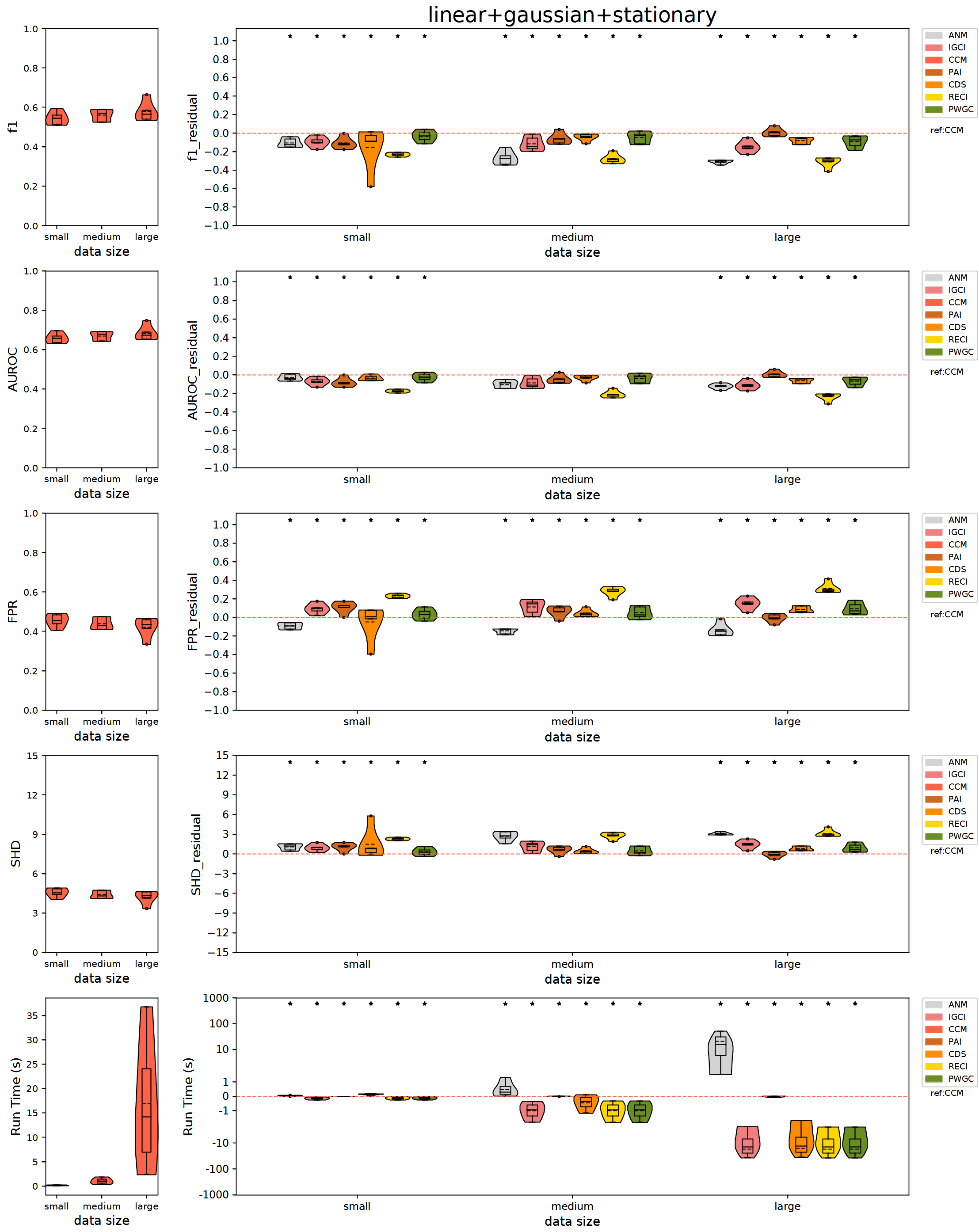}
    }
    \end{figure} 
    \begin{figure}[htbp]
    \centering
     \subfigure[Linear and Non-Gaussian Datasets (pairwise)]{
    \label{fig.sub.2}
    \includegraphics[width=0.9\textwidth]{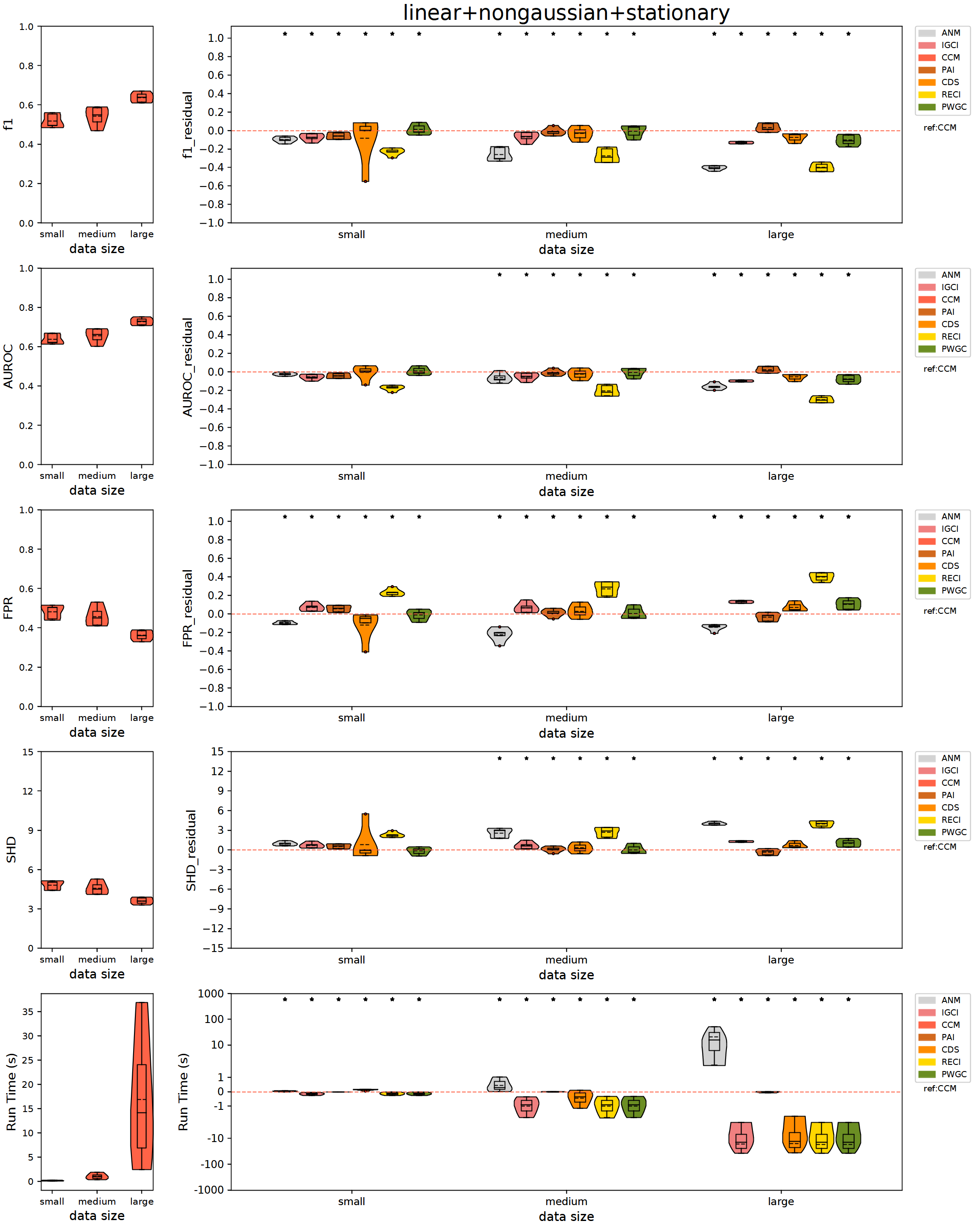}
    }
    \end{figure} 
\begin{figure}[htbp]
    \centering
     \subfigure[Nonlinear and Gaussian Datasets (pairwise)]{
    \label{fig.sub.3}
    \includegraphics[width=0.9\textwidth]{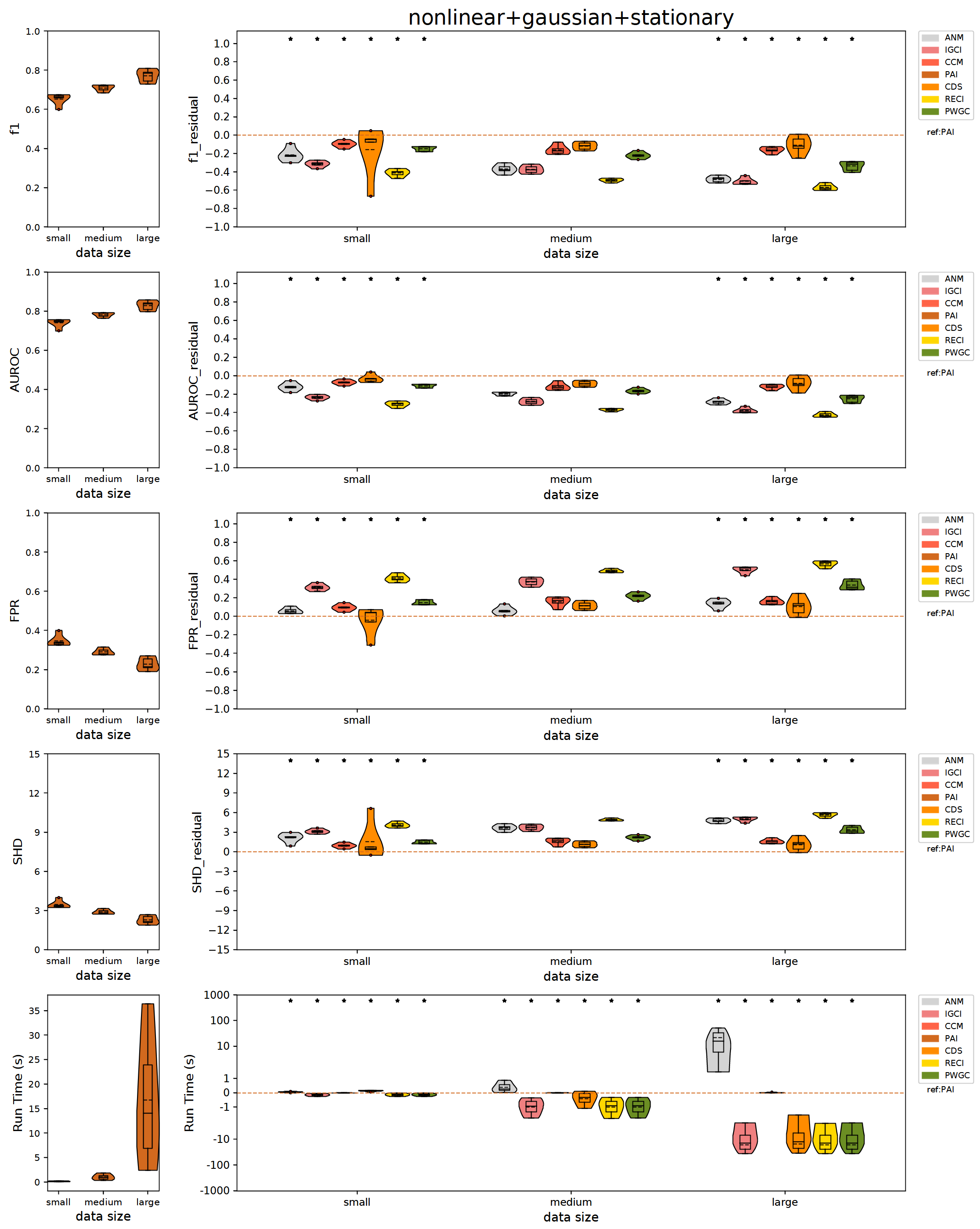}
    }
    \end{figure} 
    \begin{figure}[htbp]
    \centering
     \subfigure[Nonlinear and Non-Gaussian Datasets (pairwise)]{
    \label{fig.sub.4}
    \includegraphics[width=0.9\textwidth]{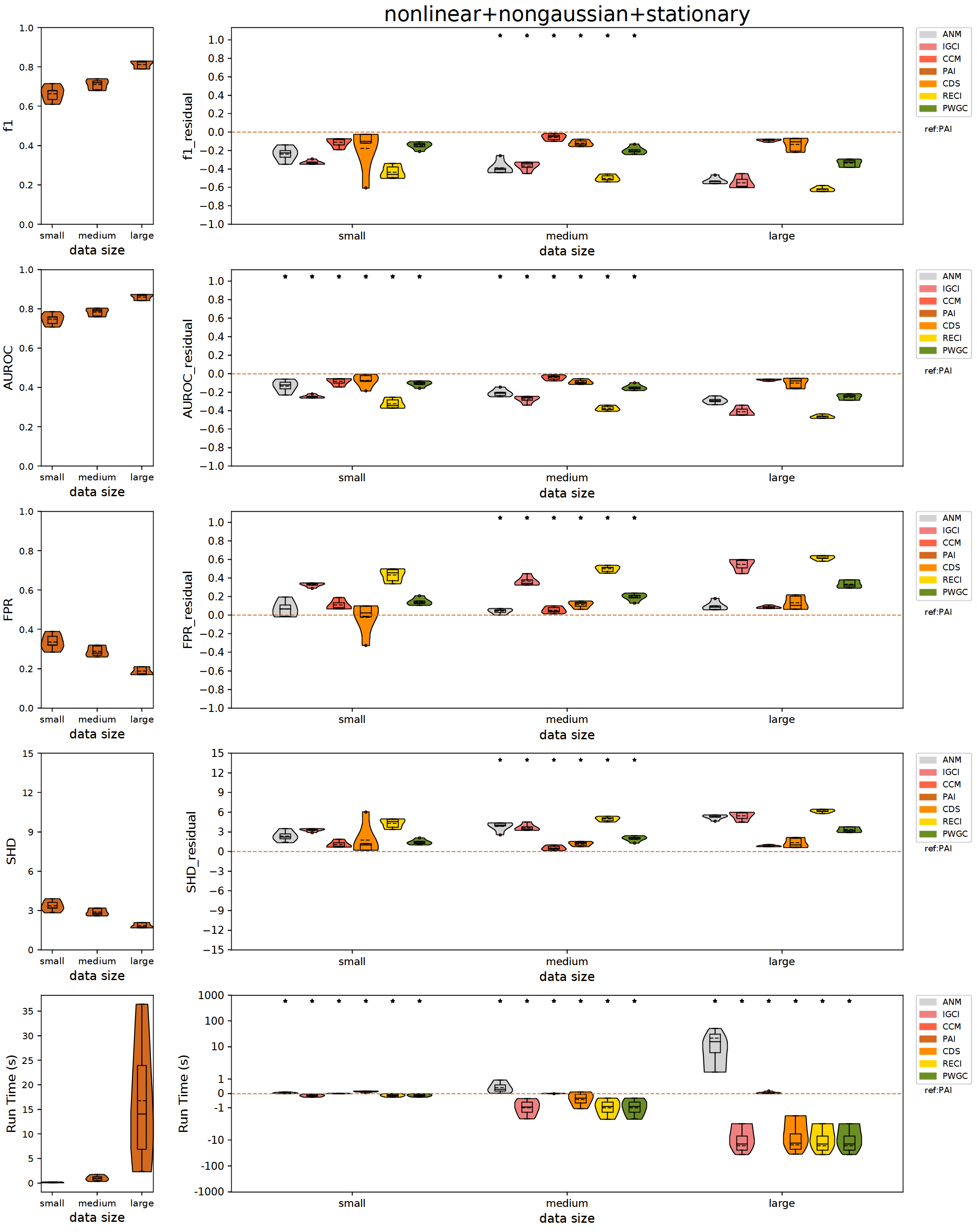}
    }
    \caption{Evaluation of pairwise algorithms (ANM, IGCI, CCM, PAI, CDS, PECI, PWGC) on synthetic datasets. The original violin plot of the reference algorithm is presented in the left column while the corresponding residual plot of alternative algorithms is presented in the right column. The presence of an asterisk denotes statistical significance as determined by the Mann-Whitney U test with a significance threshold of $p<0.05$.}
    \end{figure}

\begin{figure}[htbp]
   \label{instantaneous_fig}
    \centering
    \subfigure[Linear and Gaussian Datasets (instantaneous)]{
    \label{fig.sub.1}
    \includegraphics[width=0.9\textwidth]{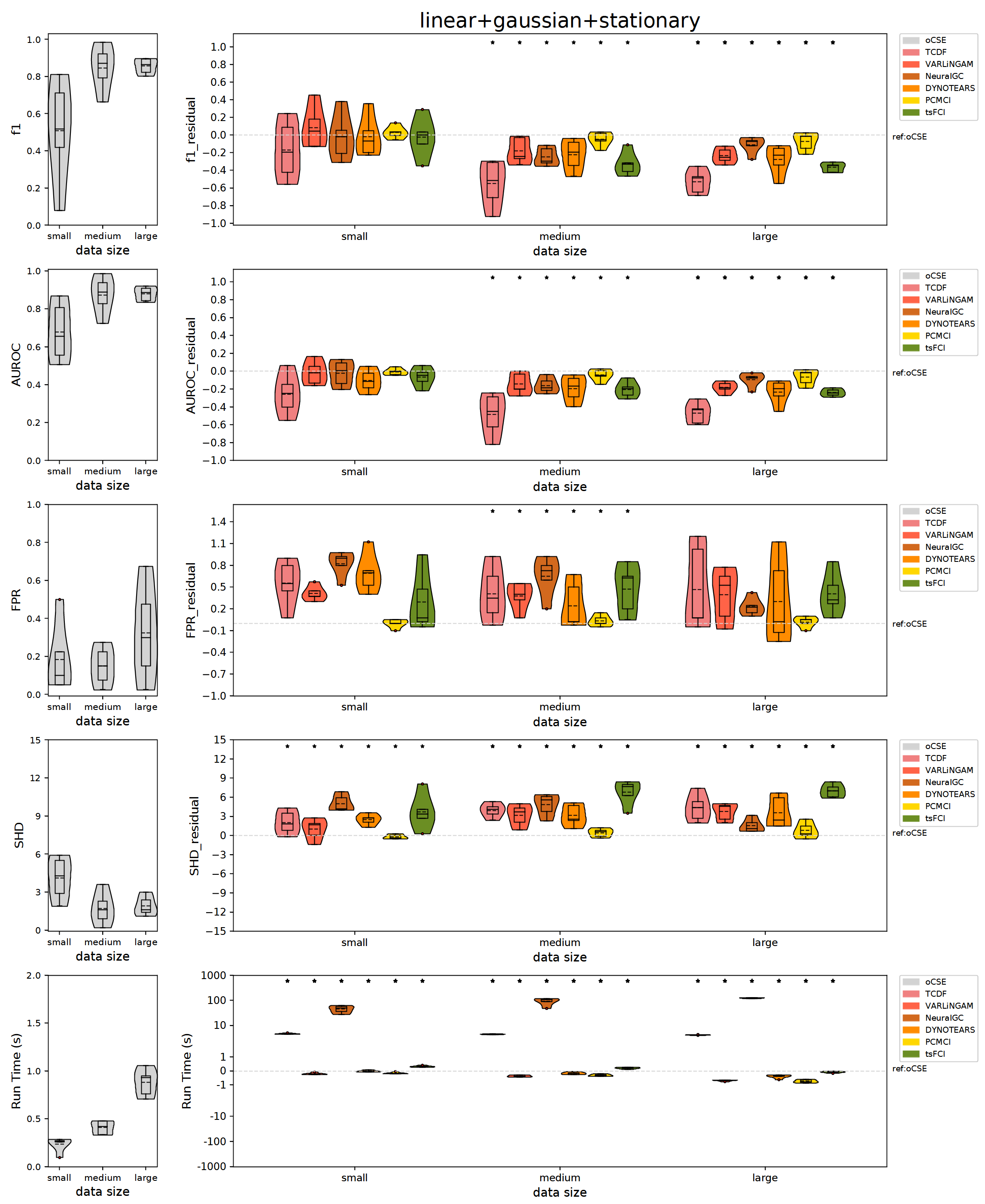}
    }
    \end{figure} 
    \begin{figure}[htbp]
    \centering
     \subfigure[Linear and Non-Gaussian Datasets (instantaneous)]{
    \label{fig.sub.2}
    \includegraphics[width=0.9\textwidth]{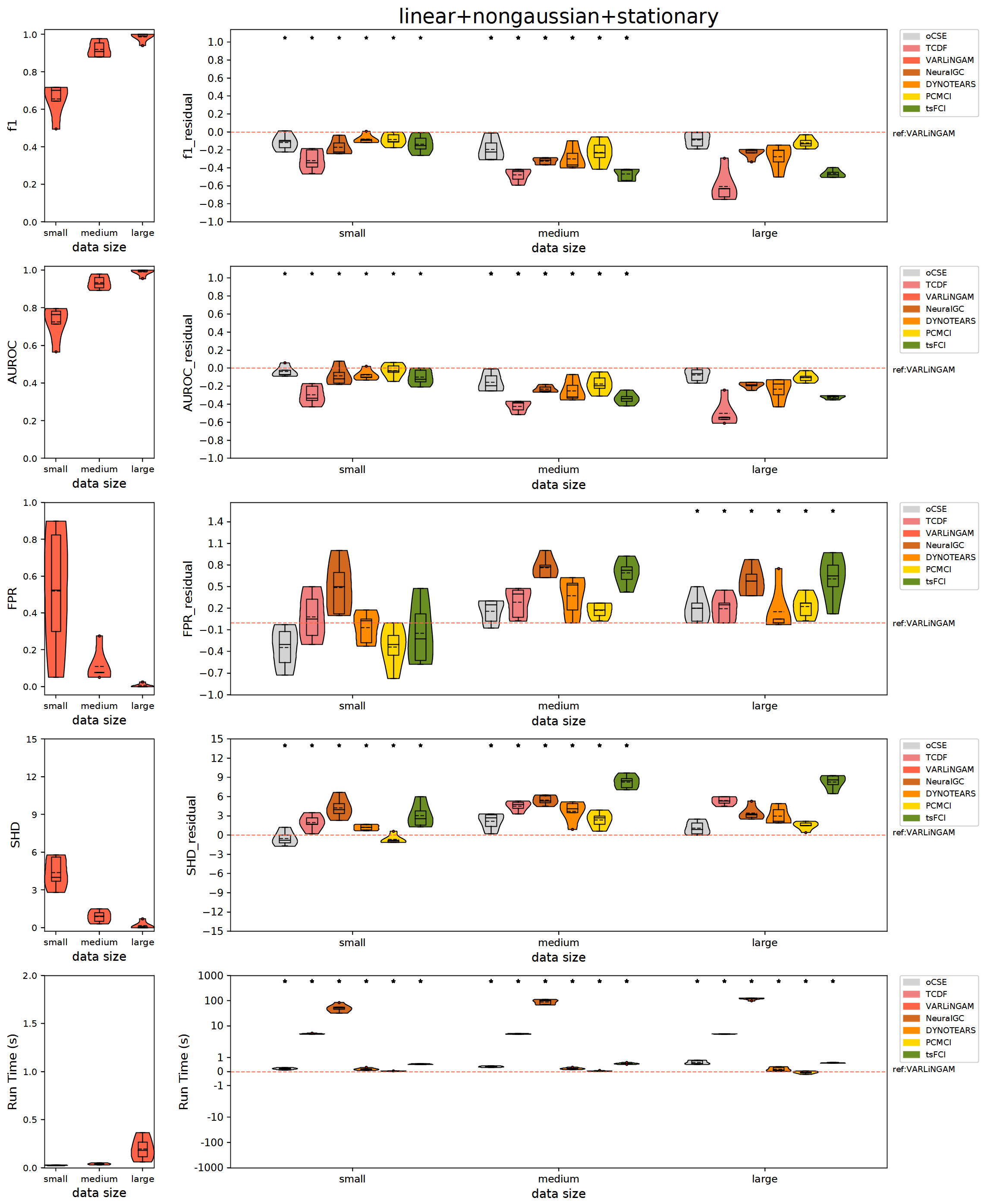}
    }
    \end{figure} 
\begin{figure}[htbp]
    \centering
     \subfigure[Nonlinear and Gaussian Datasets (instantaneous)]{
    \label{fig.sub.3}
    \includegraphics[width=0.9\textwidth]{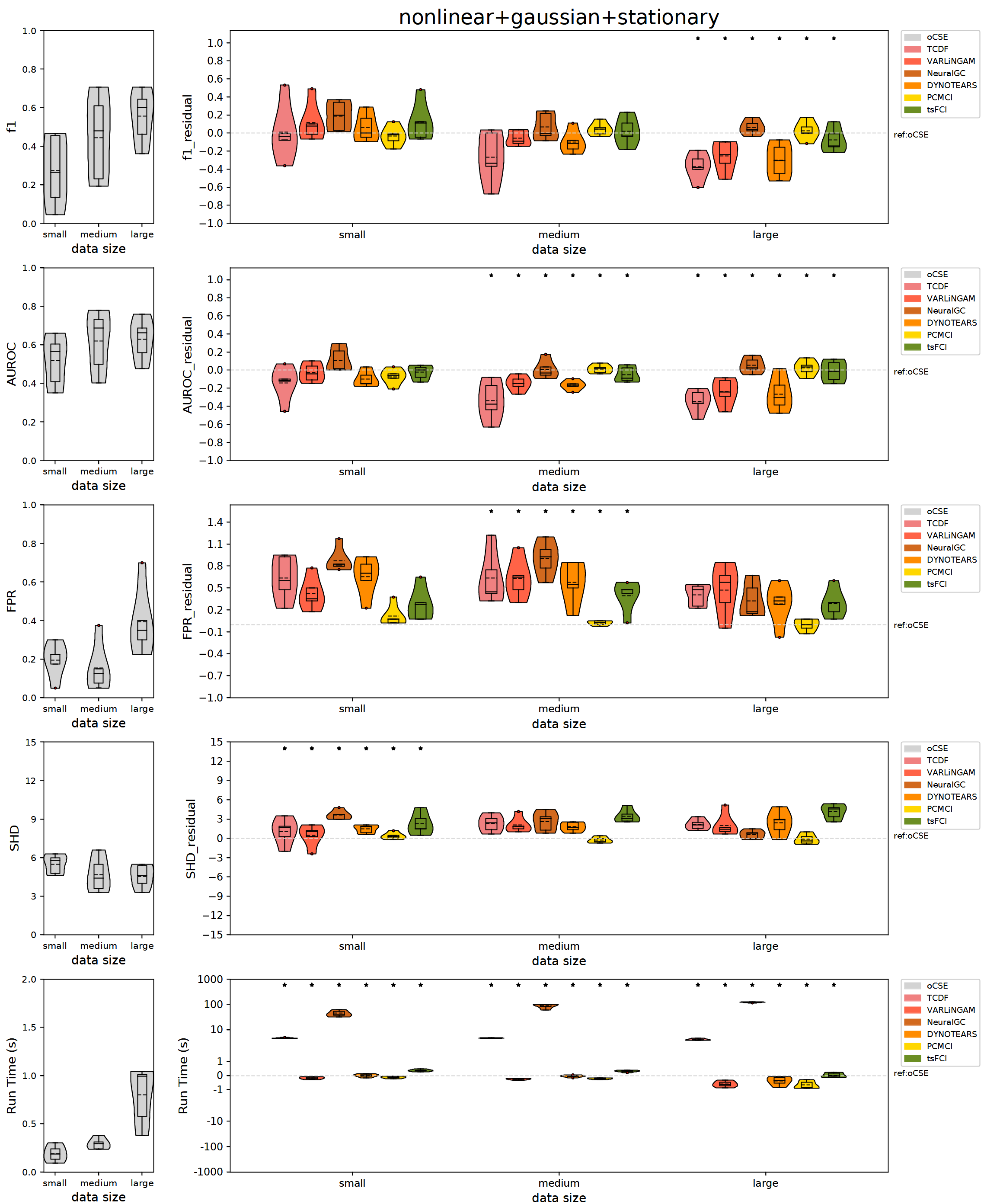}
    }
    \end{figure} 
    \begin{figure}[htbp]
    \centering
     \subfigure[Nonlinear and Non-Gaussian Datasets (instantaneous)]{
    \label{fig.sub.4}
    \includegraphics[width=0.9\textwidth]{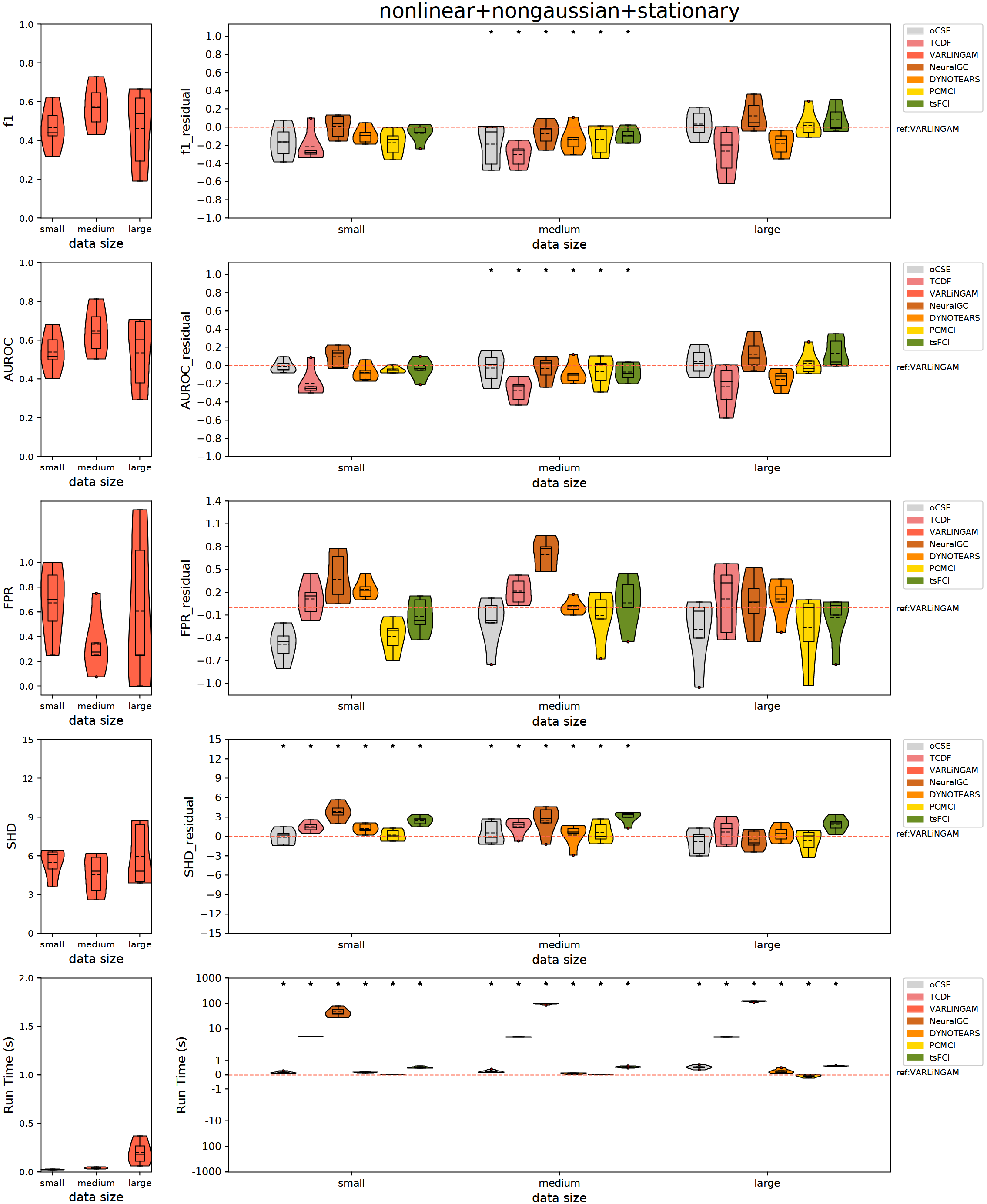}
    }
    \caption{Evaluation of instantaneous causal algorithms (oCSE, TCDF, VARLiNGAM, NeuralGC, DYNOTEARS, PCMCI, tsFCI) on synthetic datasets. The original violin plot of the reference algorithm is presented in the left column while the corresponding residual plot of alternative algorithms is presented in the right column. The presence of an asterisk denotes statistical significance as determined by the Mann-Whitney U test with a significance threshold of $p<0.05$.}
    \end{figure}

    \begin{figure}[htbp]
\label{time-delay_fig}
    \centering
    \subfigure[Linear and Gaussian Datasets (time-delay)]{
    \label{fig.sub.1}
    \includegraphics[width=0.98\textwidth]{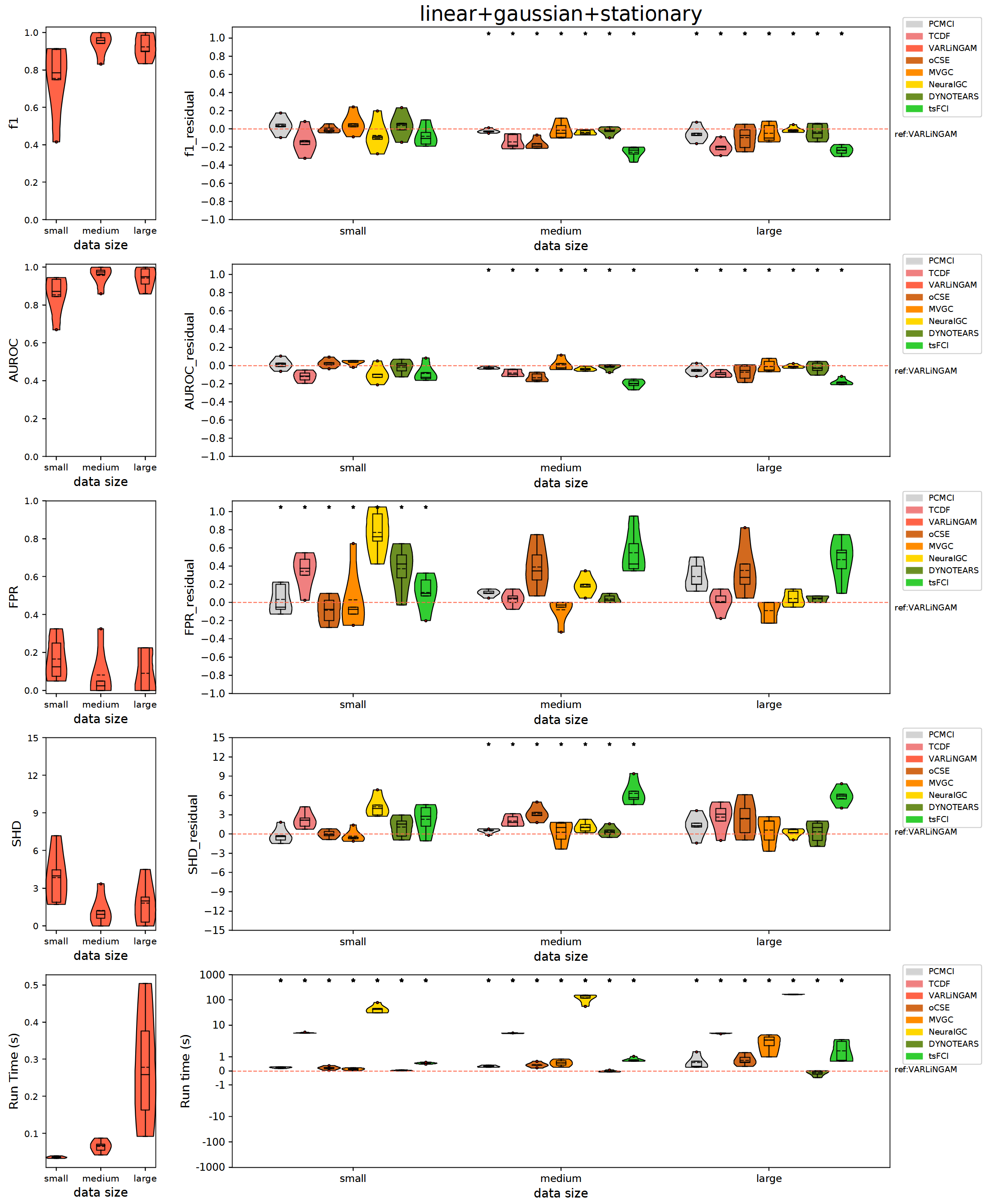}
    }
    \end{figure} 
    \begin{figure}[htbp]
    \centering
     \subfigure[Linear and Non-Gaussian Datasets (time-delay)]{
    \label{fig.sub.2}
    \includegraphics[width=0.98\textwidth]{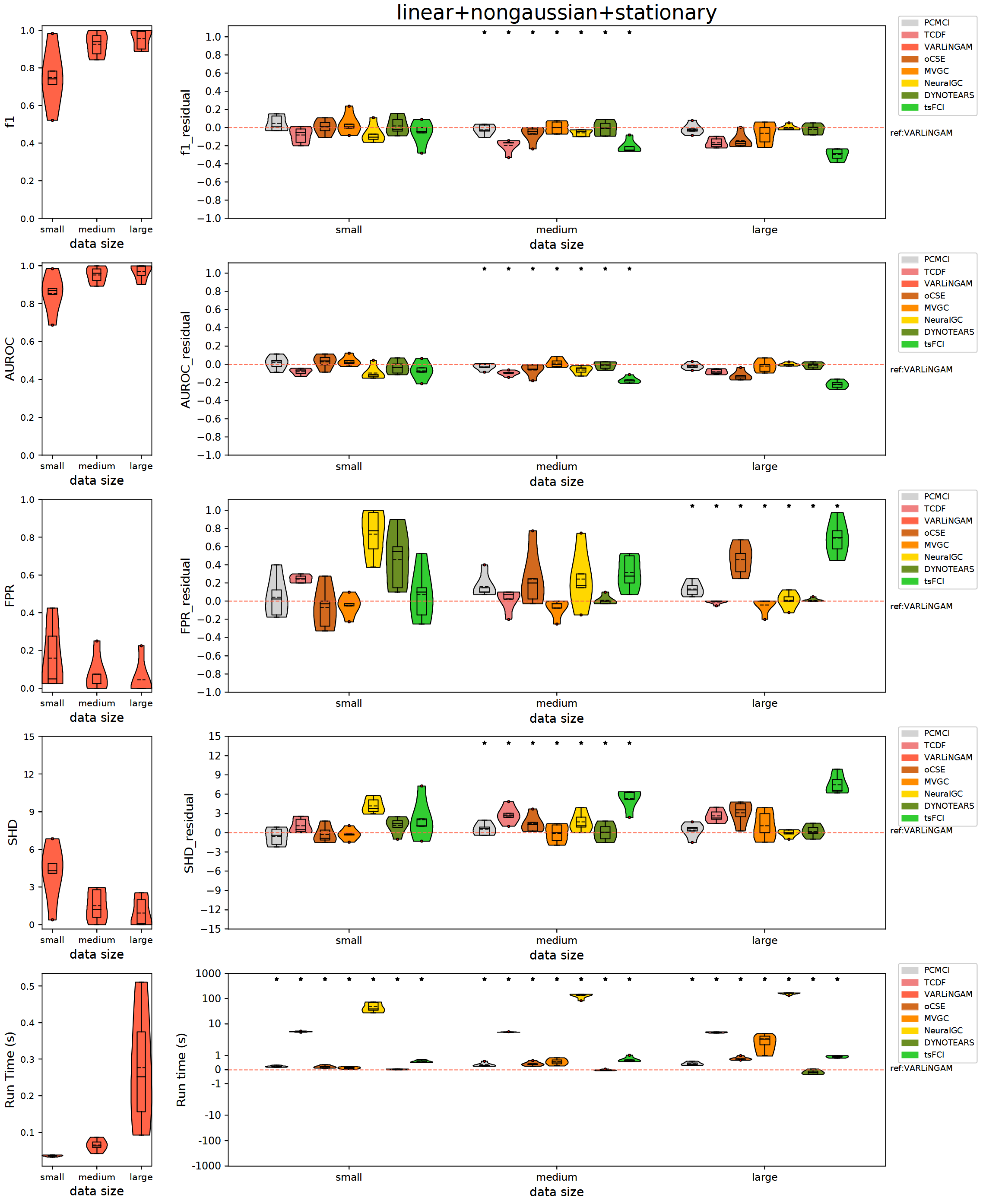}
    }
    \end{figure} 
\begin{figure}[htbp]
    \centering
     \subfigure[Nonlinear and Gaussian Datasets (time-delay)]{
    \label{fig.sub.3}
    \includegraphics[width=0.98\textwidth]{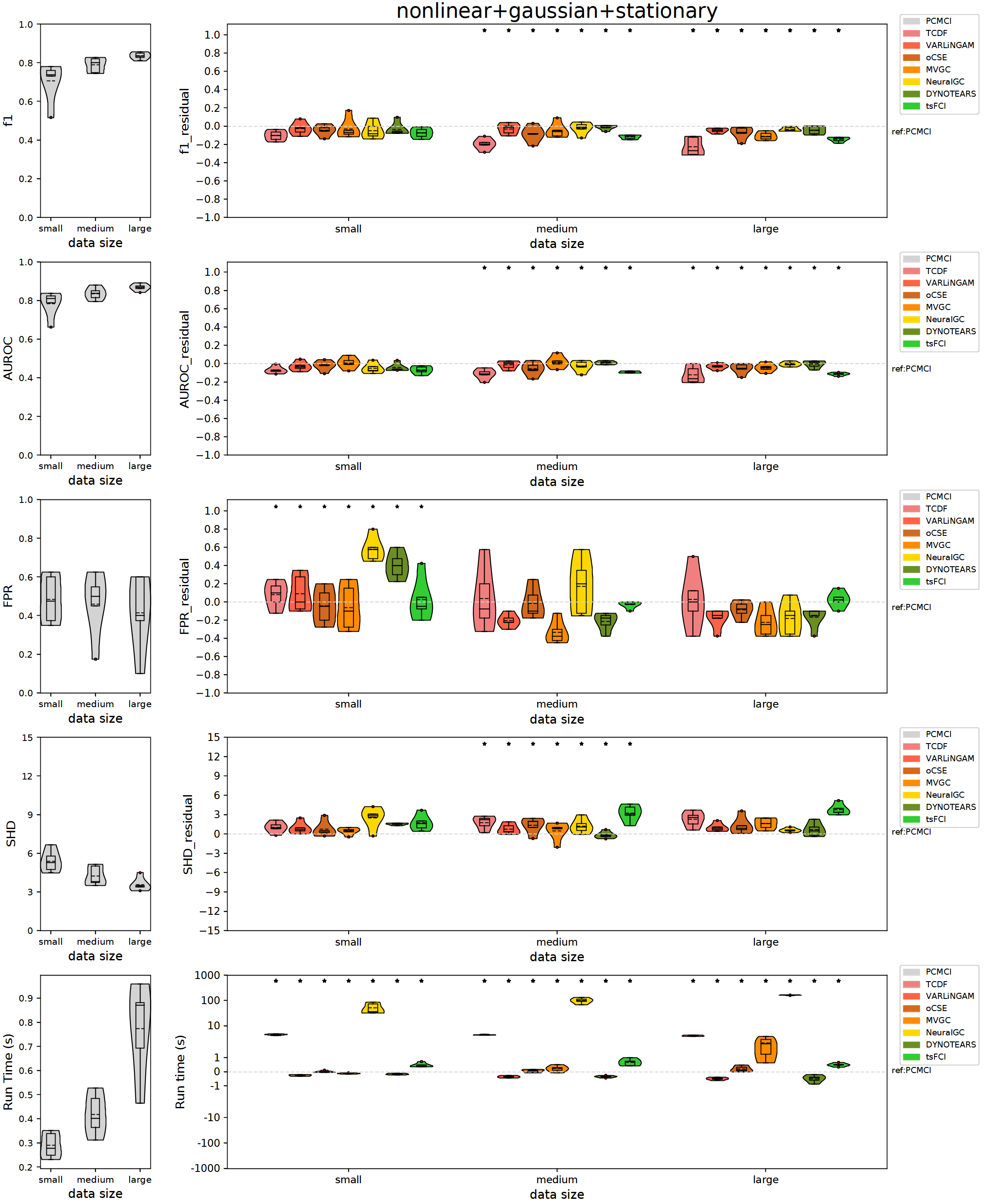}
    }
    \end{figure} 
    \begin{figure}[htbp]
    \centering
     \subfigure[Nonlinear and Non-Gaussian Datasets (time-delay)]{
    \label{fig.sub.4}
    \includegraphics[width=0.98\textwidth]{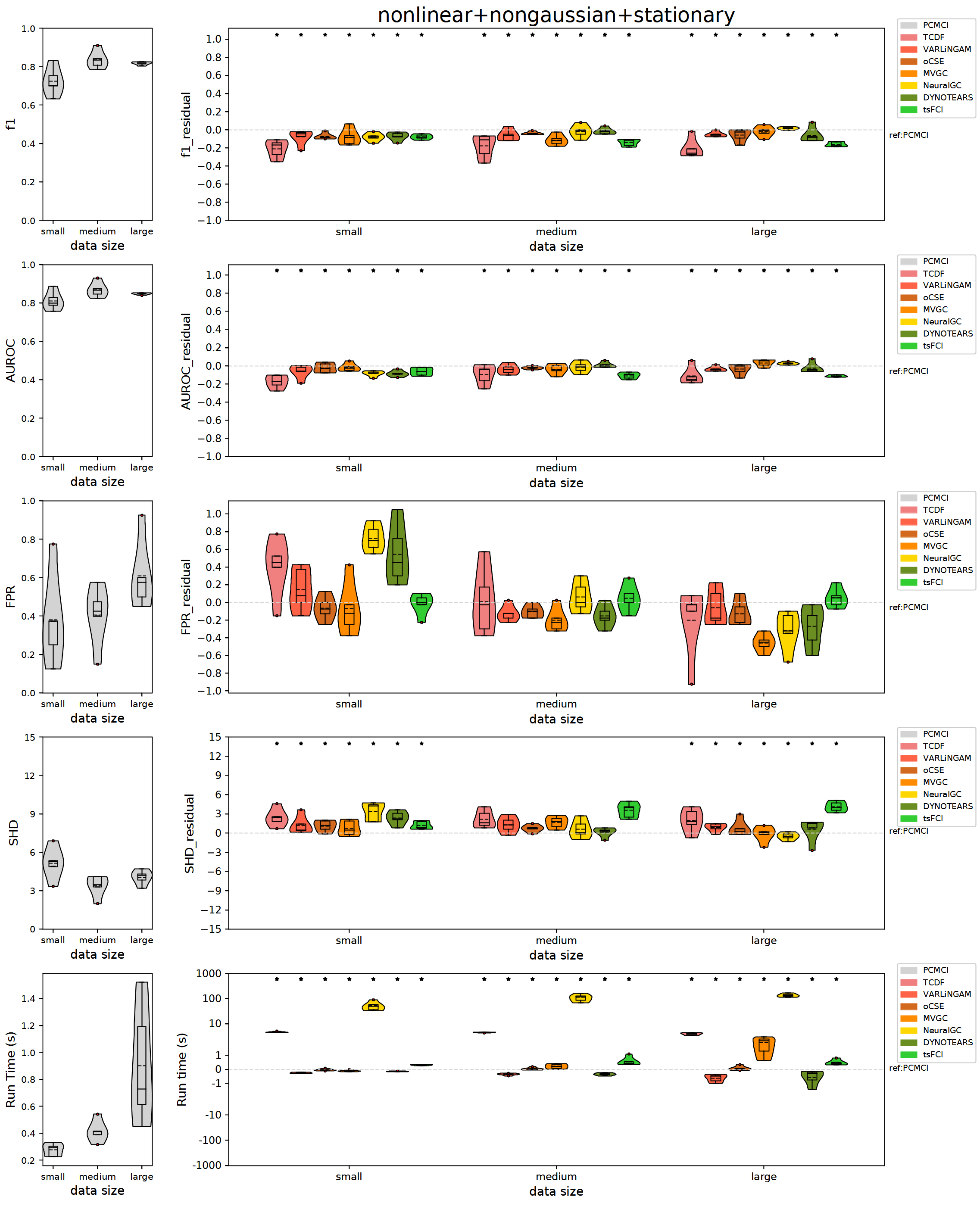}
    }
      
    \caption{Evaluation of time-delay causal algorithms (MVGC, PCMCI, TCDF, VARLiNGAM, NeuralGC, oCSE, DYNOTEARS, tsFCI) on synthetic datasets. The original violin plot of the reference algorithm is presented in the left column while the corresponding residual plot of alternative algorithms is in the right column. An asterisk denotes statistical significance determined by the Mann-Whitney U test with $p<0.05$.}
    \end{figure}

    \begin{figure}[htbp]
\label{iid_fig}
    \centering
    \subfigure[Linear and Gaussian Datasets (i.i.d.)]{
    \label{fig.sub.1}
    \includegraphics[width=0.98\textwidth]{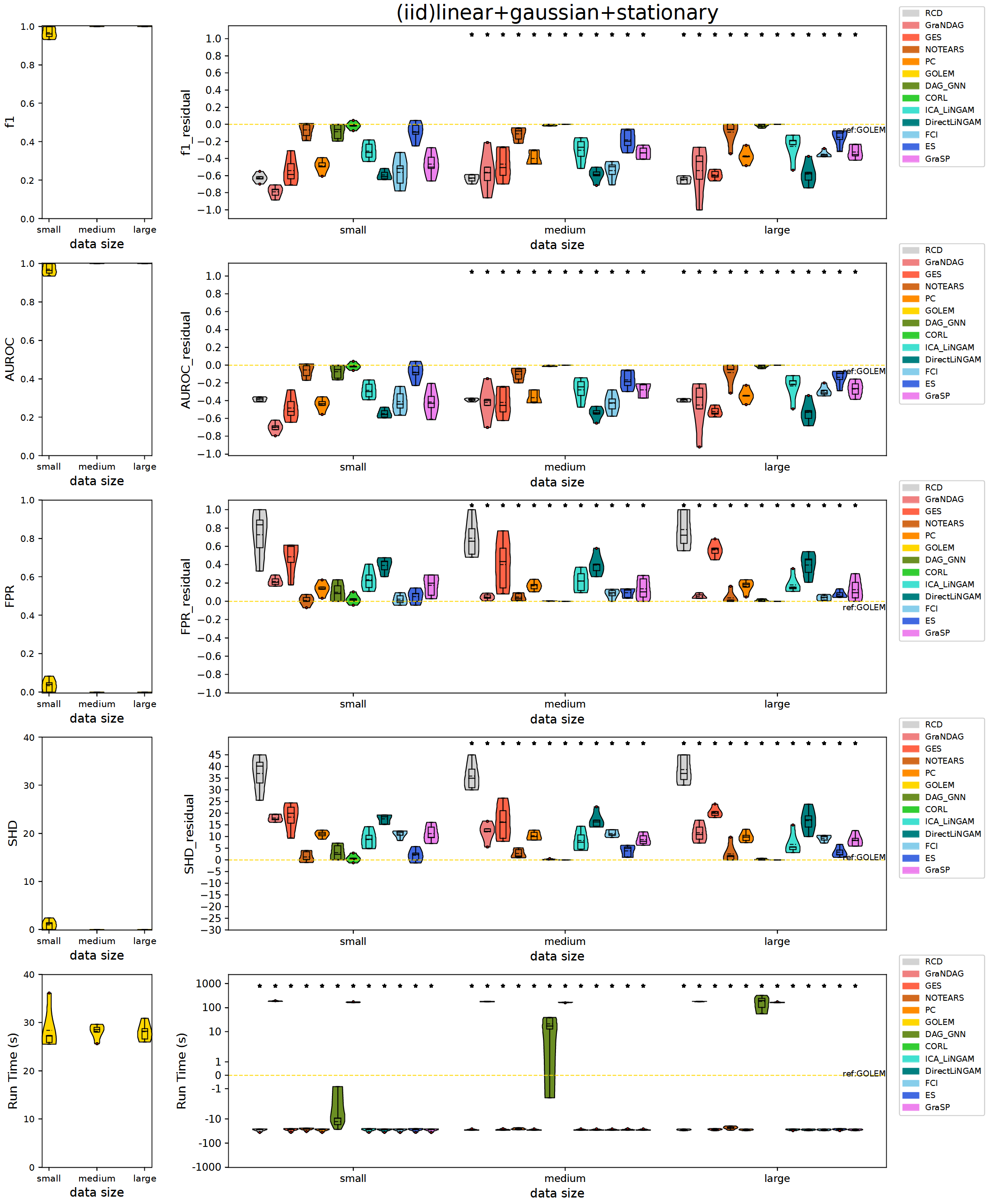}
    }
    \end{figure} 
    \begin{figure}[htbp]
    \centering
     \subfigure[Linear and Non-Gaussian Datasets (i.i.d.)]{
    \label{fig.sub.2}
    \includegraphics[width=0.98\textwidth]{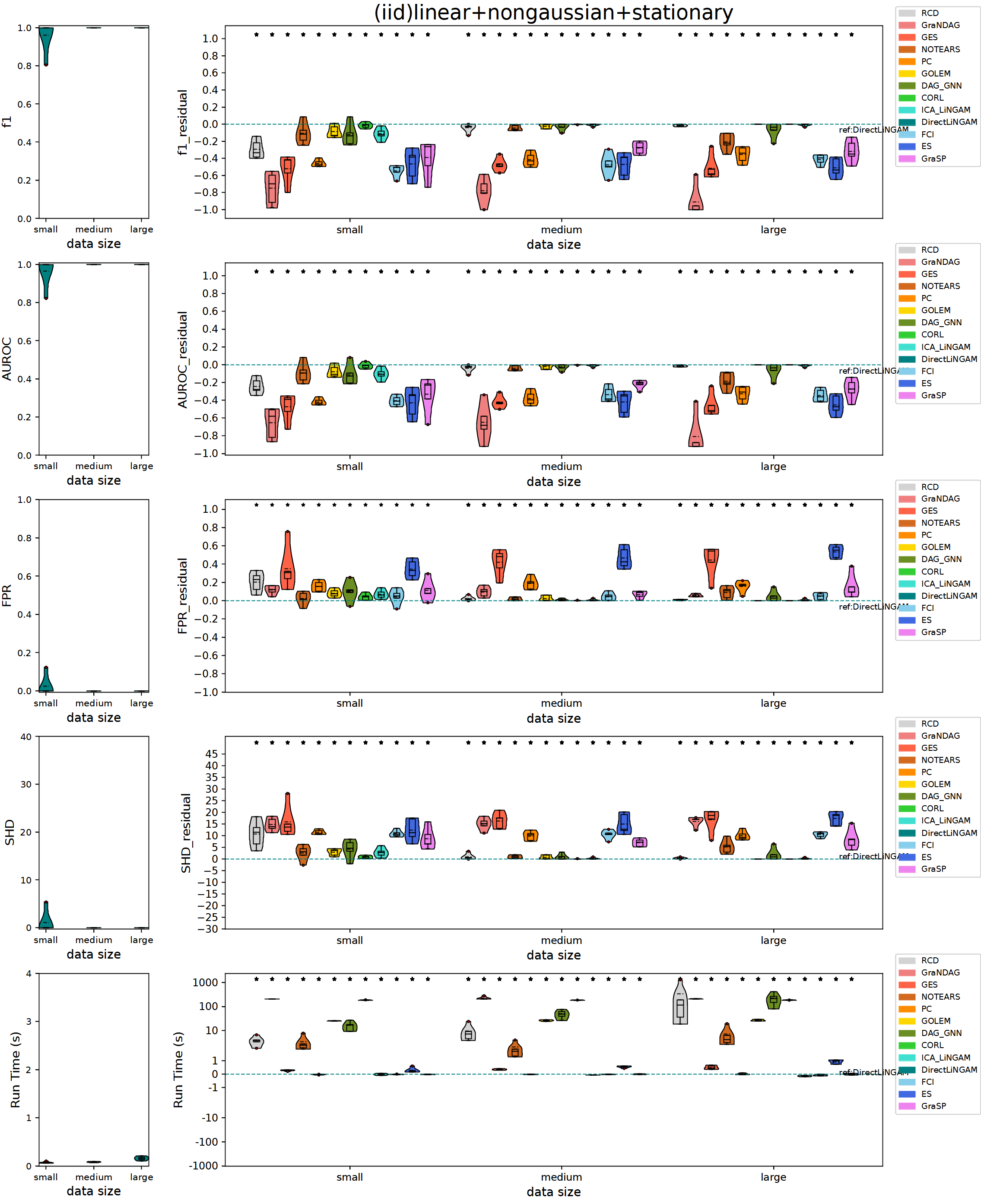}
    }
    \end{figure} 
\begin{figure}[htbp]
    \centering
     \subfigure[Nonlinear and Gaussian Datasets (i.i.d.)]{
    \label{fig.sub.3}
    \includegraphics[width=0.98\textwidth]{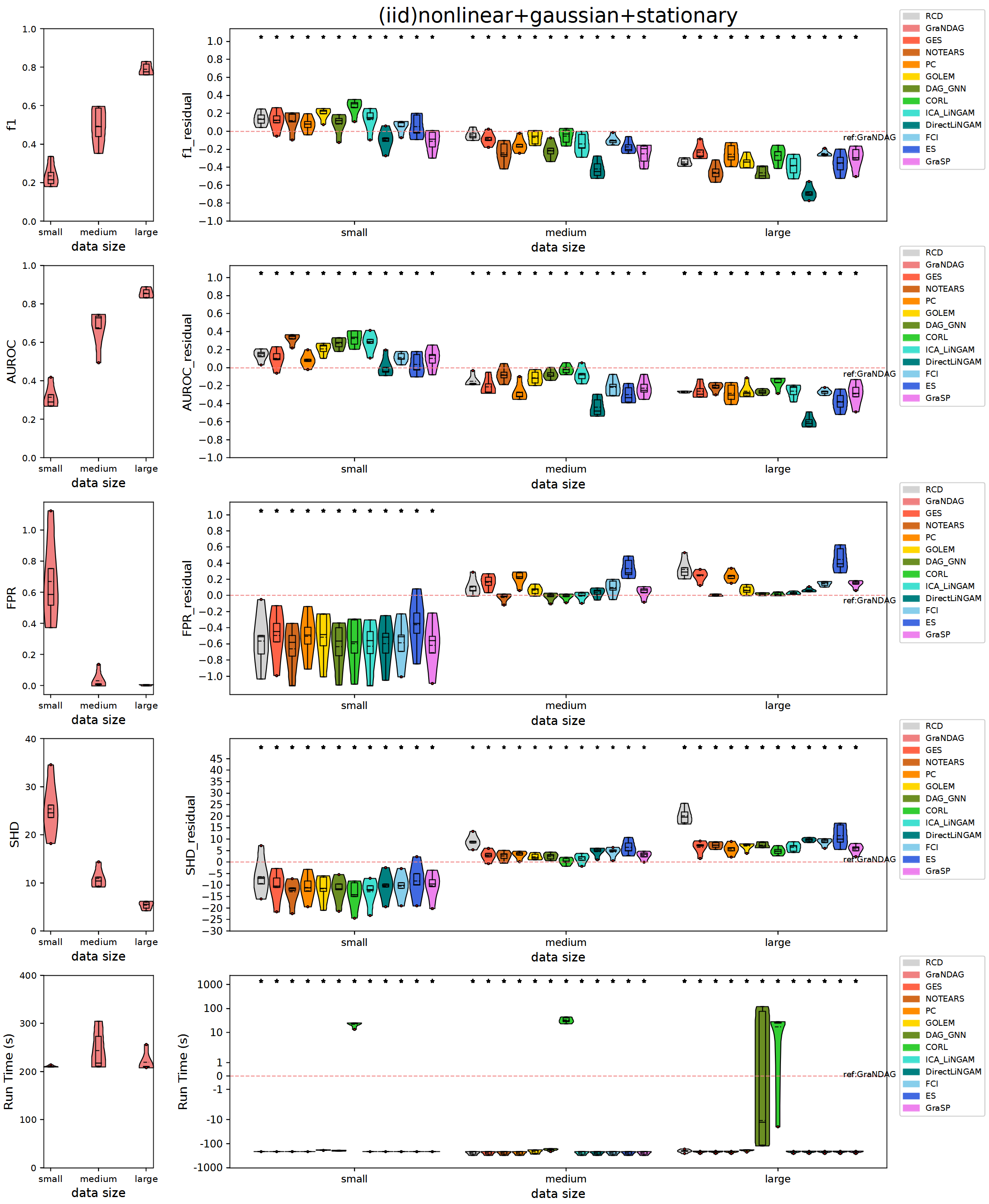}
    }
    \end{figure} 
    \begin{figure}[htbp]
    \centering
     \subfigure[Nonlinear and Non-Gaussian Datasets (i.i.d.)]{
    \label{fig.sub.4}
    \includegraphics[width=0.98\textwidth]{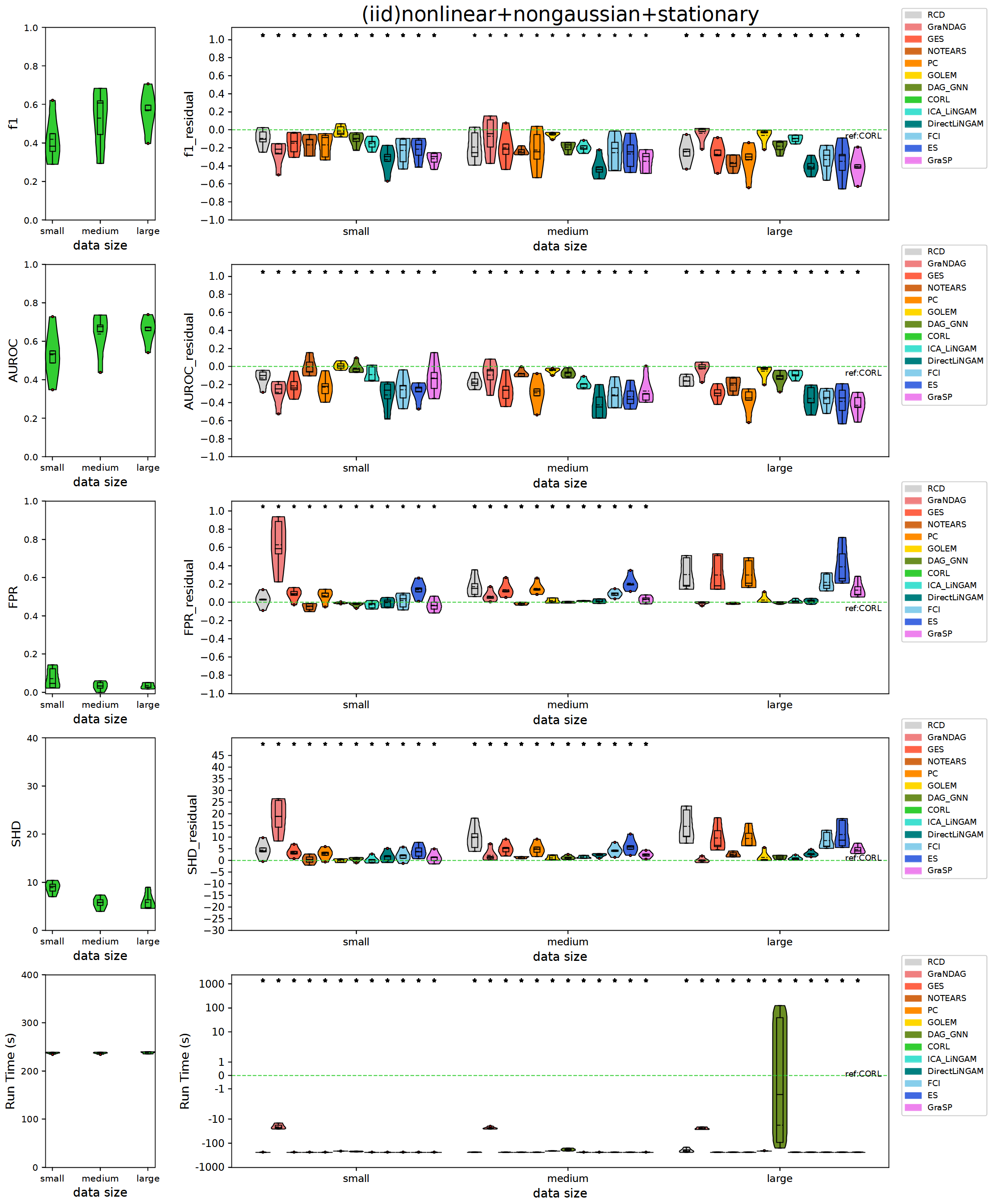}
    }
    \caption{Evaluation of i.i.d. causal algorithms (RCD, GraNDAG, GES, NOTEARS, PC, GOLEM, DAG-GNN, CORL, ICALiNGAM, DirectLiNGAM, FCI, ES, GRaSP) on synthetic datasets. The original violin plot of the reference algorithm is presented in the left column while the corresponding residual plot of alternative algorithms is in the right column. An asterisk denotes statistical significance determined by the Mann-Whitney U test with $p<0.05$.}
    \end{figure}

\end{appendix}

\vskip 0.2in

\end{document}